\documentclass[sigconf,balance=false]{acmart}
\usepackage{popets}

\setcopyright{none}
\copyrightyear{2025}
	
\acmYear{2025}
\acmVolume{2025}
\acmNumber{}
\acmDOI{}
\acmISBN{}
\acmConference{}
\usepackage{preamble}
\hypersetup{
    filecolor=black,
    urlcolor=gray,
}

\usepackage[normalem]{ulem}

\begin{document}

\title{Learning Privacy from Visual Entities}

%%%%%%%%%%%%%%%% Authors' Info %%%%%%%%%%%%%%%%%
\author{Alessio Xompero}
\orcid{0000-0002-8227-8529}
\affiliation{%
  \institution{Queen Mary University of London}
  % \city{London}
  % \country{U.K.}
  \city{}
  \state{}
  \country{}
  }
\email{a.xompero@qmul.ac.uk}

\author{Andrea Cavallaro}
\orcid{0000-0001-5086-7858}
\affiliation{%
  \institution{Idiap Research Institute\\\'{E}cole Polytechnique F\'{e}d\'{e}rale de Lausanne}
  % \city{Lausanne}
  % \country{Switzerland}
  \city{}
  \state{}
  \country{}
  }
\email{andrea.cavallaro@epfl.ch}

\renewcommand{\shortauthors}{Alessio Xompero and Andrea Cavallaro}

\begin{abstract}
Subjective interpretation and content diversity make predicting whether an image is private or public a challenging task. Graph neural networks combined with convolutional neural networks (CNNs), which consist of 14,000 to 500 millions parameters, generate features for visual entities (e.g., scene and object types) and identify the entities that contribute to the decision. In this paper, we show that using a simpler combination of transfer learning and a CNN to relate privacy with scene types optimises only 732 parameters while achieving comparable performance to that of graph-based methods. On the contrary, end-to-end training of graph-based methods can mask the contribution of individual components to the classification performance. Furthermore, we  show that a high-dimensional feature vector, extracted with CNNs for each visual entity, is unnecessary and complexifies the model. The graph component has also negligible impact on performance, which is driven by fine-tuning the CNN to optimise image features for privacy nodes.
\end{abstract}


\maketitle

\section{Introduction}

Image privacy classification is the task of assigning a label (e.g., public or private) to an image~\cite{Zerr2012CIKM_PicAlert,Orekondy2017ICCV,Zhao2022ICWSM_PrivacyAlert,Tonge2016AAAI,Tonge2018AAAI,Baranouskaya2023ICIP,Stoidis2022BigMM,Yang2020PR,Yang2022AAAI}.  Privacy classification can help prevent accidentally sharing images with private information~\cite{Zerr2012CIKM_PicAlert,Tonge2020TWEB} and explain what private information can be extracted from images~\cite{Ferrarello2022DRS, AriasCabarcos2023PoPETs}.
However, determining whether an image contains private information is challenging because privacy determinations are context-dependent and images vary in the content that people consider private~\cite{Ferrarello2022DRS}. 
Existing classification methods detect pre-defined private content (e.g.~credit card, license plate, semi-nudity)~\cite{Orekondy2017ICCV} or multiple entities (e.g.~objects and scene)~\cite{Tonge2016AAAI,Tonge2018AAAI}. Identifying relationships between these entities is key for classification~\cite{Marino2017CVPR} and understanding what makes an image private~\cite{Orekondy2017ICCV}.

Image privacy classification is also difficult due to limited training data, e.g.~between 5,000 images~\cite{Zhao2022ICWSM_PrivacyAlert} and 30,000 images~\cite{Orekondy2017ICCV,Zerr2012CIKM_PicAlert}. Because of this, existing methods either fine-tune the parameters of selected layers of a convolutional neural network (CNN) pre-trained on a much larger, source-domain dataset (e.g., ImageNet~\cite{Deng2009CVPR_ImageNet}), or replace and train the last fully connected (FC) layer on the downstream task (image privacy)~\cite{Orekondy2017ICCV,Tran2016AAAI_PCNH,Han2022MTA_PrivacyMLMS}. 

Graph-based methods~\cite{Jiao2020BIGCOM_IEye,Yang2022AAAI,Yang2020PR,Stoidis2022BigMM} apply graph neural networks (GNNs) in combination with CNNs~\cite{Yang2022AAAI}, or prior knowledge~\cite{Yang2020PR,Stoidis2022BigMM}, to encode the relationships between detected visual entities. 
Graph-based Image Privacy (GIP)~\cite{Yang2020PR} and Graph Privacy Advisor (GPA)~\cite{Stoidis2022BigMM} are trained end-to-end while also fine-tuning the CNNs to initialise the features used by the graph component~\cite{Yang2020PR,Stoidis2022BigMM}. These works claimed graph-based models as best-performing on publicly available datasets~\cite{Stoidis2022BigMM,Yang2020PR}. However, we will show that a substantially smaller model achieves the same performance. Our work is the first to demonstrate that using transfer learning~\cite{Bishop2024_DeepLearning} with a pre-trained CNN and only training the added FC layer suffices to achieve performance comparable to much larger graph-based models~\cite{Stoidis2022BigMM,Yang2020PR}.

In this paper, we evaluate the impact of fine-tuning the CNN to determine the real contribution of the graph component for the task. We consider the approach of recognising visual entities and  discuss various design choices to learn privacy from these entities. We especially focus on graph-based methods that aim at making the model inherently explainable through the graph itself. Specifically, we explore the following research questions:
\begin{itemize}
    \item What components of a graph-based method impact the classification performance for image privacy?
    \item Do the features refined by the graph component improve the performance for image privacy?
\end{itemize}

We estimate the cost of designing and training a graph-based method for image privacy classification and perform an in-depth analysis of design choices and training strategies to identify the component(s) driving the classification performance. We demonstrate that performance gains are only marginally due to the graph component and mostly due to the fine-tuning of the CNNs. We also show that the design of the model architecture is unnecessarily complicated when extracting large feature vectors from the CNNs. Additionally, we show that using transfer learning~\cite{TanTransferLearning,Bishop2024_DeepLearning} with a CNN suffices to achieve the highest classification performance on PrivacyAlert~\cite{Zhao2022ICWSM_PrivacyAlert} and Image Privacy Dataset~\cite{Yang2020PR}. By adding a FC layer while keeping all the parameters of the CNN fixed and pre-trained on a different source dataset, our approach only optimises 732 parameters for the downstream task, compared to the 500 million parameters of GIP or the 14,000 parameters of GPA.  

The paper is organised as follows. We discuss existing works for image privacy classification in Section~\ref{sec:realtedwork}. Section~\ref{sec:problemformulation} formulates the problem. Preliminaries on transfer learning and GNNs are presented in Section~\ref{sec:preliminaries}. We present how to combine pre-trained CNNs with a GNN to provide explainability of the model and include relationships between visual entities in Section~\ref{sec:graphdesign}. 
The experimental validation of Section~\ref{sec:experiments} compares the relative impact of individual components in graph-based methods and discusses the limitations of using only detected objects as visual entities for graph-based methods. Finally, we draw the conclusions in Section~\ref{sec:conclusion}. 

\section{Related work}
\label{sec:realtedwork}

Early privacy classifiers used hand-crafted features, such as SIFT~\cite{Lowe2004IJCV}, histogram of gradients~\cite{Dalal2005CVPR_HOG}, colour histograms, edges, or face aspect ratio for recognising the presence of private entities in an image~\cite{Zerr2012SIGIR,Squicciarini2014ACM_HSM,Buschek2015INTERACT}. CNNs enabled the extraction of features and classification of an image with a single model whose parameters can be learned in an end-to-end manner~\cite{Tonge2018MSM,Tonge2019WWW,Zhong2019BigData_RPM,Han2022MTA_PrivacyMLMS}. 
To overcome the problem of limited data in this task, CNNs are trained with a transfer learning strategy to predict an image as private~\cite{Tran2016AAAI_PCNH} (binary classification) or predict the likelihood of 68 privacy concepts in an image as user-independent, multi-label classification~\cite{Orekondy2017ICCV}. An object classifier and a privacy-specific classifier, both  connected to the same output layer, can be jointly trained by applying transfer learning only to the object classifier~\cite{Tran2016AAAI_PCNH}. 

Instead of training a model end-to-end, a different approach is to decouple the feature extraction from the classification step. In this approach, pre-trained CNNs are used to extract features, identify entities (and a confidence score), or compute additional features that can be used for training a classifier specialised on image privacy. For example, {\em deep features} are extracted from the last FC layers of a CNN pre-trained on ImageNet~\cite{Deng2009CVPR_ImageNet} for object recognition (1,000 classes) and used as input to a Support Vector Machine~\cite{Tonge2016AAAI,Tonge2020TWEB}, logistic regressor or multi-layer perceptron (MLP)~\cite{Baranouskaya2023ICIP}. To identify presence of the entities in the image, deep tags select the top-\textit{k} classes (or tags) with the most confident probabilities, as extracted by a deep neural network~\cite{Tonge2016AAAI}. Deep tags, represented as a binary vector after thresholding the probabilities of the selected classes, are provided as input feature to a classifier. Object tags were initially identified by a CNN pre-trained on ImageNet for object recognition~\cite{Tonge2016AAAI}, but object tags alone cannot always discriminate private images from public images. Scene tags, detected by a CNN pre-trained on Places365 for scene recognition~\cite{Zhou2018TPAMI_Places365}, are concatenated to the object tags to account for contextual information~\cite{Tonge2018AAAI}. Alternatively, a vector of eight privacy-specific human-interpretable features can be constructed from multiple pre-trained CNNs and provided as input to the classifier~\cite{Baranouskaya2023ICIP}. These features include the number of people localised in the image, the probability of the image being outdoors, or the presence of sensitive content (e.g. adult, violent, racy) as probabilities.

{\em Graph structures} account for the relationship between detected entities and are applied to various vision tasks\footnote{We refer the reader to Chang et al.'s work~\cite{Chang2023TPAMI_SceneGraphsSurvey} for a broader and in-depth survey of scene graphs applied to multiple tasks not limited to image classification.}, from object recognition, detection of human-object interaction, to advertisement classification~\cite{Chang2023TPAMI_SceneGraphsSurvey,Marino2017CVPR,kalanat2022symbolic}. 
None of these methods have yet been used for image privacy classification, but alternative graph-based methods have been devised for this task~\cite{Stoidis2022BigMM,Yang2020PR,Yang2022AAAI,Jiao2020BIGCOM_IEye,Yu2017TIFS_iPrivacy}.
By using the frequency of co-occurrent objects in a large set of social images, a tree-based classifier is constructed and jointly trained with a CNN for semantic segmentation to identify sensitive objects~\cite{Yu2017TIFS_iPrivacy}. 
An undirected and heterogenous two-layer semantic graph is designed with one layer consisting of nodes representing entities such as objects, person, and scene, and the other layer consisting of nodes representing attributes, such as bareness (percentage of pixels detected on a person as skin in the YCbCr colour space) or face score~\cite{Jiao2020BIGCOM_IEye}. Nodes are connected with edges representing relationships between entities and with edges linking entities to the attributes. To classify an image as private, interpretable and person-based rule sets are learnt from a user's dataset. These rule sets  consist of boolean expressions related to each other by a logical AND operation~\cite{Jiao2020BIGCOM_IEye}. 

\begin{table*}[t!]
    \centering
    \footnotesize
    \caption{Characteristics of image privacy classifiers. Top block: methods classifying images as private with end-to-end learning models, either directly or by extracting image-level feature vectors. Middle block: methods recognising visual entities, such as objects and scenes, and their features before further refinement and classification. Bottom block: methods based on recognising visual entities used in this work. Note the design differences between our models and corresponding methods (e.g., GIP, GPA).
    \vspace{-10pt}
    }
    \begin{tabular}{c cccccc ccccc ccccc}
    \toprule
    Method & \multicolumn{6}{c}{Objects} & \multicolumn{5}{c}{Scenes} & \multicolumn{5}{c}{Approach}     \\
    \cmidrule(lr){2-7}\cmidrule(lr){8-12}\cmidrule(lr){13-17}
    & Card  & Conf  & Pres  & Deep  & Num  & Dataset  & Conf  & Pres  & Deep  & Num & Dataset & TL    & FT    & Graph & GNN & Classifier \\
    \midrule
    Deep feats~\cite{Tonge2016AAAI}     & \wbox & \wbox & \wbox & \wbox & -    & -        & \wbox & \wbox & \wbox & -   & -       & \wbox & \wbox & \wbox & -   & SVM   \\
    *VPA~\cite{Orekondy2017ICCV}         & \wbox & \wbox & \wbox & \wbox & -    & -        & \wbox & \wbox & \wbox & -   & -       & \bbox & \bbox & \wbox & -   & -     \\
    DRAG~\cite{Yang2022AAAI}            & \wbox & \wbox & \wbox & \wbox & -    & -        & \wbox & \wbox & \wbox & -   & -       & \wbox & \wbox & \bbox & GCN & FC    \\   
    \midrule
    PCNH~\cite{Tran2016AAAI_PCNH}       & \wbox & \wbox & \wbox & \bbox & 204 & ImageNet & \wbox & \wbox & \wbox & -   & -       & \bbox & \wbox & \wbox & -   & FC   \\
    iPrivacy~\cite{Yu2017TIFS_iPrivacy} & \wbox & \wbox & \wbox & \bbox & 1000    & ImageNet        & \wbox & \wbox & \wbox & -   & -       & \bbox & \wbox & \bbox & -   & Tree  \\
    $^\diamond$PCS1~\cite{Tonge2020TWEB} & \wbox & \wbox & \bbox & \wbox & 1000   & ImageNet     & \wbox & \wbox & \wbox & -   & -       & \wbox & \wbox & \wbox & -   & Rule  \\
    $^\diamond$PCS2~\cite{Xompero2024CVPRW_XAI4CV} & \bbox & \bbox & \wbox & \wbox & 80   & COCO     & \wbox & \wbox & \wbox & -   & -       & \wbox & \wbox & \wbox & -   & Rule  \\
    $^\diamond$PCS3~\cite{Xompero2024CVPRW_XAI4CV} & \bbox & \bbox & \wbox & \wbox & 80   & COCO     & \wbox & \wbox & \wbox & -   & -       & \wbox & \wbox & \wbox & -   & Rule  \\
    $^\diamond$L8PS~\cite{Baranouskaya2023ICIP} & \bbox & \bbox & \wbox & \wbox & 80   & COCO     & \bbox & \wbox & \wbox & 365   & Places       & \wbox & \wbox & \wbox & -   & LR/MLP  \\
    MLP~\cite{Xompero2024CVPRW_XAI4CV} & \bbox & \bbox & \wbox & \wbox & 80 & COCO & \wbox & \wbox & \wbox & -- & -- & \wbox & \wbox & \wbox & -- & MLP \\
    GA-MLP~\cite{Xompero2024CVPRW_XAI4CV} & \bbox & \bbox & \wbox & \wbox & 80 & COCO & \wbox & \wbox & \wbox & -- & --  & \wbox & \wbox & \wbox & GA & MLP \\
    IEye~\cite{Jiao2020BIGCOM_IEye}     & \bbox & \wbox & \bbox & \wbox & 9000    & COCO + ImageNet        & \wbox & \bbox & \wbox & 365   & Places       & \wbox & \wbox & \bbox & -   & Rule     \\
    Deep Tags~\cite{Tonge2016AAAI}      & \wbox & \wbox & \bbox & \wbox & 1000 & ImageNet & \wbox & \wbox & \wbox & -   & -       & \wbox & \wbox & \wbox & -   & SVM   \\
    GIP~\cite{Yang2020PR}               & \wbox & \wbox & \wbox & \bbox & 81   & COCO     & \wbox & \wbox & \wbox & -   & -       & \bbox & \bbox & \bbox & GRM & MLP   \\
    Deep Tags~\cite{Tonge2018AAAI}      & \wbox & \wbox & \bbox & \wbox & 1000 & ImageNet & \wbox & \bbox & \wbox & 365 & Places  & \wbox & \wbox & \wbox & -   & SVM   \\
    GPA~\cite{Stoidis2022BigMM}         & \bbox & \wbox & \wbox & \wbox & 81   & COCO     & \wbox & \wbox & \bbox & 365 & Places  & \bbox & \bbox & \bbox & GRM & MLP   \\   
    \midrule
    MLP & \bbox & \wbox & \wbox & \wbox & 80 & COCO & \wbox & \wbox & \wbox & -- & -- & \wbox & \wbox & \wbox & -- & MLP \\
    GA-MLP & \bbox & \wbox & \wbox & \wbox & 80 & COCO & \wbox & \wbox & \wbox & -- & --  & \wbox & \wbox & \wbox & GA & MLP \\
    GIP & \wbox & \wbox & \wbox & \bbox & 80   & COCO     & \wbox & \wbox & \wbox & - & -  & \bbox & \wbox & \bbox & GRM & MLP \\
    GPA & \bbox & \wbox & \wbox & \wbox & 80   & COCO     & \wbox & \wbox & \wbox & - & -  & \bbox & \wbox & \bbox & GRM & MLP   \\
    S2P & \wbox & \wbox & \wbox & \wbox &  -- & -- & \wbox & \wbox & \bbox & 365 & Places & \bbox & \wbox & \wbox & -- & FC \\
    \bottomrule \addlinespace[\belowrulesep]
    \multicolumn{17}{l}{\parbox{0.90\linewidth}{\scriptsize{
    *Visual Privacy Advisor (VPA) is the only method performing multi-label classification of 67 entities predefined as private, and 1 public (\textit{safe}). Privacy of an image is computed as the maximum probability among the 68 entities. $^\diamond$Methods focused on the object \textit{person} and discard the others.\\
    KEY -- Card:~cardinality; Conf:~confidence; Pres:~presence as a binary value obtained via thresholding on confidence or using top-k selection; Deep:~features extracted by a convolutional neural network; Num:~number of pre-defined categories from a given dataset; TL:~transfer learning, FT:~Fine-tuning; GNN:~Graph Neural Network, SVM:~Support Vector Machine; GCN:~Graph Convolutional Network~\cite{Kipf2017ICLR_GCN}; GRM:~Graph Reasoning Model~\cite{Wang2018IJCAI}, FC:~fully connected layer; LR:~Logistic regressor; MLP:~multi-layer perceptron; PCSX:~person-centric strategy X with X being 1, 2 or 3; L8PS:~LR with eight privacy-specific features; GA:~graph-agnostic model reproducing the blocks of a GCN; \bbox:~considered, \wbox:~not considered; --:~not applicable.}}}
    \end{tabular}    
    \label{tab:relatedworks}
    \vspace{-10pt}
\end{table*}

Spatially-correlated feature channels, which are extracted by a pre-trained and fixed CNN, are clustered into pre-defined groups and aggregated into feature maps representing each cluster (region-aware feature maps)~\cite{Yang2022AAAI}. A correlation-based graph is constructed by adaptively identifying the correlation between the clusters using the self-attention mechanism, and used within a Graph Convolutional Network~\cite{Kipf2017ICLR_GCN} to refine the region-aware feature maps. The refined feature maps are concatenated with an image-level feature map to predict the privacy of an image. 
A prior graph that relates a pre-defined vocabulary of 80 objects with two class nodes (public and private) can be constructed based on the estimated frequency of each object with respect to each class in a training dataset~\cite{Yang2020PR}. 
A GNN~\cite{Wang2018IJCAI} refines the features by propagating, mixing, and smoothing local and global information. This approach was modified by constructing a prior graph that relates the objects via co-occurrence and replaces the deep features with only the number of object instances in an image (cardinality)~\cite{Stoidis2022BigMM}. A pre-trained classifier also extracts the logits associated with the scene tags and uses a trainable FC layer to map the logits to the two class nodes. 

Table~\ref{tab:relatedworks} summarises the characteristics of the methods described in this section.
GIP~\cite{Yang2020PR} and GPA~\cite{Stoidis2022BigMM} are the most relevant for our work because the CNN and GNN components can easily be decoupled and their implementation is publicly available. As the two models are trained end-to-end, it is unclear which factor is driving their classification decision. In the next sections, we will review the design of graph-based methods using GIP and GPA as references, and we will validate the impacts of individual components to the performance.

\section{Problem definition}
\label{sec:problemformulation}

Let the model $f_{\theta}(\cdot)$ predict if an input image $I$ is private, i.e.~$f_{\theta}(I) = y \in \{0,1\}$, where 0 denotes the label public and 1 the label private. The model parameters $\theta$ are trained on an annotated dataset, $\mathcal{T}=\{(I, y)_n\}_{n=1}^N$, where $N$ is the number of images (e.g., $N=10,000$, \textit{Problem 1}: limited training data). Moreover, the number of images labelled as public is much higher than the number of images labelled as private (\textit{Problem 2}: class imbalance). 

For example, existing datasets~\cite{Zhao2022ICWSM_PrivacyAlert,Orekondy2017ICCV,Zerr2012CIKM_PicAlert} collected images from online sources with keywords from a multi-category privacy taxonomy (e.g., \textit{nudity, medical, drinking/party, appearance/facial expression, religion/culture, personal information}) and instructs users to annotate these images according to guidelines that differ across datasets. These annotations are either already binary or mapped to the binary labelling, and only the binary labels are made publicly available. Appendix~\ref{app:labelinconsistencies} discusses privacy annotations and inconsistencies in the existing datasets.

We consider the problem of recognising visual entities such that the privacy model is refined as $y=f_{\theta}(d_{\eta}(I))$. The model $d_{\eta}(\cdot)$, where $\eta$ represents the model parameters, recognises a pre-defined
set of visual entities (e.g, generated by an object detector or a scene classifier).

\section{Preliminaries}
\label{sec:preliminaries}

\subsection{Transfer learning}

\begin{figure}[t!]
    \centering
    \includegraphics[width=0.98\linewidth]{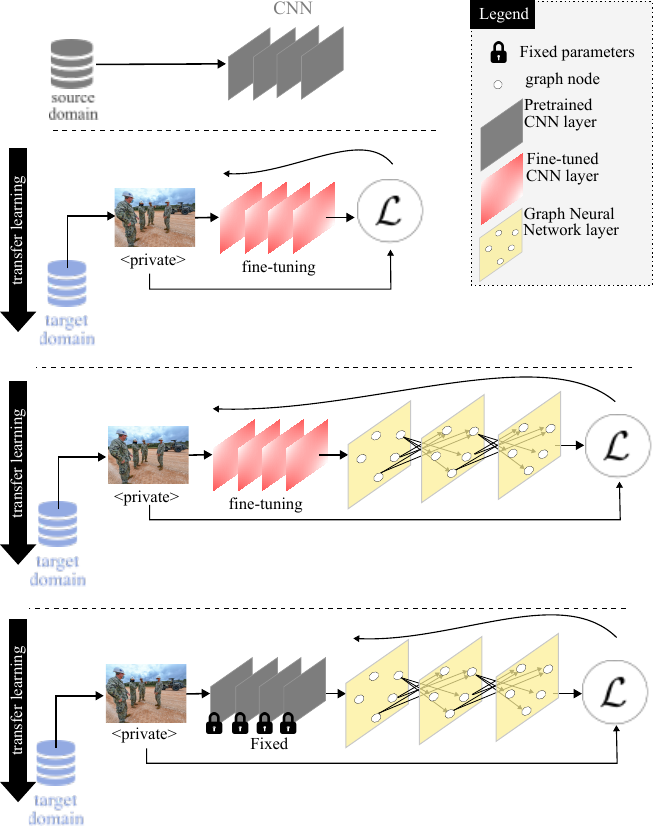}
    \caption{Illustration of different training strategies to understand the relative contribution of individual components in a learning-based model for image privacy classification (illustration inspired and adapted from \cite{Modas2021ICIP}). On top, a transfer learning based strategy that fine-tunes the layers of a convolutional neural network (CNN), initially pre-trained on a source domain, to the downstream task (image privacy) on a target domain. In the middle, an  architecture that stacks a CNN with a graph neural network (GNN), and is trained end-to-end while also fine-tuning the CNN parameters. This strategy masks the relative contribution of the GNN to the overall performance. On the bottom, the training involves only the GNN while keeping the parameters of the CNN fixed to understand the relative contribution of the GNN.
    }
    \label{fig:mainfig}
    \vspace{-10pt}
\end{figure}

To overcome the problem of limited training data, transfer learning~\cite{Bishop2024_DeepLearning} uses an additional training set for a different task (source domain), $\mathcal{S}=\{(I, z)_m\}_{m=1}^M$ with $|\mathcal{S}| = M >> N$ and $z$ being a multi-class label (e.g., objects~\cite{Deng2009CVPR_ImageNet}), to pre-train the parameters of $f_{\theta}$ (pre-training). The model is then further trained on the dataset $\mathcal{T}$ (target domain) by replacing the last layer of the model with a FC layer that transforms the features from the previous layer into the logits of the output classes of the new domain~\cite{Bishop2024_DeepLearning}. A different transfer learning strategy is \textit{fine-tuning} where all or some parameters of the pre-trained model are optimised to the task in the target domain (see Fig.~\ref{fig:mainfig})~\cite{Bishop2024_DeepLearning,Modas2021ICIP,TanTransferLearning}. For example, a CNN (e.g., ResNet~\cite{He2016CVPR_ResNet}) can be trained on ImageNet~\cite{Deng2009CVPR_ImageNet} (more than 14 million images) for object recognition or on Places~\cite{Zhou2018TPAMI_Places365} for scene recognition before being refined on a dataset for image privacy~\cite{Zhao2022ICWSM_PrivacyAlert,Yang2020PR}.

Similarly to previous works~\cite{Tonge2016AAAI,Tonge2018AAAI,Tonge2020TWEB}, we consider a pre-trained CNN and apply transfer learning by keeping the parameters fixed to recognise visual entities (e.g., scenes). Instead of applying the sigmoid function to the logits outputted by the model for all pre-defined scenes, we add a trainable FC layer (\textit{scene-to-privacy layer}) that transforms the scene logits to two logits, $f(\cdot): \mathbb{R}^{365} \rightarrow \mathbb{R}^2$, followed by softmax to obtain the probability distribution of private and public classes. The parameters of this layer, $W_s$, are randomly initialised and then optimised during the end-to-end training of the model. 
For simplicity, we refer to this method that relates scenes to privacy as S2P. 

As a binary classification task, the supervised training of the model minimises the cross-entropy loss,
\begin{equation}
    \mathcal{L} = - \frac{1}{B} \sum_{b=1}^B \sum_{j \in \mathcal{V}_p} \left( y_b \log \left( p_{j,b} \right) \right),
    \label{eq:celoss}
\end{equation}
where $B$ is the number of images in a training batch $\mathcal{B} \subset \mathcal{T}$, $y_b$ is the annotated label (public or private) for image $b$ in the training batch, and $p_{j,b}$ is the probability outputted by the classifier for the privacy node $j$ and image $b$.

\subsection{Graph neural networks}

Let $\mathcal{G} = (\mathcal{V}, \mathcal{E})$ be a graph with nodes $\mathcal{V}$ and edges $\mathcal{E}$. The structure of the graph is represented by an adjacency matrix relating nodes to each other (edges), $\mtrx{A}_{K\times K}$, where $K = |\mathcal{V}|$ is the number of nodes and $|\cdot|$ is the cardinality of a set. The elements of $\mtrx{A}_{K\times K}$ are binary values when $\mathcal{G}$ is unweighted.
$\mathcal{G}$ is undirected if nodes are connected to each other in both directions, otherwise the graph is directed. Undirected and unweighted graphs, and undirected graphs with the same weights in both directions, have a symmetric adjacency matrix, i.e. $\mtrx{A} = \mtrx{A}^\top$, where $\top$ is the transpose symbol. 

A GNN~\cite{Scarselli2009TNN} has $L$ layers that (1) refine features, and (2) propagate, aggregate, and update the features based on the graph structure.
Stacking multiple layers defines a composition of permutation-equivariant parametrised functions~\cite{Muller2024TMLR}.

Each node $v \in \mathcal{V}$ is represented by a feature vector $\mtrx{f}_v \in \mathbb{R}^D$, where $D$ is the dimensionality of the vector. The matrix $\mtrx{X} \in \mathbb{R}^{K\times D}$, stacking the feature vectors of all nodes, is the input to a GNN. The propagation of the features across nodes at each layer of the network is also referred to as message passing. For each layer $l$, a matrix of trainable parameters, $\mtrx{W}_l \in \mathbb{R}^{D\times D'}$, is trained to refine the node features as $\mtrx{X}' = \mtrx{X}\mtrx{W}_l$. If $D'=D$, the dimensionality of the features remains the same across the layers. For simplicity, we use $D$ and $D'$ as the input and output dimensionalities of the layer in a generic way, and their value can differ across layers.

The features of node $v$ are then aggregated with the features from the neighbour nodes  $\mathcal{N}_v$ based on the graph structure $\mtrx{A}$. The aggregation can be performed with a simple sum operation where the same node $v$ can be included (denoted as self-loop), or with more complex operations, such as convolution for Graph Convolutional Networks~\cite{Kipf2017ICLR_GCN}, attention for Graph Attention Networks~\cite{Velickovic2018ICLR_GAT} or gating for Gated GNNs (GGNNs)~\cite{Li2016ICLR_GGNN}. Aggregated features are  updated, for example, with a non-linearity point-wise operation such as Rectified Linear Unit (ReLU), and passed to the next layer~\cite{Scarselli2009TNN,Muller2024TMLR,Dwivedi2023JMLR}. This local message passing  implies that the information available at a specific node $v$ is propagated only to nodes that are at 1-hop distance from one layer to another. Nodes at a $L$-hop distance will require $L$ layers to receive the information of node $v$~\cite{battaglia2018relationalinductivebiasesdeep}. 

Graphs and GNNs can be used for various task, such as supervised and semi-supervised node classification~\cite{Kipf2017ICLR_GCN,Dwivedi2023JMLR}, link prediction~\cite{Dwivedi2023JMLR}, and graph classification~\cite{Dwivedi2023JMLR}. For graph classification, special nodes can be included or the features refined at the last layer can be aggregated from all nodes into a single feature vector and provided as input to a classifier. For a comprehensive overview of GNNs and recent methods, we refer the reader to existing surveys~\cite{battaglia2018relationalinductivebiasesdeep,Bronstein2017SPM,Chen2024TPAMI,Dwivedi2023JMLR,Gilmer2017ICML,Han2022NeurIPS_ViGNN,Han2023ICCV_VHGNN,Li2023TPAMI_DeepGCN,Muller2024TMLR,Rampavsek2022NeurIPS_GPS,Wu2021SurveyGNN}.

%%%%%%%%%%%%%%%%%%%%%%%%%%%%%%%%%%%%%%%%%%%%
\section{Graph-based classifiers}
\label{sec:graphdesign}

\begin{figure*}[t!]
    \centering
    \includegraphics[width=0.98\linewidth]{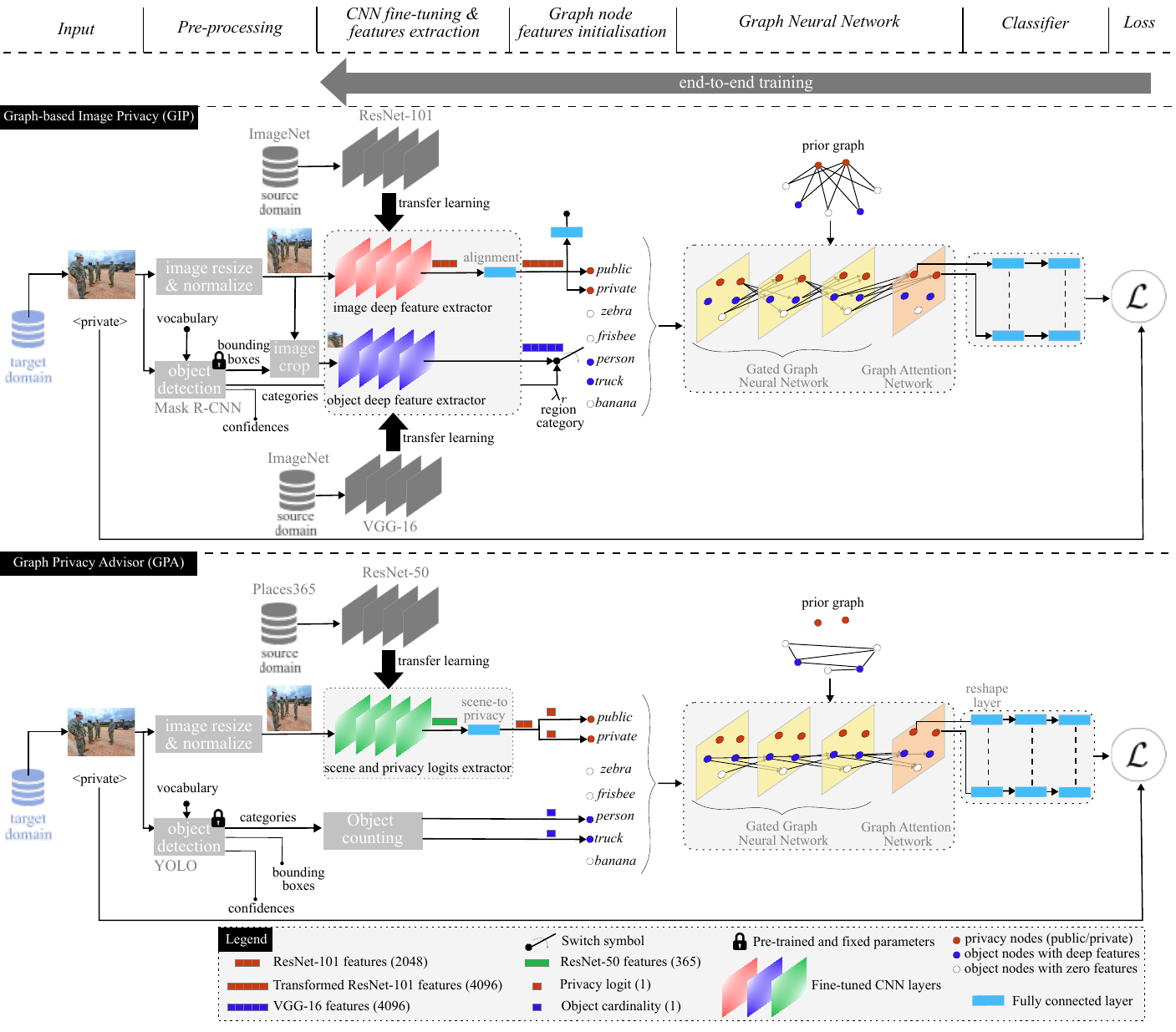}
    \caption{Illustrative diagrams of GIP~\cite{Yang2020PR} (top) and GPA~\cite{Stoidis2022BigMM} (bottom), and their end-to-end training. The methods have two main components: Convolutional Neural Networks (CNNs) to extract deep features and a Graph Neural Network (GNN) to refine the features based on a graph computed a priori. The graph is designed with two node types: privacy nodes (public and private) and object nodes. The number of object nodes is determined by a pre-defined and fixed-size vocabulary (e.g., 80 categories in the COCO dataset~\cite{Lin2018ECCV_COCO}). After initialising the node features, the GNN refines the features based on the prior graph. The features of the privacy nodes at the last layer are used as input to a classifier that consists of multiple fully connected layers with shared parameters. Both models are trained end-to-end with a cross-entropy loss $\mathcal{L}$, also guiding the fine-tuning of the CNNs. Note that the image is resized and normalised based on the statistics computed from ImageNet. Note that some of the connections in the GNNs blocks are omitted from the visualisation.
    }
    \label{fig:graphmodeldiagram}
    \vspace{-10pt}
\end{figure*}

We review and compare two graph-based methods, GIP~\cite{Yang2020PR} and GPA~\cite{Stoidis2022BigMM} (see Fig.~\ref{fig:graphmodeldiagram}).  We discuss the design choices of graph node features, GNN architecture, and  classifier.

\subsection{Graphs relating visual entities and privacy}

For image privacy classification, existing methods define a heterogeneous graph $\mathcal{G}_p = (\mathcal{V}_o, \mathcal{V}_p, \mathcal{E})$ consisting of object nodes $\mathcal{V}_o$ and privacy nodes $\mathcal{V}_p$~\cite{Yang2020PR,Stoidis2022BigMM}. Object nodes represent object categories that can be predicted by a pre-trained object detector~\cite{Lin2018ECCV_COCO}. Let $K_o = |\mathcal{V}_o|$ be the number of object categories. As these methods address image privacy as a binary classification task, the privacy nodes are designed to represent the two outputs, public and private classes: $K_p = |\mathcal{V}_p| = 2$. Let $\mathcal{V} = \mathcal{V}_o \cup \mathcal{V}_p$ be the set of all nodes with $K = K_o + K_p$ being the total number of graph nodes. The number of nodes depends on the pre-defined vocabulary of the datasets where  CNNs are trained on. The edges of the graph, $\mathcal{E}$, are designed to relate object nodes to the privacy nodes~\cite{Yang2020PR} or object nodes to each other~\cite{Stoidis2022BigMM}. After constructing a general graph that is not specific to an image, these methods~\cite{Yang2020PR,Stoidis2022BigMM} use the prior graph and the node features extracted from an image $I$ as input to the GNN to classify the image. Image privacy is therefore addressed as a graph-level classification by these methods.

Object and privacy nodes can be related to each other based on the frequency of the objects in the training set $\mathcal{T}$ with respect to each privacy class~\cite{Yang2020PR} as
\begin{equation}
\label{eq:objfreq}
    \mtrx{A}_{v,y} = \frac{M_{v,y}}{M_y},
\end{equation}
where $M_{v,y}$ is the number of images associated to the label $y$ and including the object $v$, and $M_y$ is the number of images associated to the label $y$. This prior graph is bipartite, weighted and undirected, allowing the features (message) flowing from the object nodes to the privacy nodes and vice-versa across the layers of the GNN. 

Alternatively, edges of the prior graph, \mbox{$\mathcal{E} =\{ \varepsilon_{i,j} | i,j \in {V}_o\}$} with $\varepsilon_{i,j} \in \{0,1\}$, can be defined by relating object nodes with each other based on the co-occurrence criterion in the training set $\mathcal{T}$, 
\begin{equation}
    \varepsilon_{i,j} = 
    \begin{cases} 
    1 & \text{if } \exists I_n \in \mathcal{T} \quad \text{s.t.} \quad C_{i,n} > 0 \wedge C_{j,n} > 0,  \\
    0   & \text{otherwise},
  \end{cases}
\end{equation}
i.e., an edge exists if two different objects co-occur in at least one image, and $C_{i,n}$ is the cardinality that counts the number of object instances localised in the image $n$ and belonging to the corresponding object category. This prior graph is unweighted and undirected. 

GIP~\cite{Yang2020PR} uses the bipartite prior graph where objects are not directly related to each other, whereas GPA~\cite{Stoidis2022BigMM} uses the second prior graph that does not consider object-class relationships.

\subsection{Node features}

Designing GNNs requires the definition of features for the graph nodes. The initialised features are then refined by the GNN. However, this definition is task-dependent and for image privacy, defining relevant features is not trivial and there is not yet common consensus across various research works~\cite{Yang2020PR,Stoidis2022BigMM,Baranouskaya2023ICIP,Tonge2020TWEB,Jiao2020BIGCOM_IEye,Yu2017TIFS_iPrivacy}. We present and review two approaches to initialise the node features.

The first approach extracts high-dimensional feature vectors from an input image using CNNs (deep features)~\cite{Yang2020PR}. Object nodes are initialised with deep features extracted by a VGG-16 network~\cite{Simonyan2014ICLR_VGG} (vector of 4,096 features) from the regions localised in the image by an object detector~\cite{He2017ICCV_MaskRCNN}, $d(\hat{I}; v)$. Let $\{(\mtrx{p}_r, \lambda_r) \}_{r=1}^R$ be the set of $R$ localised regions where $\mtrx{p}_r = I([x_r, y_r, w_r, h_r])$ is the region identified by the detected bounding box with top-left corner ($x_r, y_r$), and width and height ($w_r, h_r$) on the image $I$; and $\lambda_r \in \mathcal{V}_o$ is the corresponding object. Therefore, let $\gamma(\hat{\mtrx{p}}_r)$ be the function, represented by the VGG-16 network, that extracts the features from the image region resized to a predefined resolution, $\hat{\mtrx{p}}_r$. 
How to initialise the privacy nodes is a challenging design choice and one option is to extract image-level features, e.g., using a ResNet-101 network~\cite{He2016CVPR_ResNet}, $r_o(\hat{I})$. This network extracts a vector of 2,048 features from the input image resized to a predefined resolution. A trainable FC layer, $g(r_o(\hat{I}), W_s): \mathbb{R}^{2,048} \rightarrow \mathbb{R}^{4,096}$ where $W_s$ contains the parameters to learn, can align the dimensionality of this vector with the feature vectors outputted by VGG-16. During the end-to-end training, the parameters of this layer are randomly initialised and then optimised for image privacy classification. Therefore, the feature vector of a graph node, $\mtrx{f}_v \in \mathbb{R}^D$ with $D=4,098$, is 
\begin{equation}
    \mtrx{f}_v = \begin{cases}
        [1, 0, g\left( r_o(I), W_s \right)], & \quad \text{if } v \in \mathcal{V}_p, \\
        [0, 1, \gamma\left( \hat{\mtrx{p}}_r \right)], & \quad \text{if } v \in \mathcal{V}_o \wedge \lambda_r=v, \\
        [0, 1, \mtrx{0}], & \quad \text{if } v \in \mathcal{V}_o \wedge d(\hat{I}; v) \rightarrow \emptyset, \\
    \end{cases}
\end{equation}
where $\wedge$ is the logical AND operator, and $\emptyset$ is the empty set. Object nodes that are not associated with any object localised in the image are initialised with a vector of zero values, $\mtrx{0}$, whose dimensionality is 4,096. Also, note the 1-hot vector to distinguish the privacy nodes ($[1,0]$) from the object nodes ($[0,1]$). 

The second approach is to design node-features that are human-interpretable as well as minimal and sufficient for the classification task while better understanding their influence and relevance to privacy~\cite{Stoidis2022BigMM}.
Object nodes can be initialised with a single feature, $C_v$, i.e. the object cardinality defined as
\begin{equation}
    C_v = | \{ (\mtrx{p}_r, \lambda_r) ; \lambda_r = v\}_{r=1}^R |.
\end{equation}
As cardinality cannot be defined for the privacy nodes, deep features representing the whole image can be transformed into a 2D vector, $\mtrx{s} = f\left( r_s(I), W_s \right)$ and $\mtrx{s} \in \mathbb{R}^2$. Each element is used for each privacy node. In this case, the features for each graph node are defined as
\begin{equation}
    \mtrx{f}_v = \begin{cases}
        [1, \mtrx{s}_v], & \quad \text{if } v \in \mathcal{V}_p, \\
        [0, C_v], & \quad \text{if } v \in \mathcal{V}_o \wedge \lambda_n=v, \\
        [0, 0], & \quad \text{if } v \in \mathcal{V}_o \wedge d(\hat{I}; v) \rightarrow \emptyset, \\
    \end{cases}
\end{equation}
where a flag is included to distinguish privacy and object nodes. 

GIP~\cite{Yang2020PR} initialises the graph nodes based on the deep features approach. Specifically, the method uses a VGG-16 pre-trained on ImageNet~\cite{Deng2009CVPR_ImageNet} and a Mask R-CNN based detector~\cite{He2017ICCV_MaskRCNN} pre-trained on COCO~\cite{Lin2018ECCV_COCO}. If multiple instances of the same object category are localised in the image, GIP uses the feature vector associated with the last instance, as ordered by the detector, to initialise the corresponding node. On the contrary, GPA~\cite{Stoidis2022BigMM} uses the second approach to initialise the node features. Specifically, GPA uses S2P to transforms the 365 scene logits to two features, $f(\cdot): \mathbb{R}^{365} \rightarrow \mathbb{R}^2$, each associated with the private and public nodes.

\subsection{Model architecture and processing}
 
GNNs requires the definition of the architecture, the aggregation function and the update function. We review the GNN used by existing graph-based models for image privacy classification.

Graph Reasoning Model\footnote{Graph Reasoning Model~\cite{Wang2018IJCAI} was originally designed for the recognition of social interactions in images by modelling a graph that relates objects with relationships.} (GRM)~\cite{Wang2018IJCAI} is an architecture that stacks a 3-layer GGNN~\cite{Li2016ICLR_GGNN} with a modified Graph Attention Network~\cite{Velickovic2018ICLR_GAT} consisting of a single layer. 
The prior graph and the node features are provided as input to the GGNN that refines and propagates the features in the hidden statuses, inspired by the Gated Recurrent Unit~\cite{chung2014empirical}. Specifically, two gating mechanisms, an update gate and a reset gate, are used to aggregate the refined features from the neighbour nodes at the node $v$ for each layer $l$. Note that GGNN splits the adjacency matrix into two square matrices, the one representing the outgoing edges and the other representing the incoming edges to each node in the graph, and concatenated together for the processing. These two matrices have their own parameters depending on the direction of the edges. We refer the reader to the GGNN paper~\cite{Li2016ICLR_GGNN} for details of the mathematical formulation. 

The refined features $\mtrx{X}'_{l}$ at the last layer $l$ of the GGNN are used as input to the modified Graph Attention Network~\cite{Yang2020PR,Velickovic2018ICLR_GAT} to update the relevance of each object with respect to public or private nodes as the prior graph might not be optimal in its design at capturing this information~\cite{Yang2020PR}. This layer uses the low-rank bi-linear pooling~\cite{Kim2017ICLR} to further refine and fuse the features between an object node and a privacy node, and computes attention coefficients normalised in the interval $[0,1]$ between the pair of nodes. The normalised attention coefficient between the privacy node $j \in \mathcal{V}_p$ and the object node $i \in \mathcal{V}_o$ is 
\begin{equation}
    \alpha_{j,i} = \sigma\left(a\left(\tanh\left(W_i\mtrx{x}'_{l,i}\right) \odot \tanh\left(W_j\mtrx{x}'_{l,j}\right)\right)\right),
\end{equation}
where $\odot$ is the Hadamard product, $a(\cdot)$ is the attention mechanism~\cite{Velickovic2018ICLR_GAT}, $\sigma(\cdot)$ is the sigmoid function, and $W_i$ and $W_j$ are matrices with learnable parameters. Attention coefficients for object nodes not related to a privacy node are set to zero, avoiding the learning of the parameters~\cite{Yang2020PR}. For each privacy node, GRM outputs a vector, $\tilde{\mtrx{x}}_j$, that concatenates the refined and updated features of the node with the refined features of the object nodes, all weighted by the attention coefficients~\cite{Yang2020PR}:
\begin{equation}
    \tilde{\mtrx{x}}_j = \left[ \mtrx{x}'_j, \ldots, \alpha_{j,i} \mtrx{x}'_i, \ldots \right], \quad j \in \mathcal{V}_p, 1 \leq i \leq |\mathcal{V}_o|.
\end{equation}
The dimensionality of the feature vector $\tilde{\mtrx{x}}_j$ is $H=(1+ K_o) \times D'$, where $D'$ is the output dimensionality at the last GNN layer. 

Both GIP and GPA use GRM as a GNN and adapt the dimensionality of the feature vectors in input, output, and across the layers of the GNN depending on their design choice.

\subsection{Classification}

The two methods use an MLP-based classifier~\cite{Yang2020PR,Stoidis2022BigMM}. This classifier has two FC layers that transform each feature vector outputted by GRM from the dimensionality $H$ to the dimensionality of the input feature $D$, to a single logit. We remind the reader that $D=4,096$ when using deep features and excluding the 1-hot vector, and $D=2$ when using cardinality and the single node type flag as features. Before each FC layer there is a dropout layer that prunes some of the connections depending on the pre-defined probability. The first FC layer is followed by a ReLU activation function. Note that GPA~\cite{Stoidis2022BigMM} includes an additional FC layer that transforms the features of the privacy nodes outputted by GRM, $\tilde{\mtrx{x}}_j$ with $j \in \mathcal{V}_p$, into an intermediate feature vector of lower dimensionality (\mbox{$H \rightarrow K_o + 1$}) before the classifier. We refer to this layer as \textit{reshape layer}. 

The privacy node with the highest probability determines if the input image is private as 
\begin{equation}
    j^\ast = \argmax_{j \in \mathcal{V}_p} \varsigma\left(\mu(\tilde{\mtrx{x}}_j, W_{\mu})  \right),
\end{equation}
where $\varsigma(\cdot)$ is the softmax operation, \mbox{$\mu(\cdot)$: $\mathbb{R}^H \rightarrow \mathbb{R}^1$} is the function representing the classifier, and the learnable parameters $W_{\mu}$ are shared when applying the FC layers to the features of the two privacy nodes. 

The GNN and the classifier are trained in a supervised way by minimising Eq.~\ref{eq:celoss}. Both GIP and GPA use transfer learning to train their pipelines, where the CNNs are pre-trained to extract visual entities (see Fig.~\ref{fig:graphmodeldiagram}). GPA fine-tunes the parameters of the CNN pre-trained for scene recognition. GIP fine-tunes the CNNs to extract the visual features for objects and the private and public nodes.

%%%%%%%%%%%%%%%%%%%%%%%%%%%%%%%%%%%%%%%%%
\section{Experimental validation}
\label{sec:experiments}

\subsection{Datasets}

We evaluate the models on the Image Privacy Dataset (IPD)~\cite{Yang2020PR} and the PrivacyAlert dataset~\cite{Zhao2022ICWSM_PrivacyAlert}. Both datasets refer to images publicly available on Flickr. These images have a large variety of content, including sensitive content, semi-nude people, vehicle plates, documents, private events. 
Images were annotated with a binary label denoting if the content was deemed to be \textit{public} or \textit{private}\footnote{Verifying and updating the annotation of the datasets is beyond the scope of this work, requires careful instructions to the annotators to avoid biases, and needs to handle with the intrinsic subjectivity nature of privacy.}  
As the images are publicly available, their label is mostly public; however, there is a $\sim$33\% of the IPD images and a $\sim$25\% of the PrivacyAlert images labelled as private. These datasets have a high imbalance towards the public class. Note that IPD combines two other existing datasets, PicAlert~\cite{Zerr2012CIKM_PicAlert} and part of VISPR~\cite{Orekondy2017ICCV}, to increase the number of private images already limited in PicAlert.

IPD, or subsequent works using the dataset~\cite{Stoidis2022BigMM}, does not provide reproducible information on how data was previously split into training, validation, and testing sets. We therefore randomly split the data into training, validation, and testing sets using a K-Fold stratified strategy (with K=3)\footnote{The labelling of the three data splits will be made publicly available for reproducibility and enabling future comparison using the same sets.}. To train and evaluate the models, we only consider the first fold in this work. PrivacyAlert has a number of images that is about 6$\times$ less than IPD and provides the images already split into training, validation, and testing sets (see Tab.~\ref{tab:datasplits}). 

\begin{table}[t!]
    \centering
    \footnotesize
    \setlength\tabcolsep{6pt}
    \caption{Splits and number of images per class for the image privacy datasets considered for evaluation.
    \vspace{-10pt}
    }
    \begin{tabular}{l rrr rrr}
    \toprule
     & \multicolumn{3}{c}{\textbf{Image Privacy Dataset}} & \multicolumn{3}{c}{\textbf{PrivacyAlert}*} \\
    \cmidrule(lr){2-4}\cmidrule(lr){5-7}
    Data split & Private & Public & Overall & Private & Public & Overall \\
    \midrule
    Training   &  5,928 & 12,662 & 18,590 &   787 & 2,347 & 3,134 \\
    Validation &  2,979 &  6,081 &  9,060 &   466 & 1,397 & 1,863 \\
    Testing    &  2,304 &  4,608 &  6,912 &   450 & 1,346 & 1,796 \\
    All        & 11,211 & 23,351 & 34,562 & 1,703 & 5,090 & 6,793 \\
    \bottomrule \addlinespace[\belowrulesep]
    \multicolumn{7}{l}{\parbox{0.94\columnwidth}{\scriptsize{* PrivacyAlert provides links to images on Flickr whose license was under Public Domain~\cite{Zhao2022ICWSM_PrivacyAlert}. Note that 7 images are no longer available: 1 public image and 1 private image from the training set, 1 public image from the validation set, and 4 public images from the testing set~\cite{Xompero2024CVPRW_XAI4CV}.}}}
    \end{tabular}    
    \label{tab:datasplits}
    \vspace{-10pt}
\end{table}

\subsection{Performance measures}

We evaluate the models using per-class precision, per-class recall, per-class F1-score, overall precision, overall recall (or balanced accuracy), and accuracy. 
We first compute the number of true positives (TP), false positives (FP), and false negatives (FN) for each class $y$. For a given class, \textit{precision} is the number of images correctly classified as class $y$ over the total number of images predicted as the class $y$: \mbox{$P_y = TP_y/(TP_y + FP_y)$}. \textit{Recall} is the number of images correctly classified as class $y$ over the total number of images annotated as class $y$: $R_y = TP_y/(TP_y + FN_y)$. \textit{Accuracy} is the total number of images that are correctly classified as either public or private over the total number of images that are correctly classified, wrongly predicted ($FP_y$) and missed to be predicted ($FN_y$) with respect to the annotated class: 
\begin{equation}
    ACC = \frac{\sum_y{TP_y}}{\sum_y{TP_y + FP_y + FN_y}}. 
    \label{eq:accuracy}
\end{equation}

Balanced accuracy is the main performance measure to better assess the class imbalance of the dataset, and is the average between the recall of the two classes. Similarly, overall precision is the average between the precision of the two classes. Given the semantics of the task, we will give a particular emphasis to the recall for the private class in the discussion. We will also discuss the results using the performance measures as percentages. 

\subsection{Methods under comparison}

We evaluate and compare S2P, the graph-based models GIP~\cite{Yang2020PR} and GPA~\cite{Stoidis2022BigMM}, and two methods relating objects with privacy~\cite{Xompero2024CVPRW_XAI4CV,Tonge2016AAAI,Tonge2018AAAI,Baranouskaya2023ICIP}.
The first method (MLP) trains an MLP classifier and allows us to better understand the need of both deep features and graph processing in GIP and GPA in terms of size and classification performance. MLP has 3 hidden FC layers, each of 16-dimensionality hidden status, and a FC layer for the binary classification with cross-entropy loss. Each of these layers is followed by ReLU and dropout~\cite{Srivastava2014DropoutAS}. The first FC layer is also preceded by a dropout layer. The MLP classifier takes as input a feature vector that concatenates the cardinality information computed for each object category localised by an object detector.
The second method (GA-MLP) trains a graph-agnostic model and allows us to understand the relevance of the graph by replicating the steps of a Graph Convolutional Network~\cite{Kipf2017ICLR_GCN} but without the graph structure~\cite{Dwivedi2023JMLR,Xompero2024CVPRW_XAI4CV}. 
As the edges of the graph are not provided, the nodes are isolated and features are not propagated across nodes~\cite{Dwivedi2023JMLR,Xompero2024CVPRW_XAI4CV}. The method initialises the features of the nodes with the cardinality of each object category localised in the image, or with a zero value if any object of other categories is not localised in the image. After applying a dropout layer, node features are provided as input to three blocks\footnote{The previous work~\cite{Xompero2024CVPRW_XAI4CV} included, for each block, a batch normalisation layer that generates a source of randomness, making the model training not replicable despite fixing the seed. In our design, we remove the batch normalisation layer.} consisting of an FC layer, ReLU, and a dropout layer~\cite{Srivastava2014DropoutAS}. The FC layers transform the feature vector of each node into a 16-dimensional feature vector in output for each block. GA-MLP aggregates the refined features via global sum pooling and then applies an MLP-based classifier~\cite{Xompero2024CVPRW_XAI4CV}. The classifier has 2 hidden FC layers, each halving the dimensionality of the input features and followed by ReLU, and an FC layer for binary classification~\cite{Dwivedi2023JMLR}. 

We refer the reader to Appendix~\ref{app:mlp} and Appendix~\ref{app:gamlp} for an analysis of alternative design choices of MLP and GA-MLP, respectively. Performance varies across various design choices, such as different object features, use of feature normalisation, use of batch normalisation, use of a weighted cross-entropy loss, and different hyper-parameter values (number of hidden neurons and number of layers for MLP). As a best-performing model cannot be easily identifiable depending on the reference performance measure, we selected the variant with the fairer comparison with GPA.

\subsection{Implementation details}
\label{subsec:implementation}

To localise objects in the images, we use YOLOv8 by Ultralytics\footnote{YOLOv8x for object detection (COCO): \url{https://docs.ultralytics.com/models/yolov8/}}, pre-trained on the COCO dataset~\cite{Lin2018ECCV_COCO}, as object detector\footnote{Note the original GIP~\cite{Yang2020PR} uses Mask R-CNN with a maximum number of object instances set to 12 and the minimum confidence set to 0.7}. We set the maximum number of object instances localised for each image to 50, with a minimum confidence of 0.6 and non-maximum suppression threshold of 0.4. The input image is resized to a resolution of 416$\times$416 pixels before applying the object detector.
The outputs of the object detector are used consistently across all models. 

For GIP and GPA, we follow their settings. For GIP, we set the hidden channel to 4,098 and the output channel to 512. The hidden channel is given by the size of the input features (4,096) and the 1-hot vector denoting the node type (2). The input image is resized to a resolution of 448$\times$448 pixels and values are normalised with respect to the statistics computed on ImageNet before applying the CNNs to extract the deep features. For GPA, we set the hidden channel and the output channel to 2 (the number of output classes). As for GIP, the input image is resized to a resolution of 448$\times$448 pixels and values are normalised with respect to the statistics computed on ImageNet before applying the scene classifier to extract and transform the logits into features for the privacy nodes. Given the different data splits, we re-compute a weighted prior graph from the training set for each dataset. This prior graph includes both the bipartite sub-graph connecting the object nodes to the privacy nodes for GIP and the sub-graph connecting the object nodes with each other for GPA. For GIP, we mask the adjacency matrix to consider only the bipartite sub-graph. For GPA, we mask the adjacency matrix to consider only the sub-graph connecting object nodes and we binarise the non-zero values to 1, following the original design~\cite{Stoidis2022BigMM}. 

For MLP and GA-MLP, we use the cardinality information of the 80 object categories as input features, and we set the probability of the dropout layers to 0, thus avoiding any dropout effects on the performance even if being lower and not optimal. For S2P, we use a fixed ResNet-50, pre-trained on Places365~\cite{Zhou2017Vision_Places}, to predict the logits of 365 scenes. The parameters of the trainable FC layer are initialised with Xavier uniform distribution~\cite{Glorot2010ICAIS_XavierInit} and zero bias. 

For reproducibility and fairness, we implemented all models in a Python-based common framework using PyTorch, starting from the original implementation of GIP and GPA\footnote{GIP code: \url{https://github.com/guang-yanng/Image_Privacy} and GPA code: \url{https://github.com/smartcameras/GPA}}. All the models are trained and validated on a Linux-based machine using one NVIDIA GeForce GTX 1080 Ti GPU with 12 GB of RAM.

%%%%%%%%%%%%%%%%%%%%%%%%%%%%%%%%%%%%%%%%%%%
\subsection{Training details} 

We set the same training parameters across all the models, partially following the settings of benchmarking GNNs~\cite{Dwivedi2023JMLR}. As optimiser, we use Adam~\cite{Kingma2015ICLR} with an initial learning rate set to 0.001 and no weight decay. We half the learning rate when the balanced accuracy has stopped improving for at least 10 epochs (patience). We set the maximum number of epochs to 1,000, but the training stops early if the learning rate is reduced to a value lower than 0.00001 or the training time lasts longer than 12~h. For both the training and validation splits, we set the batch size to 100. Given the larger size of GIP, we increase the training time to a maximum of 72~h and we reduce the batch size to 2. We save the model at the epoch with the highest balanced accuracy in the validation split, and this model is used for the evaluation on the testing split. For reproducibility of models and experiments, we set the seed to an arbitrary value of 789. Note that we do not analyse variations in the performance due to different seeds, which is beyond the scope of this paper.

%%%%%%%%%%%%%%%%%%%%%%%%%%%%%%%%%%%%%%%%%%%
\begin{table*}[t!]
    \centering
    \footnotesize
    \setlength\tabcolsep{7pt}
    \caption{Relevance of various design choices in GIP when re-trained and evaluated on the testing set of PrivacyAlert and IPD.
    \vspace{-10pt}
    }
    \begin{tabular}{c ccc c ccc ccc cc}
    \toprule
    & \multicolumn{3}{c}{\textbf{Graph node features init. (deep) }} & & \multicolumn{3}{c}{\textbf{PrivacyAlert}} & \multicolumn{3}{c}{\textbf{IPD}} & \multicolumn{2}{c}{\textbf{\# of trainable parameters}}\\
    \cmidrule(lr){2-4}\cmidrule(lr){6-8}\cmidrule(lr){9-11}\cmidrule(lr){12-13}
    & Object nodes & Privacy nodes & Type & WD & R (Priv.) & BA & ACC & R (Priv.) & BA & ACC & Optimised & Total \\
    \midrule
    * & -- & Fine-tuning & -- & -- & -- & -- & -- & 75.40 & 74.40 & 74.33 & 50,901,058 & 50,901,058\\
    * & Fine-tuning & Fine-tuning & \bbox & \bbox & -- & -- & -- & 75.10 & 77.30 & 77.09 & 514,600,884 & 514,600,884 \\
    $^\dagger$ & Fine-tuning & Fine-tuning & \bbox & -- & 45.50 & 67.85 & 79.35 & 59.00 & 74.55 & 81.72 & 514,600,884 & 514,600,884 \\
    \midrule
    $^\diamond$ & Fine-tuning & Fine-tuning & \bbox & \wbox & 47.33 & 68.06 & 78.40 & 99.57 & 50.47 & 34.10 & 514,600,884 & 514,600,884 \\    
    & Fixed & Fixed & \bbox & \wbox & 100.00 & 50.26 & 25.45 & 100.00  & 50.00 & 33.33 & 337,840,180 & 514,600,884 \\
    \rowcolor{mylightgray}& Fixed & Zeros & \bbox & \wbox & 75.78 & 70.88 & 68.43 & 63.06 & 63.42 & 63.54 & 329,439,282 & 463,699,826\\
    $^\diamond$ & Fixed & Zeros & \wbox & \wbox & 73.11 & 67.68 & 64.98 & 35.46 & 59.65 & 67.71 & 329,287,682 & 463,548,226 \\
    \midrule
    $^\diamond$ & Fine-tuning & Fine-tuning & \bbox & \bbox & 66.22 & 70.96 & 73.33 & 56.90 & 68.71 & 72.64 & 514,600,884 & 514,600,884 \\ 
    & Fixed & Fixed & \bbox & \bbox & 0.22 & 49.63 & 74.28 & 38.37 & 56.40 & 62.41 & 337,840,180 & 514,600,884 \\
    \bottomrule \addlinespace[\belowrulesep]
    \multicolumn{13}{l}{\parbox{0.9\linewidth}{\scriptsize{* Results are taken from the GIP paper~\cite{Yang2020PR}. The first row is the only ResNet based model. Note that PrivacyAlert was not yet available as dataset. $^\dagger$ Results are taken from the GPA paper~\cite{Stoidis2022BigMM}. $^\diamond$ Results for IPD are provided by the best model that is obtained at the first epoch. After the first epoch, the model degenerates to predicting only the private class. Highlighted in gray the best variant. KEY - init.:~initialisation, WD:~use of weight decay in the optimiser; Priv.:~private class, R:~recall, BA:~Balanced accuracy, ACC:~accuracy, --: unknown; \wbox:~not used, \bbox:~used.}}}
    \end{tabular}
    \label{tab:analysisgip}
    \vspace{-10pt}
\end{table*}

\subsection{Models size}

The models require a different number of trainable parameters to be optimised for image privacy classification. We provide and compare the number of trainable parameters and the number of optimised parameters. Note that the methods at inference can include additional components, such as the CNN for object detection, but we do not count their parameters in the total number when these parameters are not optimised for the downstream task. 

\textit{GIP} has 514,600,884 trainable parameters. All these parameters are optimised for the downstream task when also fine-tuning the two CNNs that account for 185,161,602 parameters: 50,901,058 for the ResNet-101 pre-trained on ImageNet and 134,260,544 for VGG-16. Specifically, the FC layer, which alignes the dimensionality of the ResNet feature vector with the one of the VGG-16 feature vector, accounts for 8,400,898 parameters; GRM has 159,561,777 parameters; and the final classifier has 169,877,505 parameters. 
\textit{GPA} optimises 14,175 trainable parameters: 73 for GRM, 167 for the classifier, 13,203 for the reshaping layer, and 732 for the scene-to-privacy layer. GPA also includes the pre-trained and fixed scene classifier and the fixed object detector, whose parameters are not counted in the total. The choice of features and their dimensionality significantly reduces the number of parameters in GPA compared to those in GIP. The scene classifier is based on a ResNet-50 model pre-trained on Places365 for scene recognition and has 24,255,917 trainable parameters. 
The scene classifier and the scene-to-privacy layer also form \textit{S2P}, resulting in 24,256,649 trainable parameters whose only 732 are optimised for the downstream task. This makes S2P the model with the least number of parameters to optimise. 
\textit{MLP} optimises 1,906 trainable parameters: the first layer has 1,328 parameters (82 nodes $\times$ 16-dimensional hidden status $+$ bias), the second and third layer have 272 parameters each; and the last  layer has 34 parameters (including bias). 
\textit{GA-MLP} has 1,250 trainable parameters, all optimised for the downstream task: the 3 FC layers has 32, 272, and 272 parameters, respectively; 164 parameters for each batch normalisation layer following the FC layers (82 nodes $\times$ 2 affine parameters\footnote{We use the BatchNorm1D function in PyTorch where the parameters $\gamma$ and $\beta$ are optimised over graph nodes in our case across  batches: \url{https://pytorch.org/docs/stable/generated/torch.nn.BatchNorm1d.html}.}); and 182 parameters for the final MLP classifier.

\subsection{Relative impacts on image privacy} 
\label{subsec:relativeimpacts}

Both GIP and GPA are trained end-to-end, optimising the parameters of both the GNN and CNNs. 
To better understand the relative contribution of their individual components, we analyse the classification performance of both models when re-trained with different training strategies and different design choices.

Table~\ref{tab:analysisgip} analyses the classification performance of GIP when using four different training strategies and design choices. We compare the model trained end-to-end, including the fine-tuning of the CNNs, with the model re-trained end-to-end but keeping the parameters of the two CNNs fixed. In the latter, only the FC layer that aligns the dimensionality of the deep features is optimised.
To remove the influence of the CNN extracting image-level features, the other two alternatives
replace the deep features of the privacy nodes with a vector of zero values. As we hypothesise that the 1-hot vector   is not relevant for the refinement of the features in the GNN nor for the final performance, the two models differ from each other simply by the inclusion of this 1-hot vector. We also reported the total number of trainable parameters and the number of trainable parameters that are optimised for the task to show the relationship between model size and classification performance.

\begin{table*}[t!]
    \centering
    \footnotesize
    \setlength\tabcolsep{6.1pt}
    \caption{Relevance of various design choices in GPA when re-trained and evaluated on the testing set of PrivacyAlert and IPD. Note the best performance achieved by pre-training the scene-to-privacy layer, using a bipartite prior graph as for GIP~\cite{Yang2020PR}, and removing the node type flag from the node feature vectors (sixth row). 
    \vspace{-10pt}
    }
    \begin{tabular}{lc cccc rrr rrr cccr}
      \toprule
      \multicolumn{2}{c}{\textbf{Prior graph}} & \multicolumn{4}{c}{\textbf{Design choices}} & \multicolumn{3}{c}{\textbf{PrivacyAlert}} & \multicolumn{3}{c}{\textbf{IPD}} & \multicolumn{4}{c}{\textbf{Number of parameters (opt.)}} \\
      \cmidrule(lr){1-2}\cmidrule(lr){3-6}\cmidrule(lr){7-9}\cmidrule(lr){10-12}\cmidrule(lr){13-16}
      Type & W & Privacy feat. & S2P & Resh. & Flag & R (Priv.) & BA & ACC & R (Priv.) & BA & ACC  & GNN & Class. & Resh. & Total \\
      \midrule
      co-occ.* $^\dagger$ & \wbox & Deep   & Training   & \bbox & \bbox &  51.40 & 72.45 & 83.28 & 73.60 & 79.80 & 80.96 & 73 & 167 & 13,203 & 14,175 \\
      co-occ.*            & \wbox & Deep   & Training   & \bbox & \bbox & 51.56 & 72.51 & 82.96 & 71.83 & 80.71 & 83.67 & 73 & 167 & 13,203 & 14,175 \\
      co-occ.*            & \wbox & Deep   & Pretrained$^\diamond$ & \bbox & \bbox & 0.00  & 50.00 & 74.94 & 100.00 & 50.00 & 33.33 & 73 & 167 & 13,203 & 14,175 \\      
      co-occ.             & \wbox & Deep   & Pretrained$^\diamond$ & \bbox & \bbox & 0.00  & 50.00 & 74.94 & 100.00 & 50.00 & 33.33  & 73 & 167 & 13,203 & 14,175 \\ 
      bipartite           & \bbox & Deep   & Pretrained$^\diamond$ & \bbox & \bbox & 61.56 & 75.21 & 82.02 & 73.96 & 81.08 & 83.45 & 73 & 167 & 13,203 & 14,175 \\ 
      \rowcolor{mylightgray}bipartite           & \bbox & Deep   & Pretrained$^\diamond$ & \bbox & \wbox & 63.11 & 75.43 & 81.57 & 74.18 & 81.08 & 83.38 & 32 & 167 & 13,203 & 14,134 \\ 
      bipartite           & \bbox & Deep   & Pretrained$^\diamond$ & \wbox & \wbox & 27.78 & 63.15 & 80.79 & 100.00 & 50.00 & 33.33 & 32 & 329 & -- & 1,093 \\
      bipartite           & \bbox & Deep   & Training   & \wbox & \wbox & 57.33 & 73.69 & 81.85 & 100.00 & 50.00 & 33.33 & 32 & 329 & -- & 1,093\\
      bipartite           & \bbox & random & --         & \wbox & \wbox &  0.00 & 50.00 & 74.94 & 100.00 & 50.00 & 33.33 & 32 & 329 & -- & 361\\
      bipartite           & \bbox & zero   & --         & \wbox & \wbox &  0.00 & 50.00 & 74.94 & 100.00 & 50.00 & 33.33 & 32 & 329 & -- & 361 \\
      \bottomrule \addlinespace[\belowrulesep]
      \multicolumn{16}{l}{\parbox{0.97\linewidth}{\scriptsize{
      *GPA has a wrong implementation of the adjacency matrix not representing the idea to connect object nodes to each other as co-occurrence in the images. $^\dagger$ Results are taken from the corresponding paper~\cite{Stoidis2022BigMM}, which use a different number of images for the dataset. $^\diamond$ The scene-to-privacy (S2P) layer was trained offline before training the GPA model. S2P has always 732 parameters to optimise. The parameters of the scene classifier are pre-trained and fixed, and not counted in this table. Highlighted in gray the best variant. 
      KEY -- opt.:~parameters that are optimised during training (other trainable parameters are fixed); W:~weighted; feat.:~features; S2P:~scene-to-privacy layer; Resh.:~reshape layer; Flag:~binary variable denoting node type in the feature vector; R:~recall; BA:~balanced accuracy; ACC:~accuracy; GNN:~Graph Neural Network; Class.:~classifier; co-occ.:~co-occurrence; Deep:~deep features,  \wbox:~not used, \bbox:~used; --:~not applicable.}}}
    \end{tabular}
    \label{tab:analysisgpa}
    \vspace{-10pt}
\end{table*}

We compare the results of the first version of GIP with the results reported in the GIP paper~\cite{Yang2020PR} on the IPD dataset, and with the results reported in the GPA paper~\cite{Stoidis2022BigMM}, which re-trained and evaluated the model on both PrivacyAlert and IPD (see first 4 rows of Table~\ref{tab:analysisgip}). Note that these models are trained on different data splits of IPD due to missing information about the splits and training. On PrivacyAlert, our re-trained GIP achieves comparable classification performance to GIP~\cite{Stoidis2022BigMM}, with an improvement of about 2~percentage points (pp) in the recall of the private class, a small improvement of 0.21~pp in balanced accuracy, and a decrease of about 1~pp in accuracy. On IPD, the original GIP~\cite{Yang2020PR} achieved an improvement of 3~pp on balanced accuracy and accuracy compared to only using the ResNet-101 classifier fine-tuned for image privacy classification, mostly caused by correctly predicting more images as public rather than private~\cite{Yang2020PR}. However, if this improvement was caused by the graph processing or the end-to-end training of the model is unclear. The GIP model re-trained by GPA's authors~\cite{Stoidis2022BigMM} has already significant differences with respect to the original model~\cite{Yang2020PR}, with the recall on the private class decreasing of about 16~pp, the balanced accuracy decreasing of about 3~pp, and the accuracy improving of almost 5~pp. This shows that the re-trained model was correctly predicting more images labelled as public than those labelled as private. However, our re-trained model tends to degenerate to predict almost all images as private, as the recall on the private class is almost 100\%, but the balanced accuracy is almost 50\%. Results are obtained from the model saved at the epoch with the best balanced accuracy on the validation set, which occurs at the beginning of the training, and afterwards, the training degenerates to predict only the private class. 
When fixing the parameters of both CNNs, the optimisation of the model parameters degenerates to predict only the private class during the end-to-end training, as shown by the results on the testing set of both datasets. On PrivacyAlert, the balanced accuracy is slightly higher than 50\%, denoting that there are images predicted as public and some of them are correctly predicted as public. Keeping the same parameter settings for the training but replacing the features of the privacy nodes with a vector of zero values results in classification performance lower than the original GIP, with values in the interval [63-64]\%, on IPD. This shows that the model is highly affected by the presence of the deep features in the privacy nodes and only using the deep features from the objects might be sufficient. However, removing the node type from the feature vector decreases the recall on the private class to only 35.46\% but improves the correct prediction of images labelled as public, increasing the accuracy of 4~pp from 63.54\%. On the contrary, on PrivacyAlert, our re-trained GIP with zero features for the privacy nodes achieved the highest recall on the private class (75.78\%) and the highest balanced accuracy (70.88\%), but the accuracy decreases of about 11~pp denoting a higher prediction of false positive in the public class (more images predicted as private). Removing the node type decreases all the three performance measures as less images are correctly predicted as private.

Table~\ref{tab:analysisgpa} analyses the classification performance of GPA with different training strategies and design choices, resulting in nine different alternatives and eighteen models across the two datasets. We consider the use of the reshape layer, the use of the flag differentiating the node types, the chosen training strategy for the scene-to-privacy layer, the replacing of the feature of the privacy nodes with a zero or random value, and the type of prior graph (unweighted co-occurrence or weighted bipartite). 

Because of our re-implementation of the framework, we-retrained GPA following the original settings~\cite{Stoidis2022BigMM} and compare their results (see first two rows of Table~\ref{tab:analysisgpa}). Both models achieves similar results on PrivacyAlert, whereas our model achieves a higher balanced accuracy and accuracy on IPD despite a decrease of almost 2~pp in the recall of the private class. However, by pre-training and fixing the parameters of the scene-to-privacy layer, the end-to-end training cannot optimise the parameters, degenerating to predict only the public class. By correcting the implementation of the adjacency matrix wrongly initialised in the original GPA, the training of the model still degenerates to predict only the public class and this is caused by the lack of connection between the object nodes and the privacy nodes. Therefore, we re-use the weighted bipartite prior graph of GIP for the other alternatives. In this case, GPA with a pre-trained and fixed scene-to-privacy layer can achieve classification performance that is similar or higher than the original GPA. This is especially the case for the recall of the private class on the PrivacyAlert dataset, improving by 10-12~pp. The use of the flag has minimal effect with an improvement of 2~pp on the recall of the private class and a decrease of about 0.5~pp on the accuracy on PrivacyAlert, whereas the results are almost the same for the three performance measures on IPD. Removing the reshape layer from the architecture has a major impact on the classification performance as the model degenerates to predict only the private class on IPD and the recall on the private class decreases to only 27.78\% on PrivacyAlert. The reshape layer also counts for most of the optimised model parameters (13,203) and by removing it, the number of optimised parameters for the task reduces from 14,175 to 1,093. However, allowing the model to train the scene-to-privacy layer while removing the reshape layer helps the model achieve higher recall on the private class and comparable balanced accuracy and accuracy to the original GPA on PrivacyAlert. Furthermore, replacing the feature of the privacy nodes with either a zero or random value makes the model degenerate to only predict the public class in PrivacyAlert and the private class in IPD. This shows how the graph processing alone is not sufficient (only 361 parameters optimised), and the use of the CNN plays a central role in the model.

%==================================================
\pgfplotstableread{freq_obj_co-occ_cls_ipd.txt}\freqoccipd
\pgfplotstableread{freq_obj_co-occ_cls_privacyalert.txt}\freqoccprivacyalert
\begin{figure}[t!]
    \centering
    \begin{tikzpicture}
    \begin{axis}[
    axis x line*=bottom,
    axis y line*=left,
    enlarge x limits=false,
    ybar stacked,
    width=0.62\columnwidth,
	bar width=10pt,
    nodes near coords,
    nodes near coords style={font=\tiny, /pgf/number format/.cd,fixed zerofill,precision=1},
    xmin=-1,xmax=8.5,
    xtick={0,1,2,3,4,5,6,7,8},
    height=0.6\columnwidth,
    ymin=0,  ymax=100,
    ylabel={Frequency (\%)},
    xlabel={\# co-occurring objects},
    label style={font=\footnotesize},
    tick label style={font=\footnotesize},
    title={IPD},
    title style={font=\footnotesize},
    ]
    \addplot+[ybar, white, fill=dc2, draw opacity=0.5] table[x=n-objs,y=train-priv]{\freqoccipd};
    \addplot+[ybar, white, fill=dc1, draw opacity=0.5] table[x=n-objs,y=train-publ]{\freqoccipd};
    \end{axis}
    \end{tikzpicture}
    \begin{tikzpicture}
    \begin{axis}[
    axis x line*=bottom,
    axis y line=none,
    enlarge x limits=false,
    ybar stacked,
    width=0.58\columnwidth,
	bar width=10pt,
    nodes near coords,
    nodes near coords style={font=\tiny, /pgf/number format/.cd,fixed zerofill,precision=1},
    xmin=-0.5,xmax=7.5,
    xtick={0,1,2,3,4,5,6,7},
    height=0.6\columnwidth,
    ymin=0,  ymax=100,
    yticklabels={},
    xlabel={\# co-occurring objects},
    label style={font=\footnotesize},
    tick label style={font=\footnotesize},
    title={PrivacyAlert},
    title style={font=\footnotesize},
    ]
    \addplot+[ybar, white, fill=dc2, draw opacity=0.5] table[x=n-objs,y=train-publ]{\freqoccprivacyalert};
    \addplot+[ybar, white, fill=dc1, draw opacity=0.5] table[x=n-objs,y=train-priv]{\freqoccprivacyalert};
    \end{axis}
    \end{tikzpicture}
   \caption{Frequency of public images (\protect\tikz \protect\draw[dc2,fill=dc2] (0,0) rectangle (1.ex,1.ex);) and private images (\protect\tikz \protect\draw[dc1,fill=dc1] (0,0) rectangle (1.ex,1.ex);) based on the subset of images with only X localised object types (\# of co-occurrent objects). The number of images differ for each stacked vertical bar.
   }
    \label{fig:objimgfreq}
    \vspace{-10pt}
\end{figure}
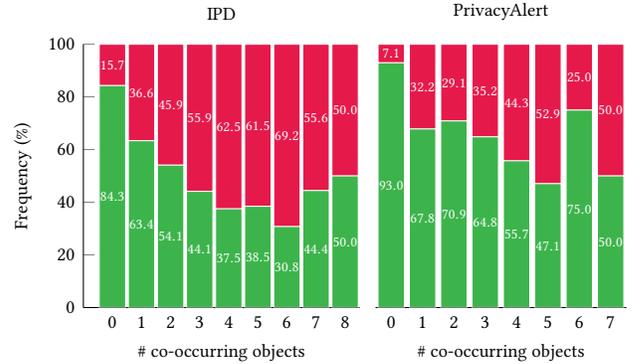

%%%%%%%%%%%%%%%%%%%%%%%%%%%%%%%%%%%%%%
\subsection{Limitations of object-only detection}
\label{app:objdet}

Both GIP~\cite{Yang2020PR} and GPA~\cite{Stoidis2022BigMM} are based on a pre-trained object detector~\cite{Redmon2018YOLOv3,He2017ICCV_MaskRCNN}. However, depending on the image content and the confidence threshold of the detector, objects can be wrongly localised or completely missed, thus resulting in a lower number or a total absence. Because of this, we analyse the number of co-occurring objects in the training sets of IPD~\cite{Yang2020PR} and PrivacyAlert~\cite{Zhao2022ICWSM_PrivacyAlert}, and their impact when designing models relating objects and privacy. Note we use the term co-occurrent objects referring to two different objects categories and not to two instances of the same category.

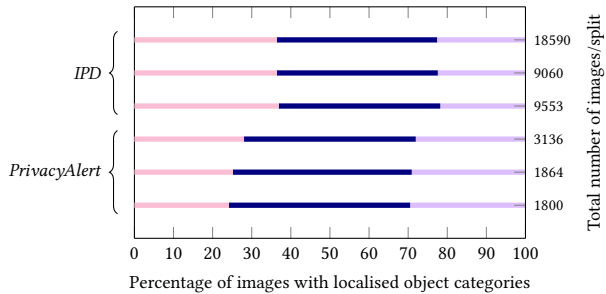
\begin{figure}[t!]
    \centering
    \begin{tikzpicture}
    \begin{axis}[
        width=.8\columnwidth,
        xbar stacked,
        bar width=2pt,
        axis y line*=left,
        ymin=0,ymax=7,
        ytick={1,2,3,4,5,6},
        yticklabels={},
        yticklabel style={name=T\ticknum},% names every xtick label node T0,T1, ...
        tick label style={font=\footnotesize},
        % y=3mm,
        height=0.55\columnwidth,
        y dir=reverse,
        xmin=0,  xmax=100,
        xtick={0,10,20,30,40,50,60,70,80,90,100},
        xlabel={Percentage of images with localised object categories},
        label style={font=\footnotesize},
    ]
    \addplot+[xbar,black,fill=dc10, draw opacity=0] table[x=survival,y=sequence]{\trackperf};
    \addplot+[xbar,black,fill=dc19, draw opacity=0] table[x=short,y=sequence]{\trackperf};
    \addplot+[xbar,black,fill=dc12, draw opacity=0] table[x=geometric,y=sequence]{\trackperf};
    \end{axis}
    \begin{axis}[
        width=.8\columnwidth, 
        xbar stacked,
    	bar width=2pt,
        axis y line*=right,
        ymin=0,ymax=7,
        ytick=data,
        ytick={1,2,3,4,5,6},
        yticklabels={18590,9060,9553,3136,1864,1800},
        tick label style={font=\scriptsize},
        ylabel={Total number of images/split},
        ylabel near ticks,
        height=0.55\columnwidth,
        y dir=reverse,
        xmin=0,
        xmax=100,
        xtick={0,10,20,30,40,50,60,70,80,90,100},
        hide x axis,
        label style={font=\footnotesize}
        ]
    \end{axis}
    \begin{scope}[decoration=brace]
        \pgfdecorationsegmentamplitude=3pt
        \draw[decorate] (T2.south west) -- (T0.north west) node[midway,left=\pgfdecorationsegmentamplitude] {\footnotesize{\textit{IPD}}};
         \draw[decorate] (T5.south west) -- (T3.north west) node[midway,left=\pgfdecorationsegmentamplitude] {\footnotesize{\textit{PrivacyAlert}}};
    \end{scope}
    \end{tikzpicture}
    \vspace{-3pt}
   \caption{Percentage of images with no localised objects  (\protect\tikz \protect\draw[dc10,fill=dc10] (0,0) rectangle (1.ex,1.ex);), with only one object type (\protect\tikz \protect\draw[dc19,fill=dc19] (0,0) rectangle (1.ex,1.ex);), with more than one object type (\protect\tikz \protect\draw[dc12,fill=dc12] (0,0) rectangle (1.ex,1.ex);). Each object type can have any number of localised instances. From top to bottom: training, validation, and testing splits for each dataset. 
   }
    \label{fig:objtimgstats}
    \vspace{-10pt}
\end{figure}

Fig.~\ref{fig:objimgfreq} shows the frequency of private images when varying the number of co-occurrent objects for each image on PrivacyAlert and IPD (different distributions). There are no images with more than 7 and 8 objects for PrivacyAlert and IPD, respectively. The presence of one or more objects is not uniquely mapped to either public or private images. Lack of localised objects is occurring mostly in public images but also in some private images.
Fig.~\ref{fig:objtimgstats} shows the percentages of images with no objects, with only one object, and with two or more objects for each data split of the two datasets. Distributions are approximately the same across the splits of each datasets. 
These statistics show that a large number of images available in these  datasets do not contain objects to localise (or the detector can miss to detect existing objects) or only one object can be localised (about 70\% of PrivacyAlert and about 80\% of IPD). Therefore, using only objects detected in images as nodes or features for a graph-based model~\cite{Stoidis2022BigMM,Yang2020PR} makes the task of recognising private images even harder.

%%%%%%%%%%%%%%%%%%
\pgfplotstableread[row sep=\\,col sep=&]{
    sequence & survival & short & geometric \\
    1 & 36.44 & 40.95 &      22.61 \\
    2 &    36.45 & 41.13 &   22.43 \\
    3 &    36.91 & 41.30 &   21.79 \\
    4 & 28.03 & 43.91 &   28.06 \\
    5 & 25.16 & 45.76&        29.08 \\
    6 & 24.17 & 46.33&        29.50 \\
    }\trackperf
%%%%%%%%%%%%%%%%%%

\subsection{Comparative analysis}

We perform an in-depth comparative analysis of GIP, GPA, MLP, GA-MLP, and S2P both in terms of classification performance and number of optimised parameters. We also consider an MLP variant that resizes and reshapes the image as input to the MLP instead of using object features (MLP-I, see details in Appendix~\ref{app:mlpimage}).
As references for the discussion, we include results for a baseline predicting all the images as public, a second baseline predicting all the images as private, a third baseline using a pseudo-random generator (Random) to sample the predictions from a uniform distribution, and two person-centric classification strategies~\cite{Xompero2024CVPRW_XAI4CV} (PCS2 and PCS3 in Table~\ref{tab:relatedworks}). PCS2 predicts an image as private based on the presence of the person category as localised by an object detector. PCS3 predicts an image as private based on the presence of the person category and its cardinality constrained to a maximum value of 2. 

\begin{table*}[t!]
    \centering
    \footnotesize
    \setlength\tabcolsep{6.5pt}
    \caption{Results and comparisons for image privacy classification on the testing sets of the two image privacy datasets. Note the comparable performance between GPA and S2P, achieving the highest balanced accuracy in both datasets.
    \vspace{-10pt}
    }
    \begin{tabular}{ll ccc c c  ccc ccc ccc}
        \toprule
        \multicolumn{1}{l}{\textbf{Dataset}} & \multicolumn{1}{l}{\textbf{Method}} & \multicolumn{3}{c}{\textbf{Object features}} & \multicolumn{1}{c}{\textbf{Scene}} & \multicolumn{1}{c}{\textbf{Image}} & \multicolumn{3}{c}{Private} & \multicolumn{3}{c}{Public} & \multicolumn{3}{c}{Overall}  \\
        \cmidrule(lr){3-5}\cmidrule(lr){8-10}\cmidrule(lr){11-13}\cmidrule(lr){14-16}
        & & Card & Conf & Deep & Logits & Deep & P & R & F1 & P & R & F1 & P & BA & ACC \\
        \midrule
        \multirow{11}{*}{IPD~\cite{Yang2020PR}}
        & All private & - & - & - & - & - & 33.33  & 100.00 & 50.00  & 0.00   & 0.00   & 0.00   & 16.67  & 50.00 & 33.33 \\
        & All public & - & - & - & - & - & 0.00   & 0.00   & 0.00   & 66.67  & 100.00 & 80.00  & 33.33  & 50.00 & 66.67 \\ 
        & Random & - & - & - & - & - & 33.68  & 50.61  & 40.44  & 67.01  & 50.17  & 57.38  & 50.35  & 50.39 & 50.32 \\ 
        \cmidrule(lr){2-16}
        & PCS2~\cite{Xompero2024CVPRW_XAI4CV} & \bbox & \bbox & \wbox & \wbox & \wbox & 17.14 & 30.12 & 21.85 & 43.77 & 27.19 & 33.54 & 30.45 & 28.66 & 28.17 \\
        & PCS3~\cite{Xompero2024CVPRW_XAI4CV} & \bbox & \bbox & \wbox & \wbox & \wbox & 23.04 & 44.49 & 30.36 & 48.07 & 25.69 & 33.49 & 35.56 & 35.09 & 31.96 \\
        \cmidrule(lr){2-16}
        & MLP-I & \wbox & \wbox & \wbox & \wbox & \wbox & 46.98 & 44.92 &  45.93 & 73.05 & 74.65 & 73.84 & 60.02 & 59.79 & 64.74 \\
        & MLP & \bbox & \wbox & \wbox & \wbox & \wbox & 59.29 & 75.09 & 66.26 & 85.63 & 74.22 & 79.52 & 72.46 & 74.65 & 74.51 \\
        & GA-MLP & \bbox & \wbox & \wbox & \wbox & \wbox & 59.04 & 75.65 & 66.32 & 85.83 & 73.76 & 79.34 & 72.44 & 74.71 & 74.39  \\
        & GIP & \wbox & \wbox & \bbox & \wbox & \bbox & 46.54 & 63.06 & 53.56 & 77.55 & 63.78 & 69.99 & 62.04 & 63.42 & 63.54 \\
        & GPA & \bbox & \wbox & \wbox & \bbox & \wbox & 75.52 & 74.18 & 74.84 & 87.20 & 87.98 & 87.59 & 81.36 & 81.08 & 83.38\\
        & S2P & \wbox & \wbox & \wbox & \bbox & \wbox & 75.83 & 72.44 & 74.10 & 86.52 & 88.45 & 87.48 & 81.18 & 80.45 & 83.12\\
        \midrule
        \multirow{11}{*}{PrivacyAlert~\cite{Zhao2022ICWSM_PrivacyAlert}}
        & All private & - & - & - & - & - & 25.00 & 100.00 & 40.00 & 0.00 & 0.00 & 0.00  & 12.50 & 50.00 & 25.00 \\ % & 20.00 & 10.00 \\
        & All public & - & - & - & - & - & 0.00 & 0.00 & 0.00 & 75.00 & 100.00 & 85.71  & 37.50 & 50.00 & 75.00 \\ % & 42.86 & 64.29 \\
        & Random & - & - & - & - & - & 74.27 & 50.67 & 60.24 & 24.23 & 47.33 & 32.05 &  49.25 & 49.00 & 49.83 \\ % & 46.15 & 53.19 \\
        \cmidrule(lr){2-16}
        & PCS2~\cite{Xompero2024CVPRW_XAI4CV} & \bbox & \bbox & \wbox & \wbox & \wbox & 39.61 & 94.00 & 55.73 & 96.29 & 52.08 & 67.60 & 67.95 & 73.04 & 62.58  \\
        & PCS3~\cite{Xompero2024CVPRW_XAI4CV}  & \bbox & \bbox & \wbox & \wbox & \wbox & 42.66 & 89.11 & 57.70 & 94.28 & 59.96 & 73.30 & 68.47 & 74.53 & 67.26 \\
        \cmidrule(lr){2-16}
        & MLP-I & \wbox & \wbox & \wbox & \wbox & \wbox & 32.11 & 31.11 & 31.60 & 77.21 & 78.01 & 77.61 & 54.66 & 54.56 & 66.26 \\
        & MLP & \bbox & \wbox & \wbox & \wbox & \wbox & 50.56 & 70.67 & 58.94 & 88.69 & 76.89 & 82.37 & 69.62 & 73.78 & 75.33 \\
        & GA-MLP & \bbox & \wbox & \wbox & \wbox & \wbox & 49.76 & 68.89 & 57.78 & 88.06 & 76.75 & 82.02 & 68.91 &  72.82 & 74.78 \\       
        & GIP & \wbox & \wbox & \bbox & \wbox & \bbox  & 42.68 & 75.78 & 54.60 & 89.07 & 65.97 & 75.80 & 65.87 & 70.88 & 68.43 \\
        & GPA & \bbox & \wbox & \wbox & \bbox & \wbox & 63.25 & 63.11 & 63.18 & 87.68 & 87.74 & 87.71 & 75.46 & 75.43 & 81.57\\
        & S2P & \wbox & \wbox & \wbox & \bbox & \wbox & 63.11 & 63.11 &  63.11 & 87.67 &  87.67 & 87.67 & 75.39 & 75.39 & 81.51 \\
      \bottomrule \addlinespace[\belowrulesep]
      \multicolumn{16}{l}{\parbox{0.975\linewidth}{\scriptsize{KEY -- Card:~cardinality, Conf:~confidence, Deep:~deep features, P:~precision, R:~recall, ACC:~accuracy, BA:~Balanced accuracy (corresponds to overall recall); PCSX:~person-centric strategy X with X being 2 or 3 (see also Table~\ref{tab:relatedworks}); MLP-I:~multi-layer perceptron that takes a resized and reshaped image as input; GA:~graph-agnostic; GIP:~Graph-based Image Privacy; GPA:~Graph Privacy Advisor; S2P:~Scene-to-Privacy.}}}    
    \end{tabular}
    \label{tab:compartiveanalysis}
    \vspace{-7pt}
\end{table*}

Table~\ref{tab:compartiveanalysis} and Fig.~\ref{fig:recallvssacc} compare the image privacy classification results of the methods and reference baselines on IPD and PrivacyAlert. For both datasets, our re-trained GPA and S2P achieve the highest classification results in terms of overall accuracy (>83\% on IPD, and >80\% in PrivacyAlert) and balanced accuracy (>80\% on IPD and >75\% in PrivacyAlert). The performance is mostly driven by the higher recall on the public class and the higher precision on the private class for IPD, while the recall on the private class is lower than MLP and GA-MLP of 1-2~pp. A similar behaviour is observed in PrivacyAlert, but the recall on the private class is lower than other methods, especially GIP of about 12~pp. The similar performance between GPA and S2P indicates that using transfer learning from the pre-trained scene classifier is sufficient for achieving such a performance and the impact of graph processing on the results is minimal. Both MLP and GA-MLP achieve lower classification performance than GPA and S2P, but similar to each other when using only object cardinality as input feature. Using the resized and reshaped image as input to an MLP classifier achieve the lowest classification performance among the learning based methods: 59.79\% and 54.56\% balanced accuracy on IPD and PrivacyAlert, respectively. The person-centric strategies~\cite{Xompero2024CVPRW_XAI4CV} achieve the lowest performance on IPD: 28.66\% for PCS2 and 35.09\% for PCS3. On PrivacyAlert, these two strategies achieve the highest recall on the private class (94\% and 89.11\%) but at the cost of the lowest recall on the public class (52.08\% and 59.96\%), showing that the strategies are too restrictive for the classification task. 

Fig.~\ref{fig:recallvssacc} compares the methods by relating recall on the private class and balanced accuracy. On PrivacyAlert, all models except MLP-I lie on the close surrounding of an imaginary line perpendicular to the diagonal that equals recall and balanced accuracy. This indicates that these models have similar classification performance with higher balanced accuracy when above the diagonal line or higher recall when below the diagonal line. On IPD, GA-MLP and MLP are close to each other and both lie almost on the top-right part of the diagonal, showing high classification performance with a balance between recall and balanced accuracy. S2P and GPA have a similar recall but with higher balanced accuracy than MLP and GA-MLP. 

Fig.~\ref{fig:sizevsacc} compares the methods by relating the number of optimised parameters and balanced accuracy on both datasets. S2P is the best choice for the task as the results are close to the top-left corner in both datasets. On IPD, GA-MLP and MLP have a slightly higher number of parameters than S2P and a decrease in the classification performance, whereas GPA has a similar performance but with one order of magnitude higher than S2P in number of optimised parameters. Unlike the performance on IPD, the performance of these four models is similar to each other when evaluated on PrivacyAlert.  
On the contrary, the number of optimised parameters for GIP and MLP is five orders of magnitude higher than that of other models, while achieving lower performance. 

%%%%%%%%%%%%%%%%%%%%%%%%%%%%
\pgfplotstableread{acc_size_ipd.txt}\sizevsacc
\pgfplotstableread{acc_size_privacyalert.txt}\sizevsaccprivacyalert

\tikzstyle{every pin}=[
font=\tiny,]

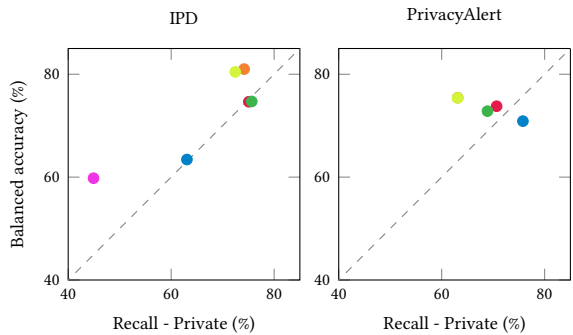
\begin{figure}[t!]
    \centering
    \begin{tikzpicture}
        \begin{axis}[
            width=0.55\columnwidth,
            height=0.55\columnwidth,
            xlabel={Recall - Private (\%)},
            ylabel={Balanced accuracy (\%)},
            ymin=40, ymax=85,
            xmin=40, xmax=85,
            xlabel near ticks,
            ylabel near ticks,
            label style={font=\footnotesize},
            tick label style={font=\scriptsize},
            title={IPD},
            title style={font=\footnotesize},
            scatter/classes={
            a={mark=*,dc1},
            b={mark=*,dc8},
            c={mark=*,dc2},
            d={mark=*,dc4},
            e={mark=*,dc5},
            f={mark=*,dc9}
            }
        ]
        \addplot+[scatter, only marks, scatter src=explicit symbolic] table[x=rec-priv,y=ba, meta=label]{\sizevsacc};
        \addplot+[dashed, color=gray,mark=none] coordinates{(0,0) (100,100)};
        \end{axis}
    \end{tikzpicture}
    \begin{tikzpicture}
        \begin{axis}[
            width=0.55\columnwidth,
            height=0.55\columnwidth,
            xlabel={Recall - Private (\%)},
            ymin=40, ymax=85,
            xmin=40, xmax=85,
            xlabel near ticks,
            ylabel near ticks,
            label style={font=\footnotesize},
            tick label style={font=\scriptsize},
            title={PrivacyAlert},
            title style={font=\footnotesize},
            scatter/classes={
            a={mark=*,dc1},
            b={mark=*,dc8},
            c={mark=*,dc2},
            d={mark=*,dc4},
            e={mark=*,dc5},
            f={mark=*,dc9}
            }
        ]
        \addplot+[scatter, only marks, scatter src=explicit symbolic] table[x=rec-priv,y=ba, meta=label]{\sizevsaccprivacyalert};
        \addplot+[dashed, color=gray,mark=none] coordinates{(0,0) (100,100)};
        \end{axis}
    \end{tikzpicture}
    \caption{Comparison of the pipelines when relating recall of private class and balanced accuracy. Best performance on the top-right corner. 
    Note that relying only on scenes as visual entities makes S2P and GPA correctly recognise less private images in PrivacyAlert, showing the different underlying distributions of the two datasets.
    Legend:
    \protect\tikz \protect\fill[dc1,fill=dc1] (1,1) circle (0.5ex);~MLP,
    \protect\tikz \protect\fill[dc8,fill=dc8] (1,1) circle (0.5ex);~MLP-I,
    \protect\tikz \protect\fill[dc2,fill=dc2] (1,1) circle (0.5ex);~GA-MLP,
    \protect\tikz \protect\fill[dc4,fill=dc4] (1,1) circle (0.5ex);~GIP,
    \protect\tikz \protect\fill[dc5,fill=dc5] (1,1) circle (0.5ex);~GPA,
    \protect\tikz \protect\fill[dc9,fill=dc9] (1,1) circle (0.5ex);~S2P.     
    }
    \label{fig:recallvssacc}
    \vspace{-7pt}
\end{figure}

\begin{figure}[t!]
    \centering
    \begin{tikzpicture}
        \begin{semilogxaxis}[
            width=0.55\columnwidth,
            height=0.55\columnwidth,
            xlabel={\# parameters},
            ylabel={Balanced accuracy (\%)},
            ymin=50, ymax=85,
            xtick={1000,10000,100000,1000000,10000000,100000000,1000000000},
            xlabel near ticks,
            ylabel near ticks,
            label style={font=\footnotesize},
            tick label style={font=\scriptsize},
            title={IPD},
            title style={font=\footnotesize},
            scatter/classes={
            a={mark=*,dc1},
            b={mark=*,dc8},
            c={mark=*,dc2},
            d={mark=*,dc4},
            e={mark=*,dc5},
            f={mark=*,dc9}
            }
        ]
        \addplot+[dashed, color=gray,mark=none] coordinates{(10000,0) (10000,100)};
        \addplot+[dashed, color=gray,mark=none] coordinates{(1000000,0) (1000000,100)};
        \addplot+[dashed, color=gray,mark=none] coordinates{(100000000,0) (100000000,100)};
        \addplot+[scatter, only marks, scatter src=explicit symbolic] table[x=size,y=ba, meta=label]{\sizevsacc};
        \end{semilogxaxis}
    \end{tikzpicture}
    \begin{tikzpicture}
        \begin{semilogxaxis}[
            width=0.55\columnwidth,
            height=0.55\columnwidth,
            xlabel={\# parameters},
            ymin=50, ymax=85,
            xtick={1000,10000,100000,1000000,10000000,100000000,1000000000},
            yticklabels={},
            xlabel near ticks,
            ylabel near ticks,
            label style={font=\footnotesize},
            tick label style={font=\scriptsize},
            title={PrivacyAlert},
            title style={font=\footnotesize},
            scatter/classes={
            a={mark=*,dc1},
            b={mark=*,dc8},
            c={mark=*,dc2},
            d={mark=*,dc4},
            e={mark=*,dc5},
            f={mark=*,dc9}
            }
        ]
        \addplot+[dashed, color=gray,mark=none] coordinates{(10000,0) (10000,100)};
        \addplot+[dashed, color=gray,mark=none] coordinates{(1000000,0) (1000000,100)};
        \addplot+[dashed, color=gray,mark=none] coordinates{(100000000,0) (100000000,100)};
        \addplot+[scatter, only marks, scatter src=explicit symbolic] table[x=size,y=ba, meta=label]{\sizevsaccprivacyalert};
        \end{semilogxaxis}
    \end{tikzpicture}
    \caption{Comparison of the pipelines when relating the number of parameters optimised for image privacy classification (\# parameters) and balanced accuracy. Best performance on the top-left corner. Fixed parameters of pre-trained CNNs and those of the object detector are not counted. The dotted line represents an increment of 2 in the order of magnitude of \# parameters. Note the logarithmic scale of the x-axis. 
    Legend:
    \protect\tikz \protect\fill[dc1,fill=dc1] (1,1) circle (0.5ex);~MLP,
    \protect\tikz \protect\fill[dc8,fill=dc8] (1,1) circle (0.5ex);~MLP-I,
    \protect\tikz \protect\fill[dc2,fill=dc2] (1,1) circle (0.5ex);~GA-MLP,
    \protect\tikz \protect\fill[dc4,fill=dc4] (1,1) circle (0.5ex);~GIP,
    \protect\tikz \protect\fill[dc5,fill=dc5] (1,1) circle (0.5ex);~GPA,
    \protect\tikz \protect\fill[dc9,fill=dc9] (1,1) circle (0.5ex);~S2P.     
    }
    \label{fig:sizevsacc}
    \vspace{-7pt}
\end{figure}
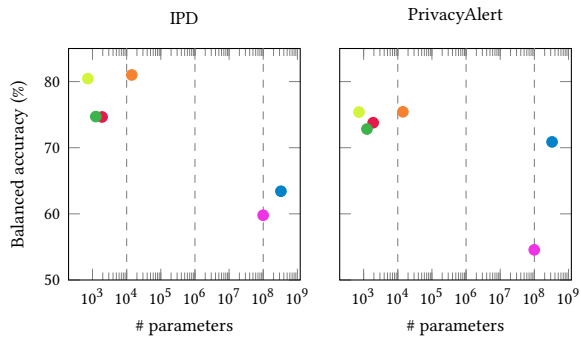
%%%%%%%%%%%%%%%%%%%%%%%%%%%%

%%%%%%%%%%%%%%%%%%%%%%%%%%%%
\begin{figure*}[t!]
    \centering
    \begin{tabular}{c}
    % Images predicted as public but labelled as private on PrivacyAlert
    \includegraphics[height=0.12\linewidth]{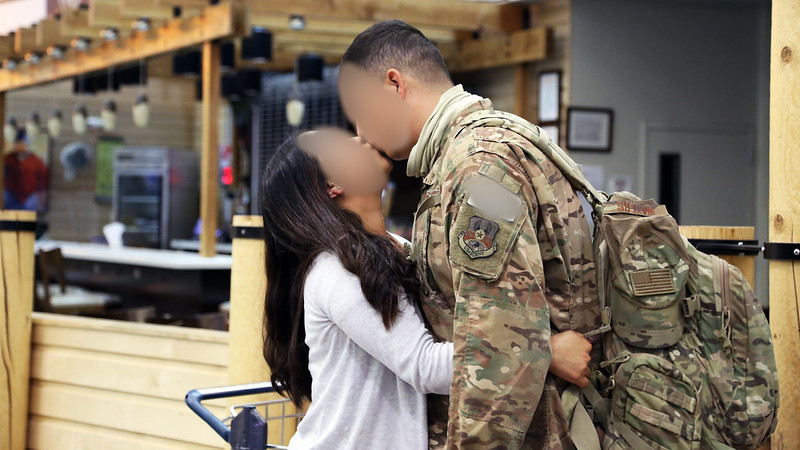}
    \includegraphics[height=0.12\linewidth]{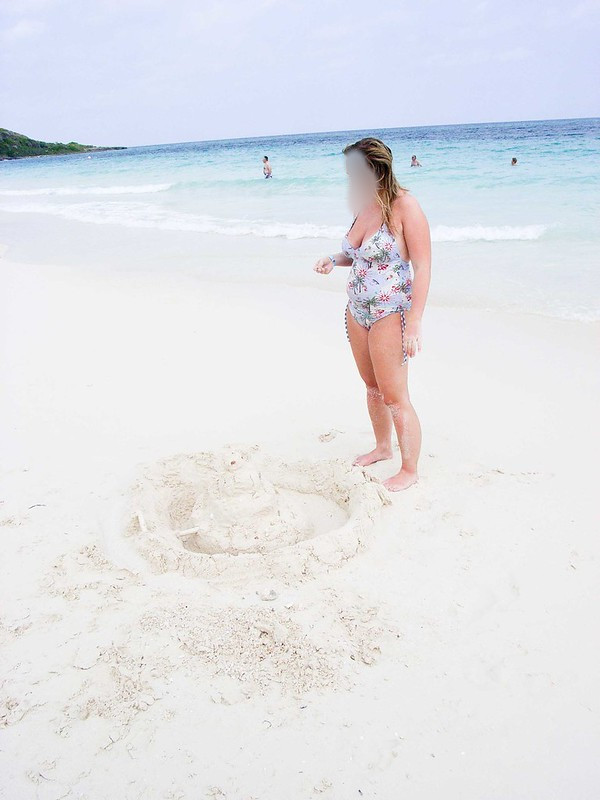}
    \includegraphics[height=0.12\linewidth]{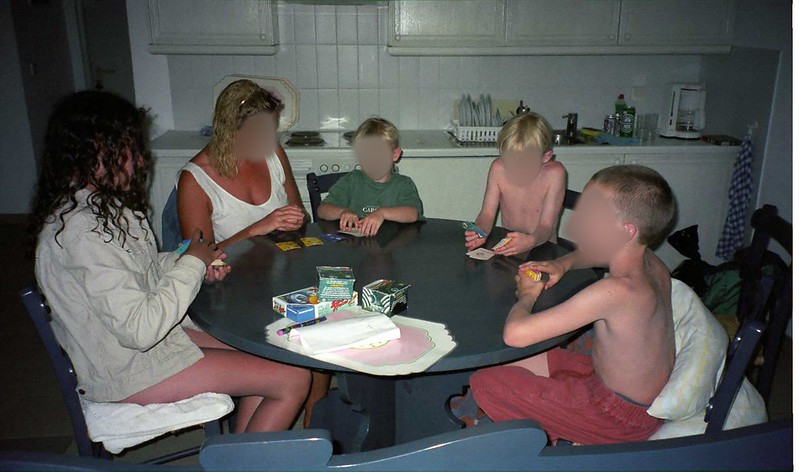}
    \includegraphics[height=0.12\linewidth]{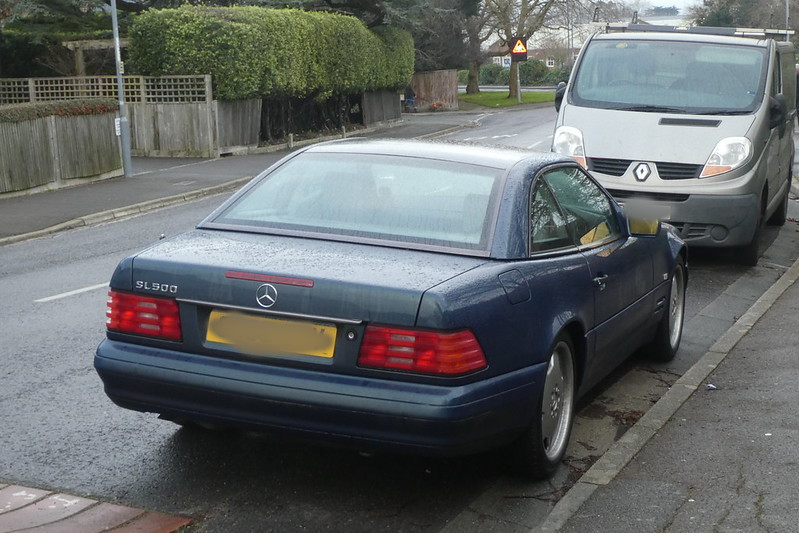}
    \includegraphics[height=0.12\linewidth]{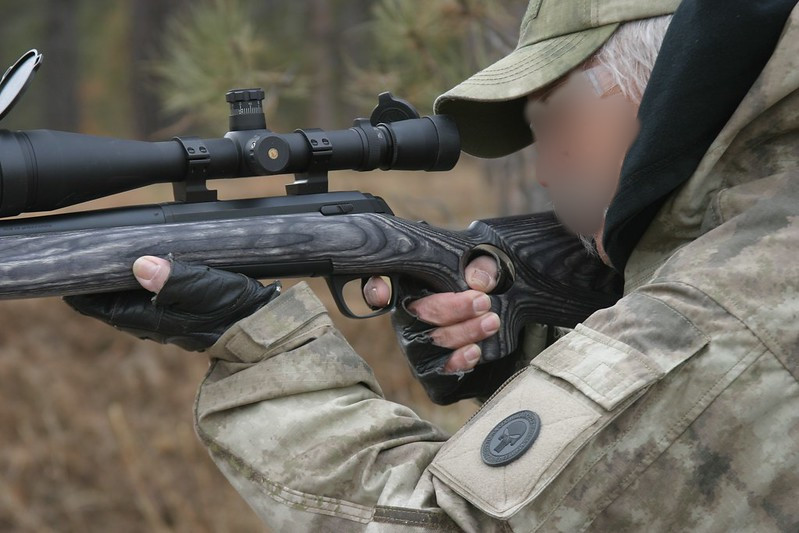}
    \includegraphics[height=0.12\linewidth]{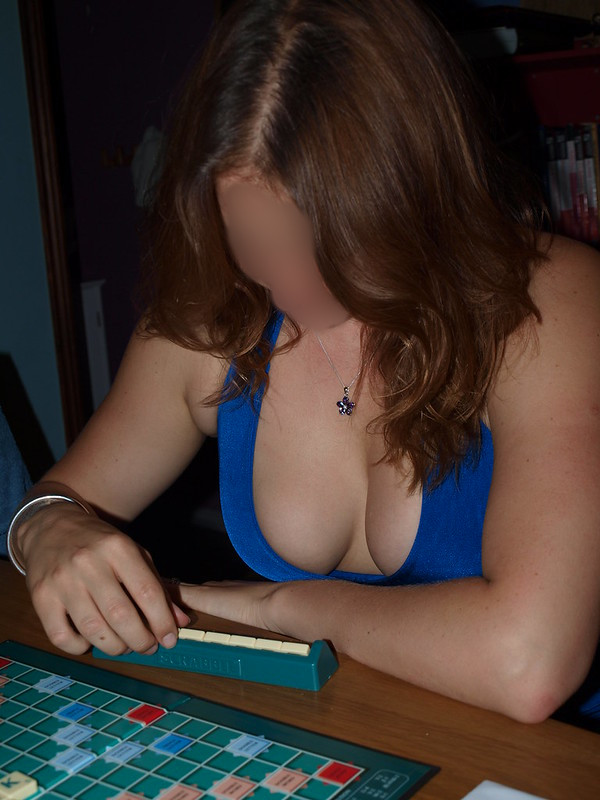}
    \\
    \includegraphics[height=0.12\linewidth]{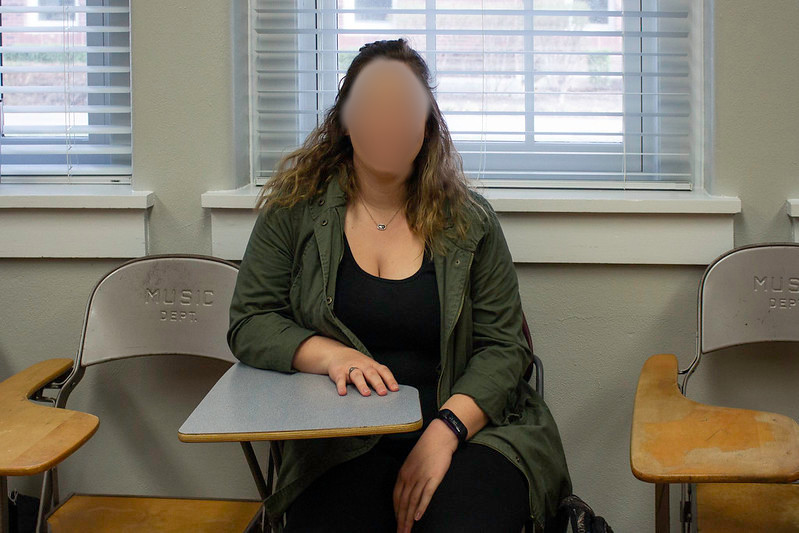}
    \includegraphics[height=0.12\linewidth]{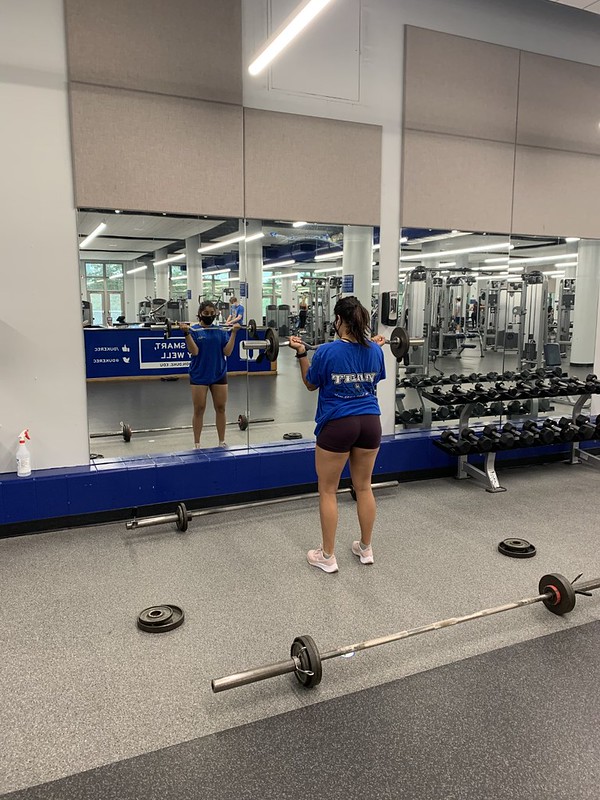}
    \includegraphics[height=0.12\linewidth]{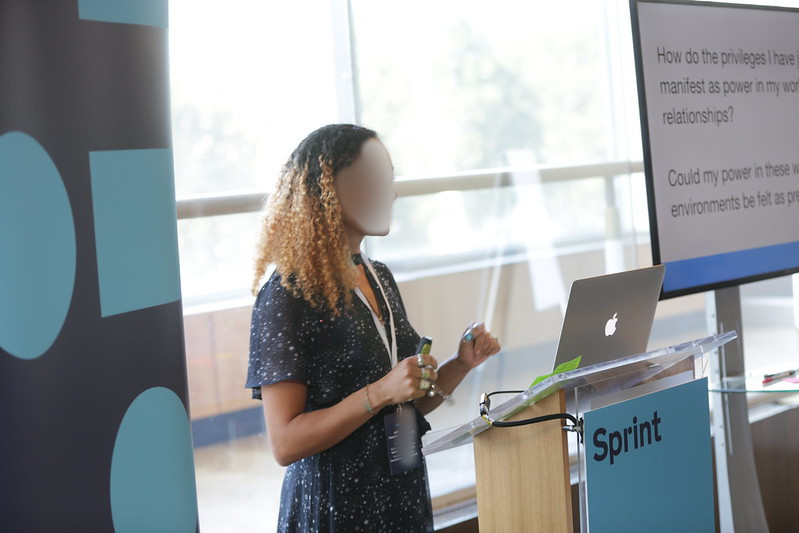}
    \includegraphics[height=0.12\linewidth]{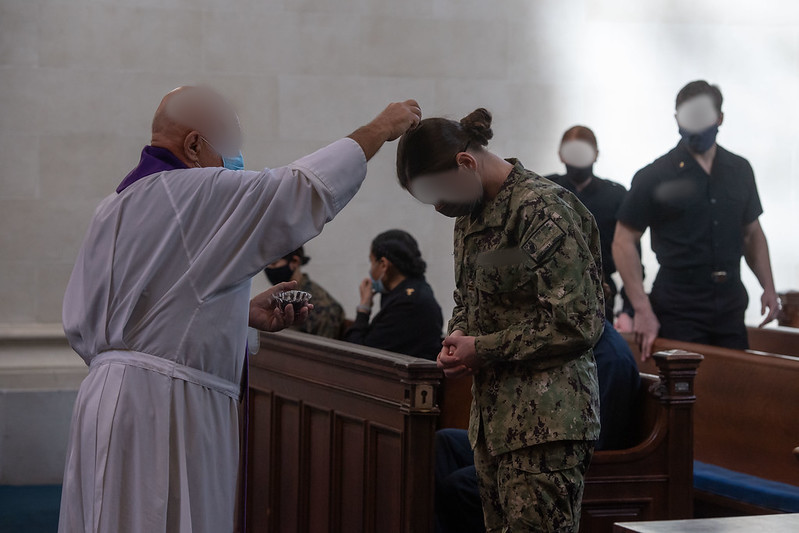}
    \includegraphics[height=0.12\linewidth]{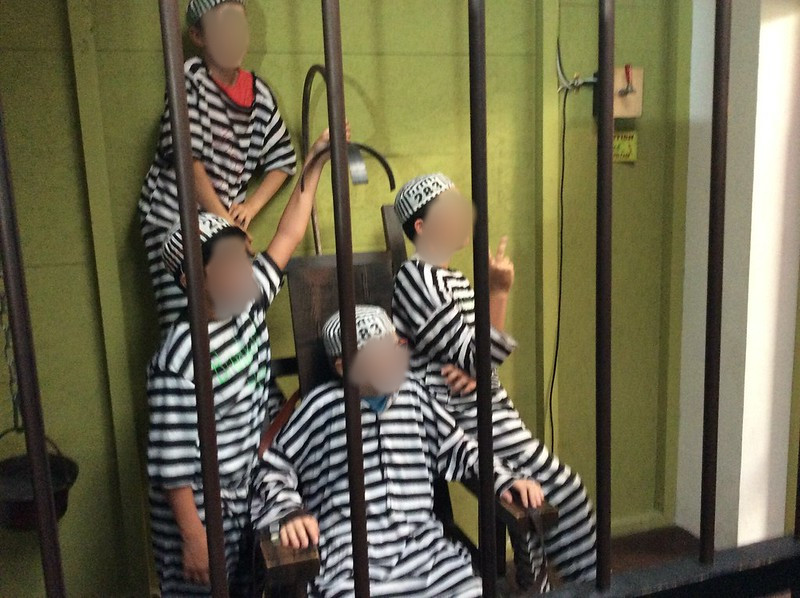}
    \includegraphics[height=0.12\linewidth]{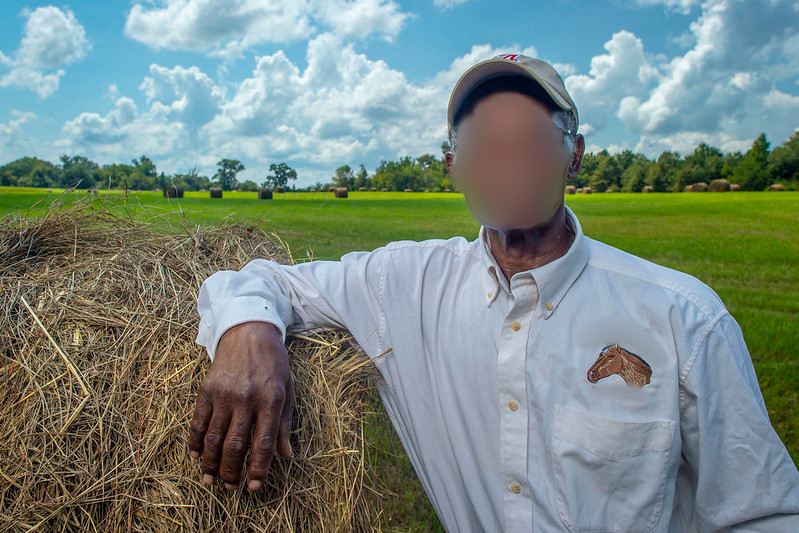}
    \\
    \midrule
    % Images predicted as private but labelled as public on PrivacyAlert
    \includegraphics[height=0.12\linewidth]{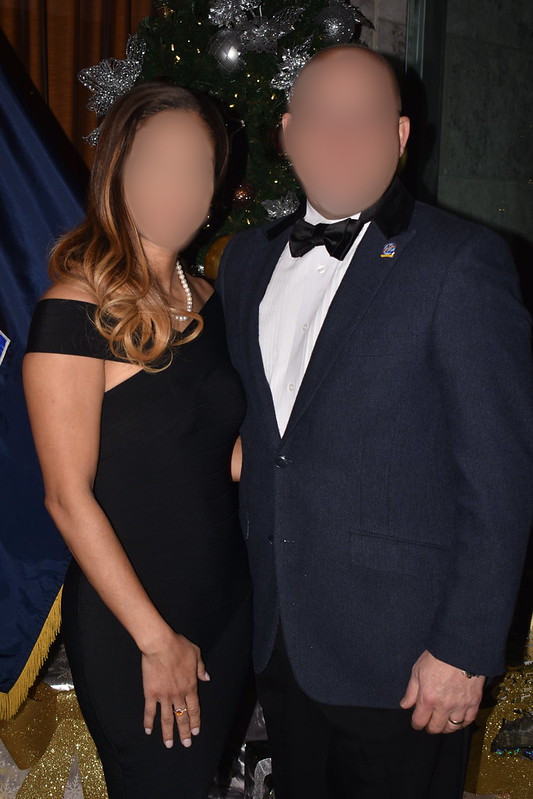}
    \includegraphics[height=0.12\linewidth]{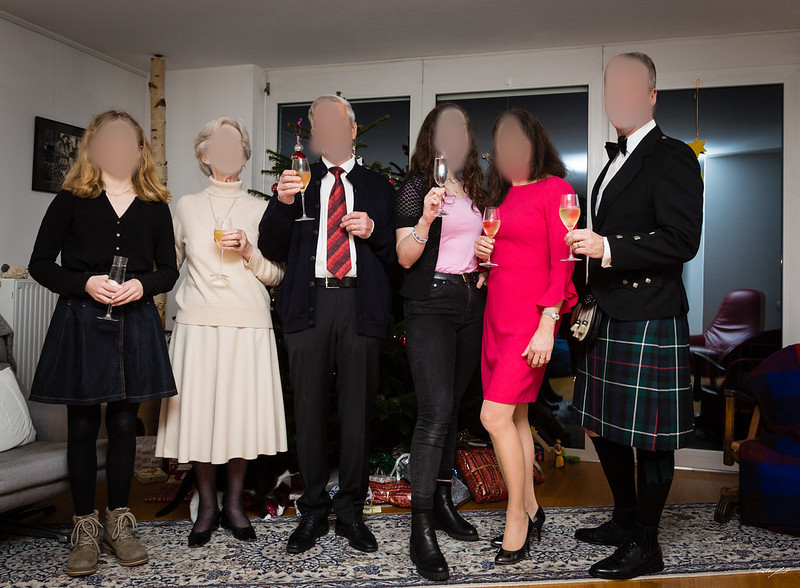}
    \includegraphics[height=0.12\linewidth]{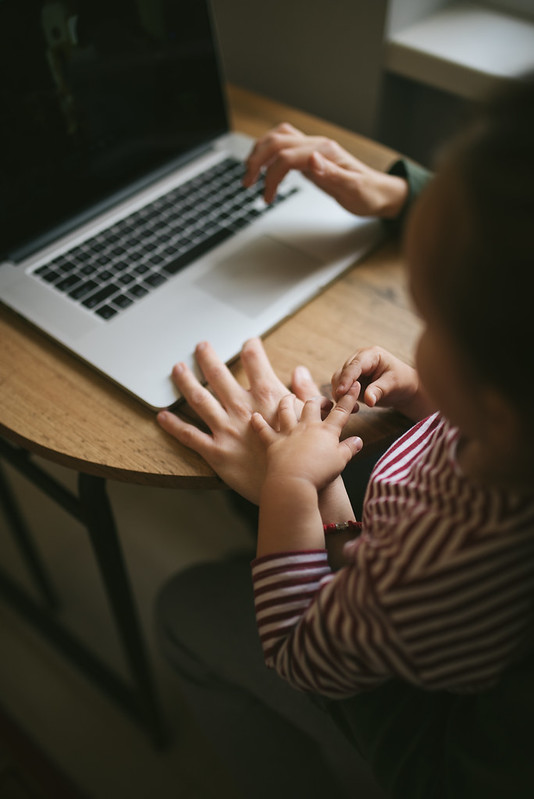}
    \includegraphics[height=0.12\linewidth]{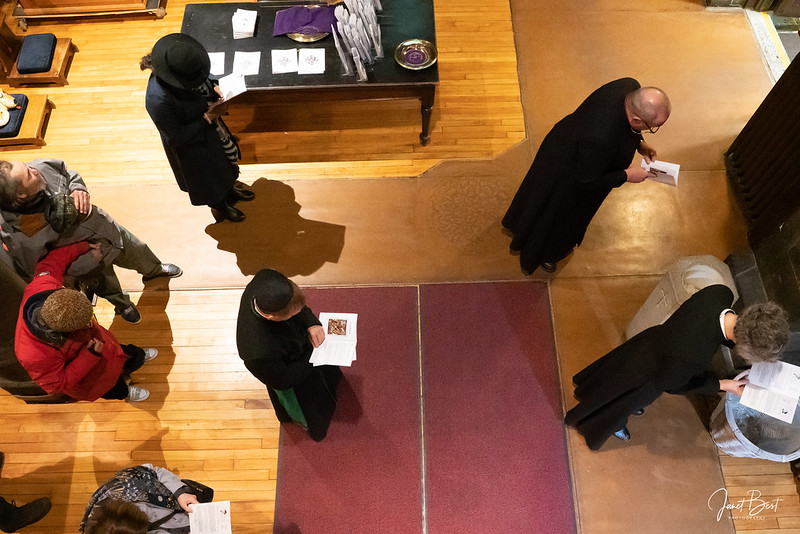}
    \includegraphics[height=0.12\linewidth]{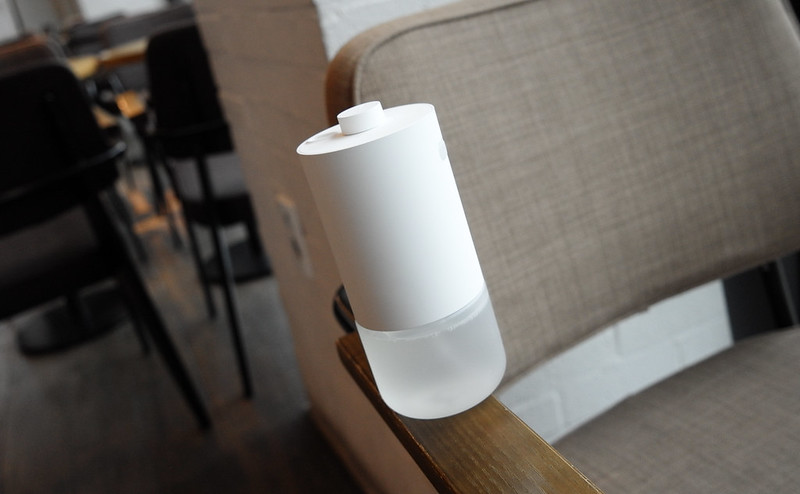}
    \includegraphics[height=0.12\linewidth]{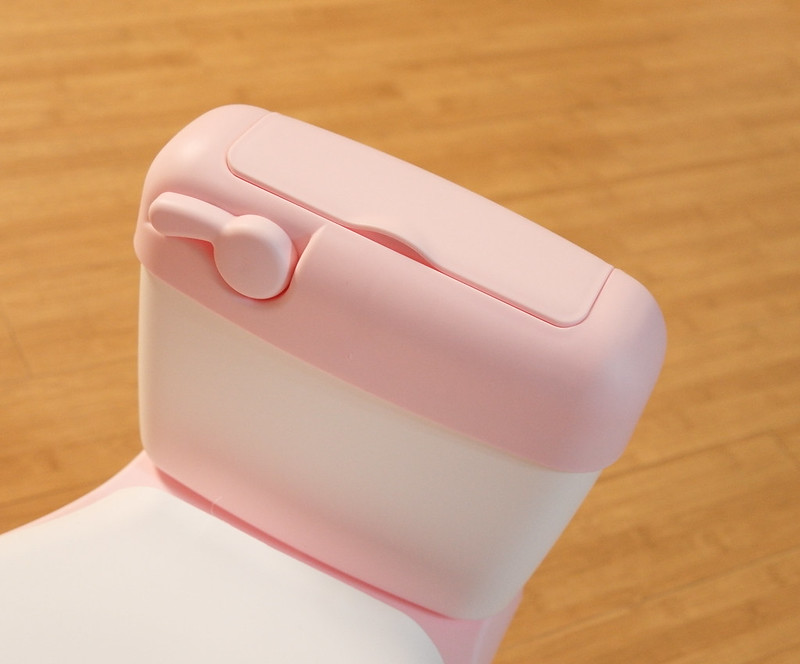}
    \includegraphics[height=0.12\linewidth]{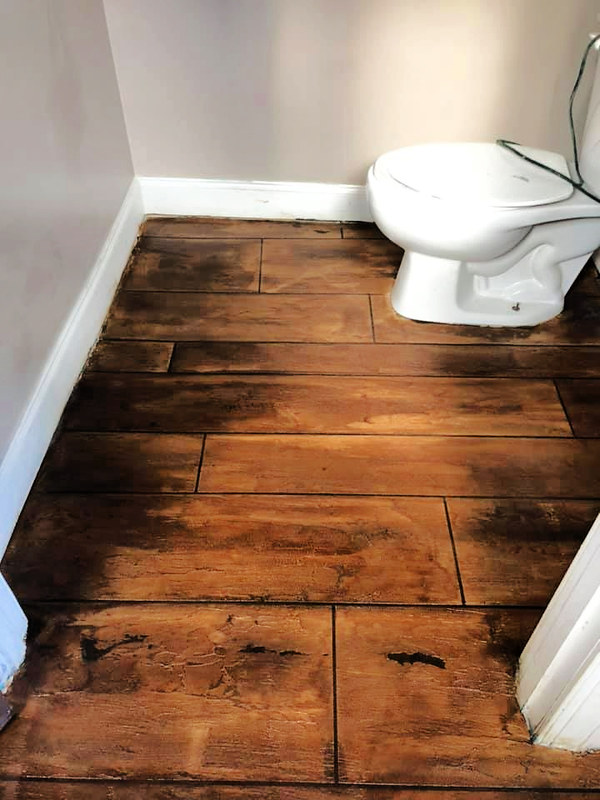}
    \\
    \includegraphics[height=0.12\linewidth]{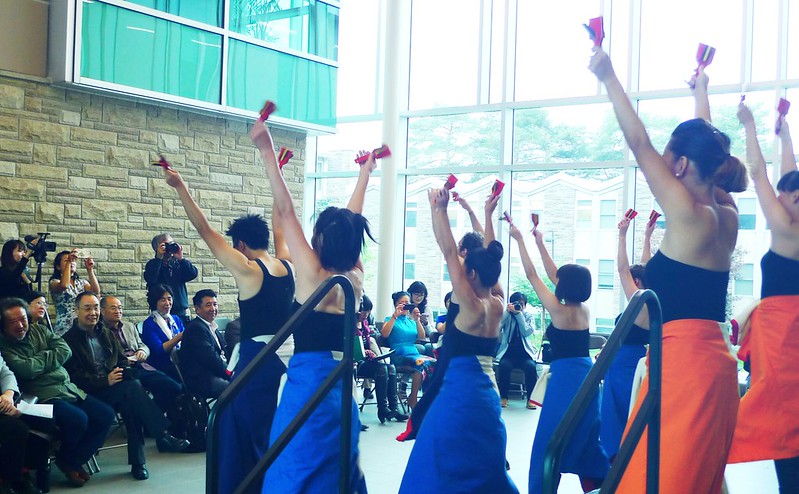}
    \includegraphics[height=0.12\linewidth]{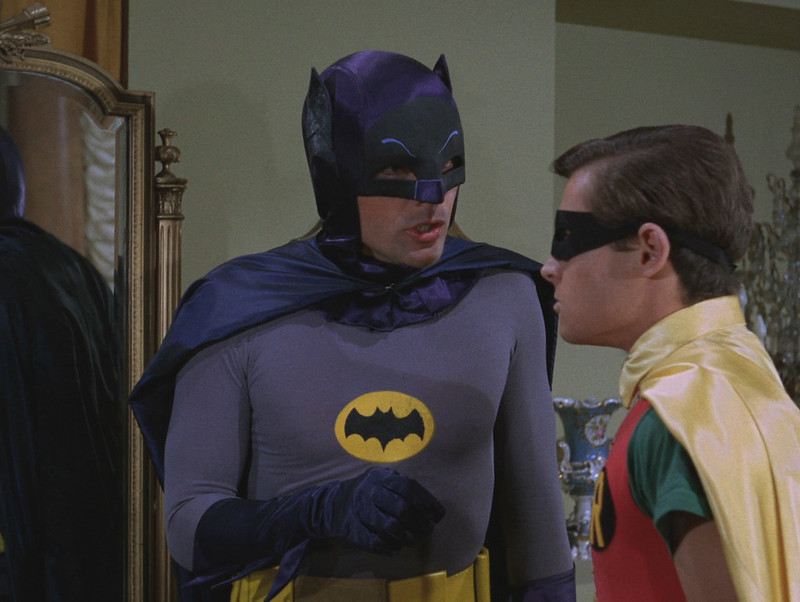}
    \includegraphics[height=0.12\linewidth]{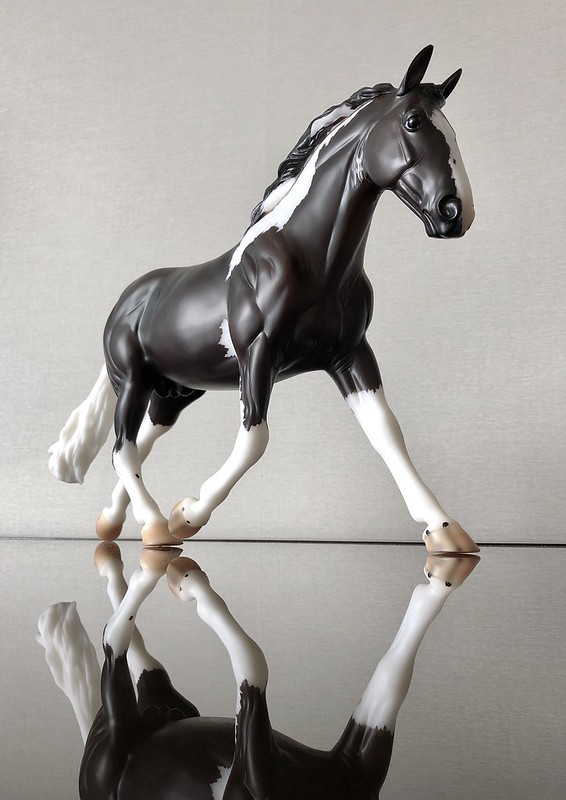}
    \includegraphics[height=0.12\linewidth]{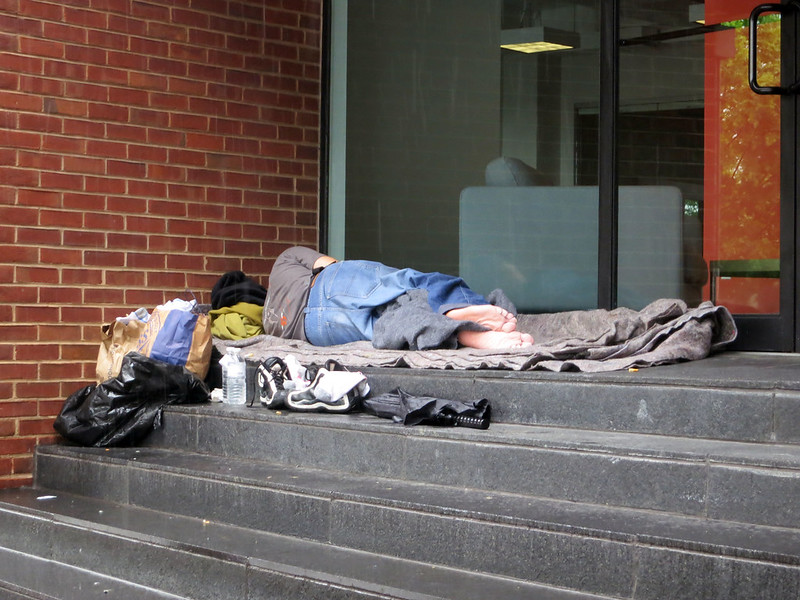}
    \includegraphics[height=0.12\linewidth]{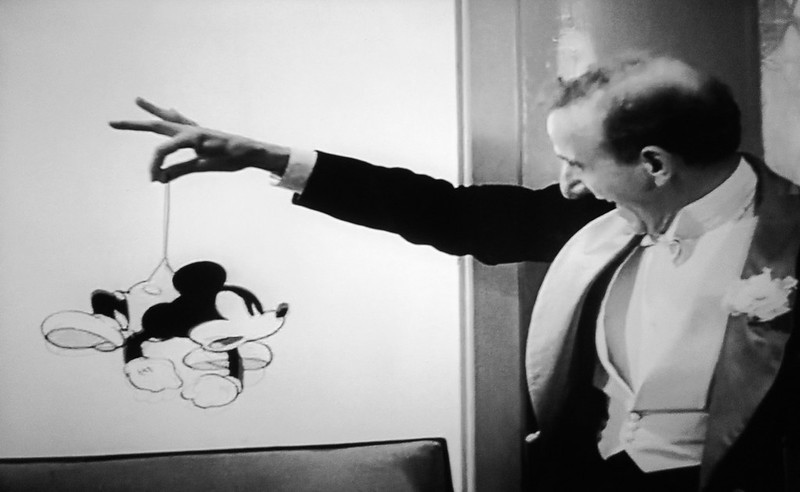}
    \includegraphics[height=0.12\linewidth]{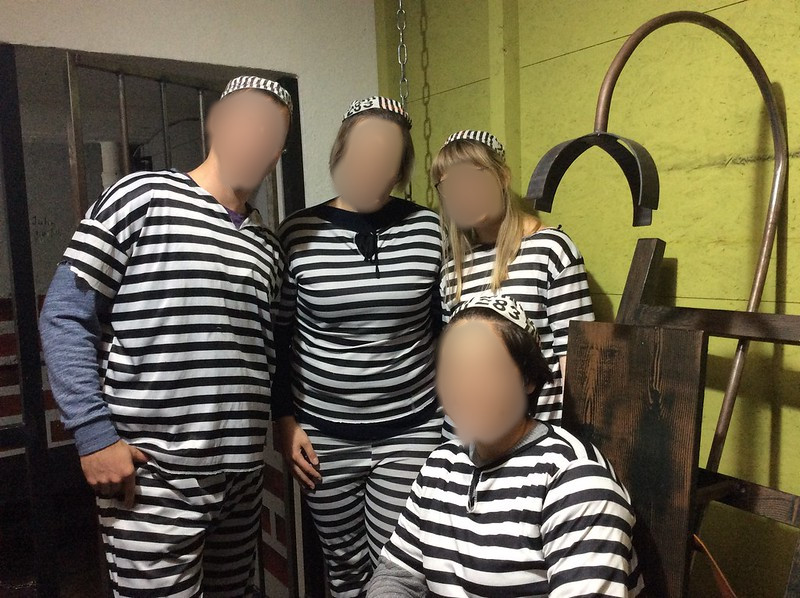}
    \end{tabular}
    \vspace{-7pt}
    \caption{Misclassification results of S2P on the testing set of PrivacyAlert. First and second rows: images labelled as private but predicted as public. Third and fourth rows: images labelled as public but predicted as private. Obfuscation added to the images.
    }
    \label{fig:s2pfalsepositives}
    \vspace{-10pt}
\end{figure*}

%%%%%%%%%%%%%5

Furthermore, in Appendix~\ref{app:extracomparison}, we compare S2P with different multimodal models, previously evaluated on PrivacyAlert~\cite{Zhao2022ICWSM_PrivacyAlert}, other methods using different classifiers (e.g., Support Vector Machines~\cite{Tonge2016AAAI}) and training strategies~\cite{Zhao2022ICWSM_PrivacyAlert}, and two variants of S2P replacing the FC layer with a 1-layer and 2-layer MLP. As expected, multimodal models achieves higher performance than S2P and the best-performing method fuses privacy predictions from specialized models using objects, scenes, and user tags~\cite{Zhao2023TWEB,Zhao2022ICWSM_PrivacyAlert}. Note that this comparison, however, is unfair due to missing images in the datasets and the use of the additional modality (results also directly taken from the evaluation done by Zhao et al.~\cite{Zhao2022ICWSM_PrivacyAlert}). Combining the ResNet-50 pre-trained on Places365 with an SVM trained for privacy achieves the best recall on the private class (81.42\% on IPD and 79.78\% on PrivacyAlert) and the best balanced accuracy (83.08\% on IPD and 78.48\% on PrivacyAlert). Results are expected given the small size of the datasets and the features extracted by a pre-trained CNN, but scaling to larger dataset size is a known drawback for SVMs. Furthermore, replacing the FC layer with a 1-layer MLP allows S2P to improve the classification performance (more comparable to ResNet-50 with SVM).

\subsection{Failure analysis to misclassifications}

Fig.~\ref{fig:s2pfalsepositives} shows misclassified images by S2P on PrivacyAlert. As expected, S2P is influenced by the predicted scenes in the backbone component and therefore is not biased towards the presence of people to determine if an image is private. The model is also not biased towards only outdoor images or indoor images as both scene types are predicted as either public or private. However, there are indoor environments that the model learned to predict as private: for example, bathroom in the last two images of the third row even if the second last photo depicts a toy as a toilet. There are scenes, such as religious environments/ceremonies and escape rooms, that are ambiguously predicted as either public or private, denoting that the model also rely on the features related to other scene types to make the decision. However, the model confuses a top-view image of a ceremony as private when none of the people is recognisable, which can also be the main factor that led the image to be annotated as public. On the contrary, visible faces in an another image showing a ceremony are not considered by the model as public. S2P also misclassifies indoor parties and performance as private compared to the dataset annotations. Office-like scenes (e.g., person in the classroom, person presenting next to a laptop and screen) are predicted as public, but these images may also denote a private setting. Additionally, the photo of a kitchen with the family and children sitting around the table is predicted as public, when one could expect this to be private. A public performance and the representation of famous characters are incorrectly predicted as private, showing that simply relying on the scene information is insufficient in these cases. Similarly, S2P fails to recognise private images of an individual in various contexts (beach, gym) or of a couple outdoors. These are places that can commonly be related to a public context and, based on the predictions, the model learned to associate these scenes to the public label. However, the model cannot distinguish when this scene type is public or private due to the lack of additional information, such as the person's presence.

\subsection{Discussion} 
\label{ssec:discussion}

We conclude that the performance of previous graph-based models~\cite{Yang2020PR,Stoidis2022BigMM} can be largely attributed to the fine-tuning of the CNNs. For example, the graph processing component of GPA~\cite{Stoidis2022BigMM}  minimally improves performance compared to S2P, which optimises only 732 parameters. Additionally, we showed that GIP~\cite{Yang2020PR} optimises an unnecessarily large number of parameters when also fine-tuning the CNNs.  
We also showed that using deep features, extracted by a fixed CNN, for object nodes and zero-value features for privacy nodes allows the GNN of GIP to refine the node features and achieve higher classification performance than a pseudo-random classifier while not degenerating to predict only one class. 

The performance of this GIP alternative is still lower than the original GIP, GPA, and S2P, and the large number of parameters to be optimised depends on the dimensionality of the node features. However, high-dimensional feature vectors extracted with CNNs might not be needed as shown by MLP and GA-MLP that achieve promising results while using the simple object cardinality as feature and less than 2,000 parameters to be optimised for the task. Using conditional rules on the confidence of person presence and cardinality~\cite{Xompero2024CVPRW_XAI4CV} is too restrictive and not sufficient for classifying images as private, as seen in IPD with respect to PrivacyAlert.

\section{Conclusion}
\label{sec:conclusion}

We proposed a simple model that relates scenes to privacy by using transfer learning combined with a CNN. We also discussed existing design choices to relate visual entities in a graph and learn privacy with a graph neural network while addressing the problem of limited training data available for the task. 
Experiments showed that the simple model achieve comparable performance to graph-based models while optimising only a small number of parameters for the image privacy task (732 instead of 14,000 or 500 millions). We also analysed various design choices and training strategies for the graph-based models to understand the relative contribution of individual components, as previously masked by the end-to-end training. We showed that the fine-tuning of the convolutional neural networks used to initialise the graph node features, especially the privacy nodes, is the driving factor for the performance. 

Future work includes the design of a graph-based model that uses relevant human-interpretable features for each visual entity and relates the entities in such a way that the recognition of private images increases, while limiting the number of parameters. 

\begin{acks}
The authors would like to thank Alina Elena Baia, Olena Hrynenko, and Darya Baranouskaya for the proofreading of the article and suggestions, and Dimitrios Stoidis to provide the images of the PicAlert dataset as used in the Graph Privacy Advisor. This work was supported by the CHIST-ERA programme through the project GraphNEx, under UK EPSRC grant EP/V062107/1.
\end{acks}

\begin{table*}[t!]
    \centering
    \footnotesize
    \caption{Comparison of annotation procedures between datasets for image privacy classification.
    \vspace{-7pt}
    }
    \begin{tabular}{cllllc}
    \toprule
    Dataset & Source & Annotation instruction & Initial labels & Labels & \# images\\
    \midrule
    PrivacyAlert~\cite{Zhao2022ICWSM_PrivacyAlert} & Flickr & \parbox{5cm}{``\textit{Assume you have taken these photos, and you are about to upload them on your favourite social network or content sharing site (e.g., Flickr, Facebook, Google+, Instagram). Please tell us whether these images are either private or public in nature. Assume that the people in the photos are those that you know.}''} & \parbox{4.5cm}{1. not being uploaded at all\\2. shared confidentially with trusted people\\3. shared with anyone in a social network\\4. shared with anyone online (OK to see)} & \parbox{1cm}{- public\\- private} & 6,793 \\
    \midrule
    PicAlert~\cite{Zerr2012CIKM_PicAlert} & Flickr & \parbox{5cm}{``\textit{Private are photos which have to do with the private sphere (like self portraits, family, friends, your home) or contain objects that you would not share with the entire world (like a private email). The rest are public. In case no decision can be made the picture should be marked as undecidable.}''} & \parbox{4.5cm}{1. private\\2. public \\3. undecidable} & \parbox{1.5cm}{- public\\- private} & \\
    \midrule
    *VISPR~\cite{Orekondy2017ICCV} & \parbox{1.5cm}{OpenImages~\cite{Kuznetsova2020IJCV_OpenImages}\\Flickr\\Twitter} & \parbox{5cm}{Selection from a predefined set of attributes (see also Table~\ref{tab:visprlabels})} & \parbox{4.5cm}{- 104 privacy attributes\\- \textit{safe}\\- unsure} & \parbox{1.5cm}{- 67 privacy\\attributes\\- \textit{safe}} & 22,167  \\
    \midrule
    IPD~\cite{Yang2020PR} &  \parbox{1.5cm}{PicAlert\\VISPR} & -- & -- & \parbox{1.5cm}{- public\\- private} & 34,562  \\
    \bottomrule \addlinespace[\belowrulesep]
    \multicolumn{6}{l}{\parbox{0.95\linewidth}{\scriptsize{*Most of the images were randomly sampled from another dataset, OpenImages~\cite{Kuznetsova2020IJCV_OpenImages}, that selected images from Flick under Public Domain license; some of the images were selected by the authors from Flickr to better balance the classes, and 50 images of credit cards were manually selected from Twitter.}}}
    \end{tabular}    
    \label{tab:datasetsprivacy}
    \vspace{-5pt}
\end{table*}

\bibliographystyle{ACM-Reference-Format}
\bibliography{main}

%%% -*-BibTeX-*-
%%% Do NOT edit. File created by BibTeX with style
%%% ACM-Reference-Format-Journals [18-Jan-2012].

\begin{thebibliography}{65}

%%% ====================================================================
%%% NOTE TO THE USER: you can override these defaults by providing
%%% customized versions of any of these macros before the \bibliography
%%% command.  Each of them MUST provide its own final punctuation,
%%% except for \shownote{}, \showDOI{}, and \showURL{}.  The latter two
%%% do not use final punctuation, in order to avoid confusing it with
%%% the Web address.
%%%
%%% To suppress output of a particular field, define its macro to expand
%%% to an empty string, or better, \unskip, like this:
%%%
%%% \newcommand{\showDOI}[1]{\unskip}   % LaTeX syntax
%%%
%%% \def \showDOI #1{\unskip}           % plain TeX syntax
%%%
%%% ====================================================================

\ifx \showCODEN    \undefined \def \showCODEN     #1{\unskip}     \fi
\ifx \showDOI      \undefined \def \showDOI       #1{#1}\fi
\ifx \showISBNx    \undefined \def \showISBNx     #1{\unskip}     \fi
\ifx \showISBNxiii \undefined \def \showISBNxiii  #1{\unskip}     \fi
\ifx \showISSN     \undefined \def \showISSN      #1{\unskip}     \fi
\ifx \showLCCN     \undefined \def \showLCCN      #1{\unskip}     \fi
\ifx \shownote     \undefined \def \shownote      #1{#1}          \fi
\ifx \showarticletitle \undefined \def \showarticletitle #1{#1}   \fi
\ifx \showURL      \undefined \def \showURL       {\relax}        \fi
% The following commands are used for tagged output and should be
% invisible to TeX
\providecommand\bibfield[2]{#2}
\providecommand\bibinfo[2]{#2}
\providecommand\natexlab[1]{#1}
\providecommand\showeprint[2][]{arXiv:#2}

\bibitem[Arias-Cabarcos et~al\mbox{.}(2023)]%
        {AriasCabarcos2023PoPETs}
\bibfield{author}{\bibinfo{person}{P. Arias-Cabarcos}, \bibinfo{person}{S. Khalili}, {and} \bibinfo{person}{T. Strufe}.} \bibinfo{year}{2023}\natexlab{}.
\newblock \showarticletitle{`Surprised, Shocked, Worried': User Reactions to Facebook Data Collection from Third Parties}.
\newblock  \bibinfo{volume}{2023}, \bibinfo{number}{1} (\bibinfo{year}{2023}), \bibinfo{pages}{384--399}.
\newblock
\urldef\tempurl%
\url{https://doi.org/10.56553/popets-2023-0023}
\showURL{%
\tempurl}


\bibitem[Baranouskaya and Cavallaro(2023)]%
        {Baranouskaya2023ICIP}
\bibfield{author}{\bibinfo{person}{D. Baranouskaya} {and} \bibinfo{person}{A. Cavallaro}.} \bibinfo{year}{2023}\natexlab{}.
\newblock \showarticletitle{Human-interpretable and Deep Features for Image Privacy Classification}. In \bibinfo{booktitle}{\emph{IEEE Int. Conf. Image Process.}}
\newblock


\bibitem[Battaglia et~al\mbox{.}(2018)]%
        {battaglia2018relationalinductivebiasesdeep}
\bibfield{author}{\bibinfo{person}{P.~W. Battaglia}, \bibinfo{person}{J.~B. Hamrick}, \bibinfo{person}{V. Bapst}, \bibinfo{person}{A. Sanchez-Gonzalez}, \bibinfo{person}{V. Zambaldi}, \bibinfo{person}{M. Malinowski}, \bibinfo{person}{A. Tacchetti}, \bibinfo{person}{D. Raposo}, \bibinfo{person}{A. Santoro}, \bibinfo{person}{R. Faulkner}, \bibinfo{person}{C. Gulcehre}, \bibinfo{person}{F. Song}, \bibinfo{person}{A. Ballard}, \bibinfo{person}{J. Gilmer}, \bibinfo{person}{G. Dahl}, \bibinfo{person}{A. Vaswani}, \bibinfo{person}{K. Allen}, \bibinfo{person}{C. Nash}, \bibinfo{person}{V. Langston}, \bibinfo{person}{C. Dyer}, \bibinfo{person}{N. Heess}, \bibinfo{person}{D. Wierstra}, \bibinfo{person}{P. Kohli}, \bibinfo{person}{M. Botvinick}, \bibinfo{person}{O. Vinyals}, \bibinfo{person}{Y. Li}, {and} \bibinfo{person}{R. Pascanu}.} \bibinfo{year}{2018}\natexlab{}.
\newblock \bibinfo{title}{Relational Inductive Biases, Deep Learning, and Graph Networks}.
\newblock
\newblock
\showeprint[arxiv]{1806.01261}~[cs.LG]


\bibitem[Bishop and Bishop(2024)]%
        {Bishop2024_DeepLearning}
\bibfield{author}{\bibinfo{person}{C.~M. Bishop} {and} \bibinfo{person}{H. Bishop}.} \bibinfo{year}{2024}\natexlab{}.
\newblock \bibinfo{booktitle}{\emph{Deep Learning: Foundations and Concepts} (\bibinfo{edition}{1} ed.)}.
\newblock \bibinfo{publisher}{Springer Cham}.
\newblock
\urldef\tempurl%
\url{https://doi.org/10.1007/978-3-031-45468-4}
\showURL{%
\tempurl}


\bibitem[Bronstein et~al\mbox{.}(2017)]%
        {Bronstein2017SPM}
\bibfield{author}{\bibinfo{person}{M. Bronstein}, \bibinfo{person}{J. Bruna}, \bibinfo{person}{Y. LeCun}, \bibinfo{person}{A. Szlam}, {and} \bibinfo{person}{P. Vandergheynst}.} \bibinfo{year}{2017}\natexlab{}.
\newblock \showarticletitle{Geometric Deep Learning: Going beyond Euclidean data}.
\newblock \bibinfo{journal}{\emph{IEEE Signal Process. Magazine}} \bibinfo{volume}{34}, \bibinfo{number}{4} (\bibinfo{year}{2017}), \bibinfo{pages}{18--42}.
\newblock


\bibitem[Buschek et~al\mbox{.}(2015)]%
        {Buschek2015INTERACT}
\bibfield{author}{\bibinfo{person}{D. Buschek}, \bibinfo{person}{M. Bader}, \bibinfo{person}{E. von Zezschwitz}, {and} \bibinfo{person}{A. De~Luca}.} \bibinfo{year}{2015}\natexlab{}.
\newblock \showarticletitle{Automatic Privacy Classification of Personal Photos}. In \bibinfo{booktitle}{\emph{Human-Computer Interaction -- INTERACT}}.
\newblock


\bibitem[Chang et~al\mbox{.}(2023)]%
        {Chang2023TPAMI_SceneGraphsSurvey}
\bibfield{author}{\bibinfo{person}{X. Chang}, \bibinfo{person}{P. Ren}, \bibinfo{person}{P. Xu}, \bibinfo{person}{Z. Li}, \bibinfo{person}{X. Chen}, {and} \bibinfo{person}{A. Hauptmann}.} \bibinfo{year}{2023}\natexlab{}.
\newblock \showarticletitle{A Comprehensive Survey of Scene Graphs: Generation and Application}.
\newblock \bibinfo{journal}{\emph{IEEE Trans. Pattern Anal. Mach. Intell.}} \bibinfo{volume}{45}, \bibinfo{number}{1} (\bibinfo{year}{2023}), \bibinfo{pages}{1--26}.
\newblock


\bibitem[Chen et~al\mbox{.}(2024)]%
        {Chen2024TPAMI}
\bibfield{author}{\bibinfo{person}{C. Chen}, \bibinfo{person}{Y. Wu}, \bibinfo{person}{Q. Dai}, \bibinfo{person}{H. Zhou}, \bibinfo{person}{M. Xu}, \bibinfo{person}{S. Yang}, \bibinfo{person}{X. Han}, {and} \bibinfo{person}{Y. Yu}.} \bibinfo{year}{2024}\natexlab{}.
\newblock \showarticletitle{A Survey on Graph Neural Networks and Graph Transformers in Computer Vision: A Task-Oriented Perspective}.
\newblock \bibinfo{journal}{\emph{IEEE Trans. Pattern Anal. Mach. Intell.}} \bibinfo{volume}{46}, \bibinfo{number}{12} (\bibinfo{year}{2024}), \bibinfo{pages}{10297--10318}.
\newblock


\bibitem[Chung et~al\mbox{.}(2014)]%
        {chung2014empirical}
\bibfield{author}{\bibinfo{person}{J. Chung}, \bibinfo{person}{C. Gulcehre}, \bibinfo{person}{K. Cho}, {and} \bibinfo{person}{Y. Bengio}.} \bibinfo{year}{2014}\natexlab{}.
\newblock \showarticletitle{{Empirical Evaluation of Gated Recurrent Neural Networks on Sequence Modeling}}. In \bibinfo{booktitle}{\emph{Adv. Neural Inf. Process. Syst. Workshop Deep Learning and Representation Learning}}. \bibinfo{address}{Montreal, Canada}.
\newblock


\bibitem[Dalal and Triggs(2005)]%
        {Dalal2005CVPR_HOG}
\bibfield{author}{\bibinfo{person}{N. Dalal} {and} \bibinfo{person}{B. Triggs}.} \bibinfo{year}{2005}\natexlab{}.
\newblock \showarticletitle{Histograms of Oriented Gradients for Human Detection}. In \bibinfo{booktitle}{\emph{Conf. Comput. Vis. Pattern Recognit.}}
\newblock


\bibitem[Deng et~al\mbox{.}(2009)]%
        {Deng2009CVPR_ImageNet}
\bibfield{author}{\bibinfo{person}{J. Deng}, \bibinfo{person}{W. Dong}, \bibinfo{person}{R. Socher}, \bibinfo{person}{L.-J. Li}, \bibinfo{person}{L. Li}, {and} \bibinfo{person}{L. Fei-Fei}.} \bibinfo{year}{2009}\natexlab{}.
\newblock \showarticletitle{ImageNet: A Large-Scale Hierarchical Image Database}. In \bibinfo{booktitle}{\emph{Conf. Comput. Vis. Pattern Recognit.}}
\newblock


\bibitem[Dwivedi et~al\mbox{.}(2023)]%
        {Dwivedi2023JMLR}
\bibfield{author}{\bibinfo{person}{V.~P.. Dwivedi}, \bibinfo{person}{C.~K. Joshi}, \bibinfo{person}{A.~T. Luu}, \bibinfo{person}{T. Laurent}, \bibinfo{person}{Y. Bengio}, {and} \bibinfo{person}{X. Bresson}.} \bibinfo{year}{2023}\natexlab{}.
\newblock \showarticletitle{{Benchmarking Graph Neural Networks}}.
\newblock \bibinfo{journal}{\emph{J. Mach. Learning Res.}} \bibinfo{volume}{24}, \bibinfo{number}{43} (\bibinfo{year}{2023}), \bibinfo{pages}{1--48}.
\newblock


\bibitem[Ferrarello et~al\mbox{.}(2022)]%
        {Ferrarello2022DRS}
\bibfield{author}{\bibinfo{person}{L. Ferrarello}, \bibinfo{person}{A. Cavallaro}, \bibinfo{person}{R. Fiadeiro}, {and} \bibinfo{person}{R. Mazzon}.} \bibinfo{year}{2022}\natexlab{}.
\newblock \showarticletitle{Reframing the Narrative of Privacy through System-thinking Design}. In \bibinfo{booktitle}{\emph{Design Research Society Biennial Conf.}}
\newblock


\bibitem[Gilmer et~al\mbox{.}(2017)]%
        {Gilmer2017ICML}
\bibfield{author}{\bibinfo{person}{J. Gilmer}, \bibinfo{person}{S.~S. Schoenholz}, \bibinfo{person}{P.~F. Riley}, \bibinfo{person}{O. Vinyals}, {and} \bibinfo{person}{G.~E. Dahl}.} \bibinfo{year}{2017}\natexlab{}.
\newblock \showarticletitle{Neural Message Passing for Quantum Chemistry}. In \bibinfo{booktitle}{\emph{Int. Conf. Machine Learning}}.
\newblock


\bibitem[Glorot and Bengio(2010)]%
        {Glorot2010ICAIS_XavierInit}
\bibfield{author}{\bibinfo{person}{X. Glorot} {and} \bibinfo{person}{Y. Bengio}.} \bibinfo{year}{2010}\natexlab{}.
\newblock \showarticletitle{Understanding the Difficulty of Training Deep Feedforward Neural Networks}. In \bibinfo{booktitle}{\emph{Int. Conf. Artificial Intell. and Statistics}}.
\newblock


\bibitem[Han et~al\mbox{.}(2022b)]%
        {Han2022NeurIPS_ViGNN}
\bibfield{author}{\bibinfo{person}{K. Han}, \bibinfo{person}{Y. Wang}, \bibinfo{person}{J. Guo}, \bibinfo{person}{Y. Tang}, {and} \bibinfo{person}{E. Wu}.} \bibinfo{year}{2022}\natexlab{b}.
\newblock \showarticletitle{Vision {GNN}: An Image is Worth Graph of Nodes}. In \bibinfo{booktitle}{\emph{Adv. Neural Inf. Process. Syst.}}
\newblock


\bibitem[Han et~al\mbox{.}(2022a)]%
        {Han2022MTA_PrivacyMLMS}
\bibfield{author}{\bibinfo{person}{Y. Han}, \bibinfo{person}{Y. Huang}, \bibinfo{person}{L. Pan}, {and} \bibinfo{person}{Y. Zheng}.} \bibinfo{year}{2022}\natexlab{a}.
\newblock \showarticletitle{Learning Multi-Level and Multi-Scale Deep Representations for Privacy Image Classification}.
\newblock \bibinfo{journal}{\emph{Multimedia Tools Appl.}} \bibinfo{volume}{81}, \bibinfo{number}{2} (\bibinfo{year}{2022}), \bibinfo{pages}{2259–2274}.
\newblock


\bibitem[Han et~al\mbox{.}(2023)]%
        {Han2023ICCV_VHGNN}
\bibfield{author}{\bibinfo{person}{Y. Han}, \bibinfo{person}{P. Wang}, \bibinfo{person}{S. Kundu}, \bibinfo{person}{Y. Ding}, {and} \bibinfo{person}{Z. Wang}.} \bibinfo{year}{2023}\natexlab{}.
\newblock \showarticletitle{Vision HGNN: An Image is More than a Graph of Nodes}. In \bibinfo{booktitle}{\emph{Int. Conf. Comput. Vis.}}
\newblock


\bibitem[He et~al\mbox{.}(2017)]%
        {He2017ICCV_MaskRCNN}
\bibfield{author}{\bibinfo{person}{K. He}, \bibinfo{person}{G. Gkioxari}, \bibinfo{person}{P. Doll{\'{a}}r}, {and} \bibinfo{person}{R.~B. Girshick}.} \bibinfo{year}{2017}\natexlab{}.
\newblock \showarticletitle{Mask {R-CNN}}. In \bibinfo{booktitle}{\emph{Int. Conf. Comput. Vis.}} \bibinfo{address}{Venice, Italy}.
\newblock


\bibitem[He et~al\mbox{.}(2016)]%
        {He2016CVPR_ResNet}
\bibfield{author}{\bibinfo{person}{K. He}, \bibinfo{person}{X. Zhang}, \bibinfo{person}{S. Ren}, {and} \bibinfo{person}{J. Sun}.} \bibinfo{year}{2016}\natexlab{}.
\newblock \showarticletitle{Deep Residual Learning for Image Recognition}. In \bibinfo{booktitle}{\emph{Conf. Comput. Vis. Pattern Recognit.}}
\newblock


\bibitem[Jiao et~al\mbox{.}(2020)]%
        {Jiao2020BIGCOM_IEye}
\bibfield{author}{\bibinfo{person}{R. Jiao}, \bibinfo{person}{L. Zhang}, {and} \bibinfo{person}{A. Li}.} \bibinfo{year}{2020}\natexlab{}.
\newblock \showarticletitle{{IEye}: Personalized Image Privacy Detection}. In \bibinfo{booktitle}{\emph{Int. Conf. Big Data Computing and Comm.}}
\newblock


\bibitem[Kalanat and Kovashka(2022)]%
        {kalanat2022symbolic}
\bibfield{author}{\bibinfo{person}{N. Kalanat} {and} \bibinfo{person}{A. Kovashka}.} \bibinfo{year}{2022}\natexlab{}.
\newblock \bibinfo{title}{Symbolic Image Detection using Scene and Knowledge Graphs}.
\newblock
\newblock
\showeprint[arxiv]{2206.04863v1}~[cs.CV]


\bibitem[Kim et~al\mbox{.}(2017)]%
        {Kim2017ICLR}
\bibfield{author}{\bibinfo{person}{J. Kim}, \bibinfo{person}{K.~W. On}, \bibinfo{person}{W. Lim}, \bibinfo{person}{J. Kim}, \bibinfo{person}{J. Ha}, {and} \bibinfo{person}{B. Zhang}.} \bibinfo{year}{2017}\natexlab{}.
\newblock \showarticletitle{{Hadamard Product for Low-rank Bilinear Pooling}}. In \bibinfo{booktitle}{\emph{Int. Conf. on Learning Represent.}}
\newblock


\bibitem[Kingma and Ba(2015)]%
        {Kingma2015ICLR}
\bibfield{author}{\bibinfo{person}{D. Kingma} {and} \bibinfo{person}{J. Ba}.} \bibinfo{year}{2015}\natexlab{}.
\newblock \showarticletitle{{Adam: A Method for Stochastic Optimization}}. In \bibinfo{booktitle}{\emph{Int. Conf. on Learning Represent.}}
\newblock


\bibitem[Kipf and Welling(2017)]%
        {Kipf2017ICLR_GCN}
\bibfield{author}{\bibinfo{person}{T.~N. Kipf} {and} \bibinfo{person}{M. Welling}.} \bibinfo{year}{2017}\natexlab{}.
\newblock \showarticletitle{Semi-Supervised Classification with Graph Convolutional Networks}. In \bibinfo{booktitle}{\emph{Int. Conf. on Learning Represent.}}
\newblock


\bibitem[Kuznetsova et~al\mbox{.}(2020)]%
        {Kuznetsova2020IJCV_OpenImages}
\bibfield{author}{\bibinfo{person}{A. Kuznetsova}, \bibinfo{person}{H. Rom}, \bibinfo{person}{N. Alldrin}, \bibinfo{person}{J. Uijlings}, \bibinfo{person}{I. Krasin}, \bibinfo{person}{J. Pont-Tuset}, \bibinfo{person}{S. Kamali}, \bibinfo{person}{S. Popov}, \bibinfo{person}{M. Malloci}, \bibinfo{person}{A. Kolesnikov}, \bibinfo{person}{T. Duerig}, {and} \bibinfo{person}{V. Ferrari}.} \bibinfo{year}{2020}\natexlab{}.
\newblock \showarticletitle{{The Open Images Dataset V4}: Unified Image Classification, Object Detection, and Visual Relationship Detection at Scale}.
\newblock \bibinfo{journal}{\emph{Int. J. Comput. Vis.}} \bibinfo{number}{3} (\bibinfo{year}{2020}).
\newblock


\bibitem[Li et~al\mbox{.}(2023)]%
        {Li2023TPAMI_DeepGCN}
\bibfield{author}{\bibinfo{person}{G. Li}, \bibinfo{person}{M. Müller}, \bibinfo{person}{G. Qian}, \bibinfo{person}{I.~C. Delgadillo}, \bibinfo{person}{A. Abualshour}, \bibinfo{person}{A. Thabet}, {and} \bibinfo{person}{B. Ghanem}.} \bibinfo{year}{2023}\natexlab{}.
\newblock \showarticletitle{DeepGCNs: Making GCNs Go as Deep as CNNs}.
\newblock \bibinfo{journal}{\emph{IEEE Trans. Pattern Anal. Mach. Intell.}} \bibinfo{volume}{45}, \bibinfo{number}{6} (\bibinfo{year}{2023}), \bibinfo{pages}{6923--6939}.
\newblock


\bibitem[Li et~al\mbox{.}(2016)]%
        {Li2016ICLR_GGNN}
\bibfield{author}{\bibinfo{person}{Y. Li}, \bibinfo{person}{D. Tarlow}, \bibinfo{person}{M. Brockschmidt}, {and} \bibinfo{person}{R.~S. Zemel}.} \bibinfo{year}{2016}\natexlab{}.
\newblock \showarticletitle{{Gated Graph Sequence Neural Networks}}. In \bibinfo{booktitle}{\emph{Int. Conf. on Learning Represent.}}
\newblock


\bibitem[Lin et~al\mbox{.}(2018)]%
        {Lin2018ECCV_COCO}
\bibfield{author}{\bibinfo{person}{T.-Y. Lin}, \bibinfo{person}{M. Maire}, \bibinfo{person}{S. Belongie}, \bibinfo{person}{J. Hays}, \bibinfo{person}{P. Perona}, \bibinfo{person}{D. Ramanan}, \bibinfo{person}{P. Doll{\'a}r}, {and} \bibinfo{person}{C.~L. Zitnick}.} \bibinfo{year}{2018}\natexlab{}.
\newblock \showarticletitle{Microsoft {COCO}: Common Objects in Context}. In \bibinfo{booktitle}{\emph{Eur. Conf. Comput. Vis.}}
\newblock


\bibitem[Lowe(2004)]%
        {Lowe2004IJCV}
\bibfield{author}{\bibinfo{person}{D.~G. Lowe}.} \bibinfo{year}{2004}\natexlab{}.
\newblock \showarticletitle{{Distinctive Image Features from Scale-invariant Keypoints}}.
\newblock \bibinfo{journal}{\emph{Int. J. Comput. Vis.}} \bibinfo{volume}{60}, \bibinfo{number}{2} (\bibinfo{year}{2004}), \bibinfo{pages}{91--110}.
\newblock


\bibitem[Lu et~al\mbox{.}(2019)]%
        {Lu2018NeurIPS_VilBERT}
\bibfield{author}{\bibinfo{person}{J. Lu}, \bibinfo{person}{D. Batra}, \bibinfo{person}{D. Parikh}, {and} \bibinfo{person}{S. Lee}.} \bibinfo{year}{2019}\natexlab{}.
\newblock \showarticletitle{{ViLBERT}: pretraining task-agnostic visiolinguistic representations for vision-and-language tasks}. In \bibinfo{booktitle}{\emph{Adv. Neural Inf. Process. Syst.}}
\newblock


\bibitem[Marino et~al\mbox{.}(2017)]%
        {Marino2017CVPR}
\bibfield{author}{\bibinfo{person}{K. Marino}, \bibinfo{person}{R. Salakhutdinov}, {and} \bibinfo{person}{A. Gupta}.} \bibinfo{year}{2017}\natexlab{}.
\newblock \showarticletitle{The More You Know: Using Knowledge Graphs for Image Classification}. In \bibinfo{booktitle}{\emph{Conf. Comput. Vis. Pattern Recognit.}}
\newblock


\bibitem[McCallister(2010)]%
        {mccallister2010guide}
\bibfield{author}{\bibinfo{person}{E. McCallister}.} \bibinfo{year}{2010}\natexlab{}.
\newblock \bibinfo{booktitle}{\emph{Guide to protecting the confidentiality of personally identifiable information}}.
\newblock \bibinfo{publisher}{Diane Publishing}.
\newblock


\bibitem[Modas et~al\mbox{.}(2021)]%
        {Modas2021ICIP}
\bibfield{author}{\bibinfo{person}{A. Modas}, \bibinfo{person}{A. Xompero}, \bibinfo{person}{R. Sanchez-Matilla}, \bibinfo{person}{P. Frossard}, {and} \bibinfo{person}{A. Cavallaro}.} \bibinfo{year}{2021}\natexlab{}.
\newblock \showarticletitle{{Improving Filling Level Classification with Adversarial Training}}. In \bibinfo{booktitle}{\emph{IEEE Int. Conf. Image Process.}}
\newblock


\bibitem[M{\"u}ller et~al\mbox{.}(2024)]%
        {Muller2024TMLR}
\bibfield{author}{\bibinfo{person}{L. M{\"u}ller}, \bibinfo{person}{M. Galkin}, \bibinfo{person}{C. Morris}, {and} \bibinfo{person}{L. Ramp{\'a}{\v{s}}ek}.} \bibinfo{year}{2024}\natexlab{}.
\newblock \showarticletitle{{Attending to Graph Transformers}}.
\newblock \bibinfo{journal}{\emph{Trans. Mach. Learning Res.}} (\bibinfo{year}{2024}).
\newblock


\bibitem[Orekondy et~al\mbox{.}(2017)]%
        {Orekondy2017ICCV}
\bibfield{author}{\bibinfo{person}{T. Orekondy}, \bibinfo{person}{B. Schiele}, {and} \bibinfo{person}{M. Fritz}.} \bibinfo{year}{2017}\natexlab{}.
\newblock \showarticletitle{{Towards a Visual Privacy Advisor: Understanding and Predicting Privacy Risks in Images}}. In \bibinfo{booktitle}{\emph{Int. Conf. Comput. Vis.}}
\newblock


\bibitem[Ramp\'{a}\v{s}ek et~al\mbox{.}(2024)]%
        {Rampavsek2022NeurIPS_GPS}
\bibfield{author}{\bibinfo{person}{L. Ramp\'{a}\v{s}ek}, \bibinfo{person}{M. Galkin}, \bibinfo{person}{V.~P. Dwivedi}, \bibinfo{person}{A.~T. Luu}, \bibinfo{person}{G. Wolf}, {and} \bibinfo{person}{D. Beaini}.} \bibinfo{year}{2024}\natexlab{}.
\newblock \showarticletitle{Recipe for a general, powerful, scalable graph transformer}. In \bibinfo{booktitle}{\emph{Adv. Neural Inf. Process. Syst.}}
\newblock


\bibitem[Redmon and Farhadi(2018)]%
        {Redmon2018YOLOv3}
\bibfield{author}{\bibinfo{person}{J. Redmon} {and} \bibinfo{person}{A. Farhadi}.} \bibinfo{year}{2018}\natexlab{}.
\newblock \showarticletitle{{YOLOv3: An Incremental Improvement}}.
\newblock \bibinfo{journal}{\emph{CoRR}}  \bibinfo{volume}{abs/1804.02767} (\bibinfo{year}{2018}).
\newblock


\bibitem[Scarselli et~al\mbox{.}(2009)]%
        {Scarselli2009TNN}
\bibfield{author}{\bibinfo{person}{F. Scarselli}, \bibinfo{person}{M. Gori}, \bibinfo{person}{Ah~C. Tsoi}, \bibinfo{person}{M. Hagenbuchner}, {and} \bibinfo{person}{G. Monfardini}.} \bibinfo{year}{2009}\natexlab{}.
\newblock \showarticletitle{{The Graph Neural Network Model}}.
\newblock \bibinfo{journal}{\emph{IEEE Trans. Neural Networks}} \bibinfo{volume}{20}, \bibinfo{number}{1} (\bibinfo{year}{2009}), \bibinfo{pages}{61--80}.
\newblock


\bibitem[Simonyan and Zisserman(2014)]%
        {Simonyan2014ICLR_VGG}
\bibfield{author}{\bibinfo{person}{K. Simonyan} {and} \bibinfo{person}{A. Zisserman}.} \bibinfo{year}{2014}\natexlab{}.
\newblock \showarticletitle{Very Deep Convolutional Networks for Large-scale Image Recognition}. In \bibinfo{booktitle}{\emph{Int. Conf. on Learning Represent.}} \bibinfo{address}{Banff, Canada}.
\newblock


\bibitem[Squicciarini et~al\mbox{.}(2014)]%
        {Squicciarini2014ACM_HSM}
\bibfield{author}{\bibinfo{person}{A.~C. Squicciarini}, \bibinfo{person}{C. Caragea}, {and} \bibinfo{person}{R. Balakavi}.} \bibinfo{year}{2014}\natexlab{}.
\newblock \showarticletitle{Analyzing Images' Privacy for the Modern Web}. In \bibinfo{booktitle}{\emph{ACM Conf. Hypertext and Social Media}}.
\newblock


\bibitem[Srivastava et~al\mbox{.}(2014)]%
        {Srivastava2014DropoutAS}
\bibfield{author}{\bibinfo{person}{N. Srivastava}, \bibinfo{person}{G.~E. Hinton}, \bibinfo{person}{A. Krizhevsky}, \bibinfo{person}{I. Sutskever}, {and} \bibinfo{person}{R. Salakhutdinov}.} \bibinfo{year}{2014}\natexlab{}.
\newblock \showarticletitle{Dropout: A Simple Way to Prevent Neural Networks from Overfitting}.
\newblock \bibinfo{journal}{\emph{J. Mach. Learn. Res.}}  \bibinfo{volume}{15} (\bibinfo{year}{2014}), \bibinfo{pages}{1929--1958}.
\newblock


\bibitem[Stoidis and Cavallaro(2022)]%
        {Stoidis2022BigMM}
\bibfield{author}{\bibinfo{person}{D. Stoidis} {and} \bibinfo{person}{A. Cavallaro}.} \bibinfo{year}{2022}\natexlab{}.
\newblock \showarticletitle{{Content-based Graph Privacy Advisor}}. In \bibinfo{booktitle}{\emph{IEEE Int. Conf. Multimedia Big Data}}.
\newblock


\bibitem[Tan et~al\mbox{.}(2018)]%
        {TanTransferLearning}
\bibfield{author}{\bibinfo{person}{C. Tan}, \bibinfo{person}{F. Sun}, \bibinfo{person}{T. Kong}, \bibinfo{person}{W. Zhang}, \bibinfo{person}{C. Yang}, {and} \bibinfo{person}{C. Liu}.} \bibinfo{year}{2018}\natexlab{}.
\newblock \showarticletitle{A Survey on Deep Transfer Learning}. In \bibinfo{booktitle}{\emph{Int. Conf. Artificial Neural Networks}}. \bibinfo{pages}{270--279}.
\newblock


\bibitem[Tonge and Caragea(2016)]%
        {Tonge2016AAAI}
\bibfield{author}{\bibinfo{person}{A. Tonge} {and} \bibinfo{person}{C. Caragea}.} \bibinfo{year}{2016}\natexlab{}.
\newblock \showarticletitle{Image Privacy Prediction Using Deep Features}. In \bibinfo{booktitle}{\emph{AAAI Conf. Artificial Intell.}}
\newblock


\bibitem[Tonge and Caragea(2018)]%
        {Tonge2018MSM}
\bibfield{author}{\bibinfo{person}{A. Tonge} {and} \bibinfo{person}{C. Caragea}.} \bibinfo{year}{2018}\natexlab{}.
\newblock \showarticletitle{On the Use of "Deep" Features for Online Image Sharing}. In \bibinfo{booktitle}{\emph{The Web Conf., Int. Workshop Modeling Social Media}}.
\newblock


\bibitem[Tonge and Caragea(2019)]%
        {Tonge2019WWW}
\bibfield{author}{\bibinfo{person}{A. Tonge} {and} \bibinfo{person}{C. Caragea}.} \bibinfo{year}{2019}\natexlab{}.
\newblock \showarticletitle{Dynamic Deep Multi-Modal Fusion for Image Privacy Prediction}. In \bibinfo{booktitle}{\emph{{WWW}}}.
\newblock


\bibitem[Tonge and Caragea(2020)]%
        {Tonge2020TWEB}
\bibfield{author}{\bibinfo{person}{A. Tonge} {and} \bibinfo{person}{C. Caragea}.} \bibinfo{year}{2020}\natexlab{}.
\newblock \showarticletitle{Image Privacy Prediction Using Deep Neural Networks}.
\newblock \bibinfo{journal}{\emph{ACM Trans. Web}} \bibinfo{volume}{14}, \bibinfo{number}{2} (\bibinfo{date}{apr} \bibinfo{year}{2020}).
\newblock


\bibitem[Tonge et~al\mbox{.}(2018)]%
        {Tonge2018AAAI}
\bibfield{author}{\bibinfo{person}{A. Tonge}, \bibinfo{person}{C. Caragea}, {and} \bibinfo{person}{A. Squicciarini}.} \bibinfo{year}{2018}\natexlab{}.
\newblock \showarticletitle{Uncovering Scene Context for Predicting Privacy of Online Shared Images}. In \bibinfo{booktitle}{\emph{AAAI Conf. Artificial Intell.}}
\newblock


\bibitem[Tran et~al\mbox{.}(2016)]%
        {Tran2016AAAI_PCNH}
\bibfield{author}{\bibinfo{person}{L. Tran}, \bibinfo{person}{D. Kong}, \bibinfo{person}{H. Jin}, {and} \bibinfo{person}{J. Liu}.} \bibinfo{year}{2016}\natexlab{}.
\newblock \showarticletitle{{Privacy-CNH}: A Framework to Detect Photo Privacy with Convolutional Neural Network using Hierarchical Features}. In \bibinfo{booktitle}{\emph{{AAAI Conf. Artificial Intell.}}}
\newblock


\bibitem[Union(1995)]%
        {EU_DataProtection}
\bibfield{author}{\bibinfo{person}{European Union}.} \bibinfo{year}{1995}\natexlab{}.
\newblock \showarticletitle{{Directive 95/46/EC of the European Parliament and of the Council on the Protection of Individuals with Regard to the Processing of Personal Data and on the Free Movement of Such Data}}.
\newblock \bibinfo{journal}{\emph{Official Journal of the European Union L281}} (\bibinfo{date}{Nov.} \bibinfo{year}{1995}), \bibinfo{pages}{31--50}.
\newblock


\bibitem[Veličković et~al\mbox{.}(2018)]%
        {Velickovic2018ICLR_GAT}
\bibfield{author}{\bibinfo{person}{P. Veličković}, \bibinfo{person}{G. Cucurull}, \bibinfo{person}{A. Casanova}, \bibinfo{person}{A. Romero}, \bibinfo{person}{P. Liò}, {and} \bibinfo{person}{Y. Bengio}.} \bibinfo{year}{2018}\natexlab{}.
\newblock \showarticletitle{{Graph Attention Networks}}. In \bibinfo{booktitle}{\emph{Int. Conf. on Learning Represent.}}
\newblock


\bibitem[Wang et~al\mbox{.}(2018)]%
        {Wang2018IJCAI}
\bibfield{author}{\bibinfo{person}{Z. Wang}, \bibinfo{person}{T. Chen}, \bibinfo{person}{J. Ren}, \bibinfo{person}{W. Yu}, \bibinfo{person}{H. Cheng}, {and} \bibinfo{person}{L. Lin}.} \bibinfo{year}{2018}\natexlab{}.
\newblock \showarticletitle{Deep Reasoning with Knowledge Graph for Social Relationship Understanding}. In \bibinfo{booktitle}{\emph{Int. Joint Conf. Artificial Intelligence}}.
\newblock


\bibitem[Wu et~al\mbox{.}(2021)]%
        {Wu2021SurveyGNN}
\bibfield{author}{\bibinfo{person}{Z. Wu}, \bibinfo{person}{S. Pan}, \bibinfo{person}{F. Chen}, \bibinfo{person}{G. Long}, \bibinfo{person}{C. Zhang}, {and} \bibinfo{person}{P.~S. Yu}.} \bibinfo{year}{2021}\natexlab{}.
\newblock \showarticletitle{A Comprehensive Survey on Graph Neural Networks}.
\newblock \bibinfo{journal}{\emph{IEEE Tran. Neural Networks and Learning Systems}} \bibinfo{volume}{32}, \bibinfo{number}{1} (\bibinfo{year}{2021}), \bibinfo{pages}{4--24}.
\newblock


\bibitem[Xompero et~al\mbox{.}(2024)]%
        {Xompero2024CVPRW_XAI4CV}
\bibfield{author}{\bibinfo{person}{A. Xompero}, \bibinfo{person}{M. Bontonou}, \bibinfo{person}{J. Arbona}, \bibinfo{person}{E. Benetos}, {and} \bibinfo{person}{A. Cavallaro}.} \bibinfo{year}{2024}\natexlab{}.
\newblock \showarticletitle{Explaining Models Relating Objects and Privacy}. In \bibinfo{booktitle}{\emph{Conf. Comput. Vis. Pattern Recognit. Workshops}}.
\newblock
\newblock
\shownote{The 3rd Explainable AI for Computer Vision (XAI4CV) Workshop}.


\bibitem[Yang et~al\mbox{.}(2020)]%
        {Yang2020PR}
\bibfield{author}{\bibinfo{person}{G. Yang}, \bibinfo{person}{J. Cao}, \bibinfo{person}{Z. Chen}, \bibinfo{person}{J. Guo}, {and} \bibinfo{person}{J. Li}.} \bibinfo{year}{2020}\natexlab{}.
\newblock \showarticletitle{{Graph-based Neural Networks for Explainable Image Privacy Inference}}.
\newblock \bibinfo{journal}{\emph{Pattern Recognit.}}  \bibinfo{volume}{105} (\bibinfo{year}{2020}), \bibinfo{pages}{1--12}.
\newblock


\bibitem[Yang et~al\mbox{.}(2022)]%
        {Yang2022AAAI}
\bibfield{author}{\bibinfo{person}{G. Yang}, \bibinfo{person}{J. Cao}, \bibinfo{person}{Q. Sheng}, \bibinfo{person}{P. Qi}, \bibinfo{person}{X. Li}, {and} \bibinfo{person}{J. Li}.} \bibinfo{year}{2022}\natexlab{}.
\newblock \showarticletitle{{DRAG}: Dynamic Region-Aware GCN for Privacy-Leaking Image Detection}. In \bibinfo{booktitle}{\emph{AAAI Conf. Artificial Intell.}}
\newblock


\bibitem[Yu et~al\mbox{.}(2017)]%
        {Yu2017TIFS_iPrivacy}
\bibfield{author}{\bibinfo{person}{J. Yu}, \bibinfo{person}{B. Zhang}, \bibinfo{person}{Z. Kuang}, \bibinfo{person}{D. Lin}, {and} \bibinfo{person}{J. Fan}.} \bibinfo{year}{2017}\natexlab{}.
\newblock \showarticletitle{{iPrivacy}: Image Privacy Protection by Identifying Sensitive Objects via Deep Multi-Task Learning}.
\newblock \bibinfo{journal}{\emph{IEEE Tran. Information Forensics and Security}} \bibinfo{volume}{12}, \bibinfo{number}{5} (\bibinfo{year}{2017}), \bibinfo{pages}{1005--1016}.
\newblock


\bibitem[Zerr et~al\mbox{.}(2012a)]%
        {Zerr2012CIKM_PicAlert}
\bibfield{author}{\bibinfo{person}{S. Zerr}, \bibinfo{person}{S. Siersdorfer}, {and} \bibinfo{person}{J. Hare}.} \bibinfo{year}{2012}\natexlab{a}.
\newblock \showarticletitle{{PicAlert! A System for Privacy-Aware Image Classification and Retrieval}}. In \bibinfo{booktitle}{\emph{ACM Int. Conf. Information and Knowledge Management}}.
\newblock


\bibitem[Zerr et~al\mbox{.}(2012b)]%
        {Zerr2012SIGIR}
\bibfield{author}{\bibinfo{person}{S. Zerr}, \bibinfo{person}{S. Siersdorfer}, \bibinfo{person}{J. Hare}, {and} \bibinfo{person}{E. Demidova}.} \bibinfo{year}{2012}\natexlab{b}.
\newblock \showarticletitle{Privacy-Aware Image Classification and Search}. In \bibinfo{booktitle}{\emph{ACM SIGIR Int. Conf. Research and Development in Information Retrieval}}.
\newblock


\bibitem[Zhao and Caragea(2023)]%
        {Zhao2023TWEB}
\bibfield{author}{\bibinfo{person}{C. Zhao} {and} \bibinfo{person}{C. Caragea}.} \bibinfo{year}{2023}\natexlab{}.
\newblock \showarticletitle{Deep Gated Multi-Modal Fusion for Image Privacy Prediction}.
\newblock \bibinfo{journal}{\emph{ACM Trans. Web}} \bibinfo{volume}{17}, \bibinfo{number}{4} (\bibinfo{year}{2023}).
\newblock


\bibitem[Zhao et~al\mbox{.}(2022)]%
        {Zhao2022ICWSM_PrivacyAlert}
\bibfield{author}{\bibinfo{person}{C. Zhao}, \bibinfo{person}{J. Mangat}, \bibinfo{person}{S. Koujalgi}, \bibinfo{person}{A. Squicciarini}, {and} \bibinfo{person}{C. Caragea}.} \bibinfo{year}{2022}\natexlab{}.
\newblock \showarticletitle{{PrivacyAlert: A Dataset for Image Privacy Prediction}}. In \bibinfo{booktitle}{\emph{Int. AAAI Conf. Web and Social Media}}.
\newblock


\bibitem[Zhong et~al\mbox{.}(2019)]%
        {Zhong2019BigData_RPM}
\bibfield{author}{\bibinfo{person}{H. Zhong}, \bibinfo{person}{H. Li}, \bibinfo{person}{A. Squicciarini}, \bibinfo{person}{S. Rajtmajer}, {and} \bibinfo{person}{D. Miller}.} \bibinfo{year}{2019}\natexlab{}.
\newblock \showarticletitle{Toward Image Privacy Classification and Spatial Attribution of Private Content}. In \bibinfo{booktitle}{\emph{IEEE Int. Conf. Big Data}}.
\newblock


\bibitem[Zhou et~al\mbox{.}(2018)]%
        {Zhou2018TPAMI_Places365}
\bibfield{author}{\bibinfo{person}{B. Zhou}, \bibinfo{person}{A. Lapedriza}, \bibinfo{person}{A. Khosla}, \bibinfo{person}{A. Oliva}, {and} \bibinfo{person}{A. Torralba}.} \bibinfo{year}{2018}\natexlab{}.
\newblock \showarticletitle{Places: A 10 million Image Database for Scene Recognition}.
\newblock \bibinfo{journal}{\emph{IEEE Trans. Pattern Anal. Mach. Intell.}} \bibinfo{volume}{40}, \bibinfo{number}{6} (\bibinfo{year}{2018}), \bibinfo{pages}{1452--1464}.
\newblock


\bibitem[Zhou et~al\mbox{.}(2017)]%
        {Zhou2017Vision_Places}
\bibfield{author}{\bibinfo{person}{B. Zhou}, \bibinfo{person}{A. Lapedriza}, \bibinfo{person}{A. Torralba}, {and} \bibinfo{person}{A. Oliva}.} \bibinfo{year}{2017}\natexlab{}.
\newblock \showarticletitle{{Places: An Image Database for Deep Scene Understanding}}.
\newblock \bibinfo{journal}{\emph{J. Vision}} \bibinfo{volume}{17}, \bibinfo{number}{10} (\bibinfo{year}{2017}).
\newblock


\end{thebibliography}

\appendix

\section{Annotation procedures and inconsistencies of image privacy datasets}
\label{app:labelinconsistencies}

We present the procedures used to annotate the image privacy datasets considered in our experiments, such as PrivacyAlert~\cite{Zhao2022ICWSM_PrivacyAlert} and IPD~\cite{Yang2020PR}, and discuss existing labelling ambiguities or inconsistencies. As IPD~\cite{Yang2020PR} combines two datasets, we also discuss PicAlert~\cite{Zerr2012CIKM_PicAlert} and VISPR~\cite{Orekondy2017ICCV}. All these datasets contain images that were collected from social networks, such as Flickr or Twitter, mostly considering the Public Domain Dedication or Public Domain Mark license. Table~\ref{tab:datasetsprivacy} compares the source images, annotation procedure including the initial labelling, and the final labels and number of images between the four datasets.

For \textit{PrivacyAlert}~\cite{Zhao2022ICWSM_PrivacyAlert}, the images were selected by using keywords belonging to a privacy taxonomy of 10 categories: \textit{nudity or sexual, other people, unorganized home, violence, medical, drinking or party, appearance or facial expression, bad character or unlawful criminal, religion or culture, and personal information}. A set of 20,000 images randomly sampled from a larger pool of images and evenly distributed across the 10 categories were manually labelled by annotators hired on Amazon Mechanical Turk. Specifically, 10,000 images (associated to the training set) were annotated by 3 annotators and the other 10,000 images (associated with the validation and training sets) were all annotated by 5 annotators. Note that the guideline used to annotate the images (see Table~\ref{tab:datasetsprivacy}) biases an annotator towards a subjective interpretation of the images as if they belong to the annotator's collection or personal sphere. 
A quality check mechanism was also adopted by providing a random sample of public and private image pairs previously annotated by trained students on image privacy prediction. Only a smaller portion of the 20,000 images have been made publicly available to provide a dataset that is not highly imbalanced towards the public class (reduced from a ratio of 9:1 to 3:1). 
Images, initially annotated with four labels, were converted into binary labels when releasing the dataset, whereas multi-class labels and the 10 privacy categories are not publicly available to perform fine-grained analyses.

\textit{PicAlert}~\cite{Zerr2012CIKM_PicAlert,Zerr2012SIGIR} contains images collected from Flickr based on a 4-month window between January and April 2010. For each step of a gamification process, 81 users, including graduate computer science students and users of social media platforms (aged between 10 and 59 years old) were instructed to annotate five photos as either public, private, or undecidable (see annotation guideline in Table~\ref{tab:datasetsprivacy}).
Available images are all annotated by at least two users with an agreement of the 75\% in the label. Note that the number of annotations for each image can largely vary in the dataset, and about 70\% of these images are labelled as public or undecidable~\cite{Zerr2012SIGIR}. However, a large number of images in the dataset contain people~\cite{Yang2020PR} and most of the images labelled as private contain people (63\%)~\cite{Stoidis2022BigMM}.

\begin{table}[t!]
    \centering
    \footnotesize
    \caption{The VISPR taxonomy of 67 privacy attributes~\cite{Orekondy2017ICCV}. 
    \vspace{-10pt}
    }
    \begin{tabular}{ll}
    \toprule
    \textbf{Category} & \textbf{Private attributes} \\
    \midrule
    \parbox{1.7cm}{Personal\\description (24)} & \parbox{6cm}{\textit{{\color{dc2}fingerprint, signature, complete nudity, full name, first name, last name, place of birth, date of birth, handwriting}, gender,\\eye colour, hair colour, complete face, partial face, tattoo,\\partial nudity, race, skin colour, traditional clothing, nationality, marital status, age group, weight group, height group}} \\
    \midrule
    Documents (8) & \parbox{6cm}{\textit{{\color{dc2}national identification, credit card, passport, driver license,\\student ID, mail, receipts, tickets}}}\\
    \midrule
    Health (3) & \parbox{6cm}{\textit{{\color{dc2}medical history}, physical disability, medical treatment}}\\
    \midrule
    Employment (2) & \parbox{6cm}{\textit{occupation, work occasion}} \\
    \midrule
    Personal life (10) & \parbox{6cm}{\textit{{\color{dc2}personal occasion, legal involvement}, religion, culture, hobbies,\\sexual orientation, sports, education history, general opinion, political opinion}} \\
    \midrule
    Relationships (6) & \parbox{6cm}{\textit{{\color{dc2}personal relationships}, social circle, professional circle,\\competitors, spectators, similar view}} \\
    \midrule
    Whereabouts (7) & \parbox{6cm}{\textit{{\color{dc2}visited landmark or street sign, complete visited location,\\partial visited location, complete home address,\\partial home address, date/time of activity}, phone number}} \\
    \midrule
    Internet activity (4) & \parbox{6cm}{\textit{{\color{dc2}email address, email content, online conversations}, username}} \\ 
    \midrule
    Automobile (3) & \parbox{6cm}{\textit{{\color{dc2}complete license plate, partial license plate}, vehicle ownership}} \\
    \bottomrule \addlinespace[\belowrulesep]
    \multicolumn{2}{l}{\parbox{0.98\linewidth}{\scriptsize{The privacy attributes are based on public guidelines for handling Personally Identifiable Information~\cite{mccallister2010guide}, namely the EU Data Protection Directive 95/46/EC~\cite{EU_DataProtection}, US Privacy Act of 1974, and community guidelines available in social networks such as Flickr and Twitter. In brackets the number of attributes for each category. Highlighted in {\color{dc2}green} the 32 privacy attributes used by Yang et al.~\cite{Yang2020PR} to select the VISPR images when composing IPD.}}}
    \end{tabular}    
    \label{tab:visprlabels}
    \vspace{-10pt}
\end{table}

\textit{VISPR}~\cite{Orekondy2017ICCV} sampled an initial pool of 100,000 images from OpenImages~\cite{Kuznetsova2020IJCV_OpenImages} and three annotators were instructed to label independent sets of images by selecting among a predefined list of privacy attributes or with an unsure or safe label in case none of the attributes could have applied (see Table~\ref{tab:datasetsprivacy} and Table~\ref{tab:visprlabels}). After this first labelling step, all images marked as unsure, those containing copyright watermark, those representing a historic photograph, those captured with a poor quality, or those containing non-English text were discarded, resulting in a dataset of 10,000 images. Other images were added to handle the class imbalance. 
VISPR is annotated with multiple labels for each image (see Table~\ref{tab:datasetsprivacy}). 
To control potential mistakes done by the annotators during the first labelling steps, a curation step was included by analysing batches of images associated with each label and manually re-annotating batches that had less than 500 images or were considered sensitive by the users. 

\textit{IPD}~\cite{Yang2020PR} combines PicAlert with a subset of private images from VISPR to reduce the highly imbalance distribution towards images labelled as public in PicAlert. Yang et al.~\cite{Yang2020PR} observed that all images of people are annotated as private in VISPR, but the dataset provides more images of private objects compared to PicAlert. Therefore, images were sampled from VISPR by considering 32 of the 67 private attributes (see Table~\ref{tab:visprlabels}).
IPD preserves the annotations for PicAlert, whereas multi-labels of VISPR are converted into a single-label annotation (private).

\begin{figure*}[t!]
    \centering
    \includegraphics[height=0.11\linewidth]{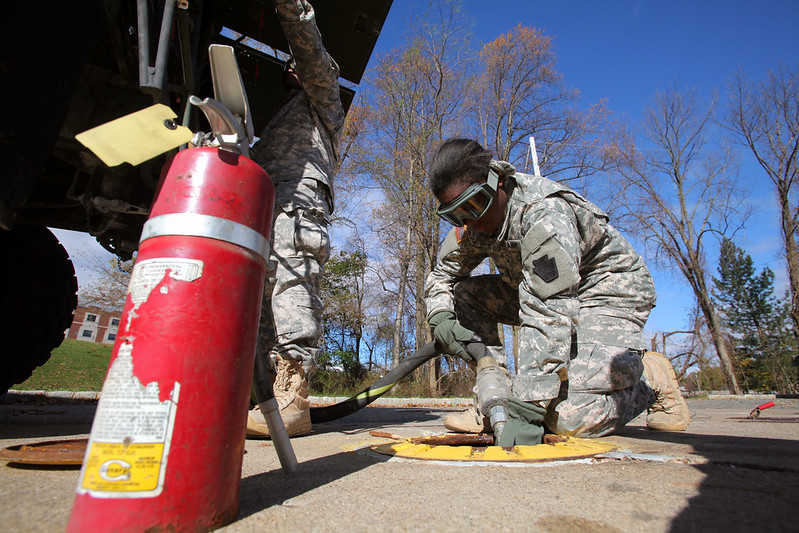} % private
    \includegraphics[height=0.11\linewidth]{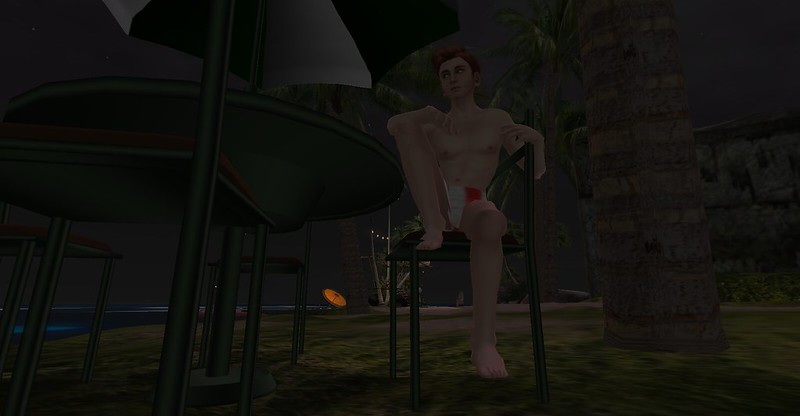} % private
    \includegraphics[height=0.11\linewidth]{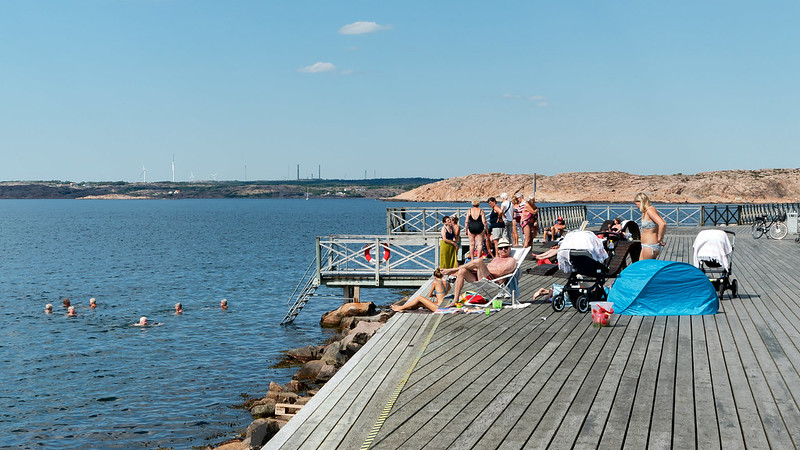} % private
    \includegraphics[height=0.11\linewidth]{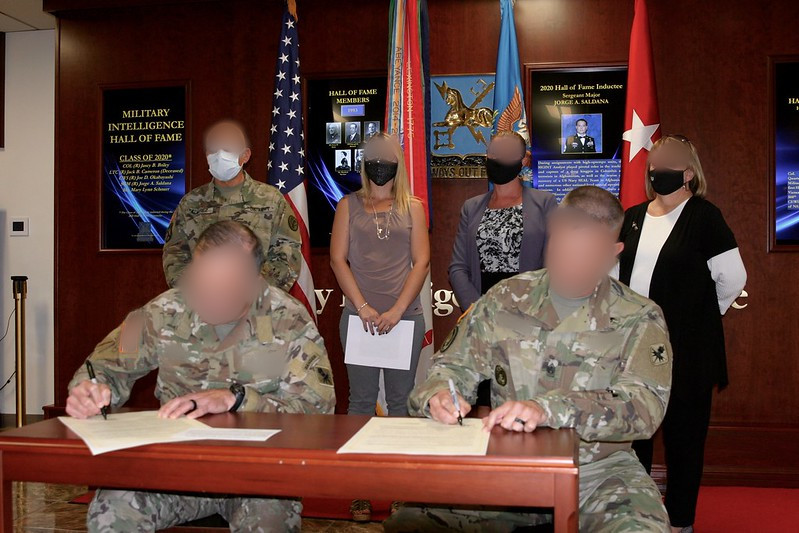} % private
    \includegraphics[height=0.11\linewidth]{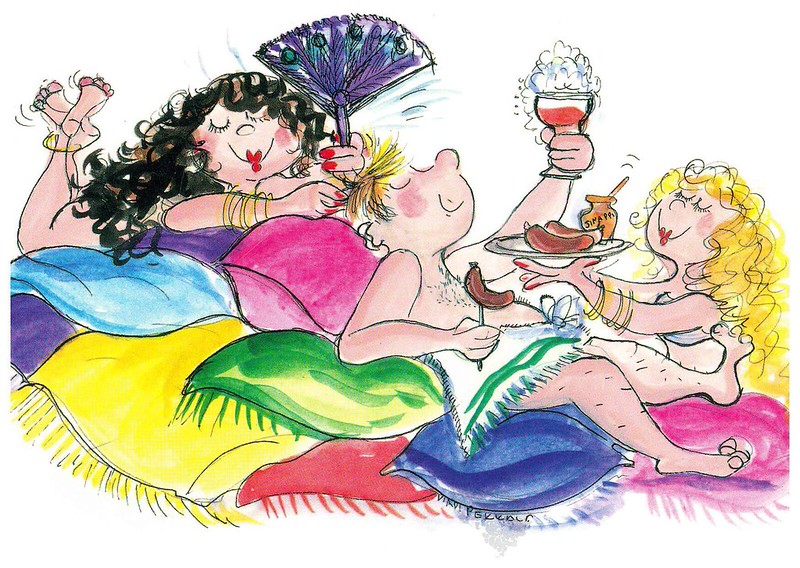} % private 
    \\
    \includegraphics[height=0.11\linewidth]{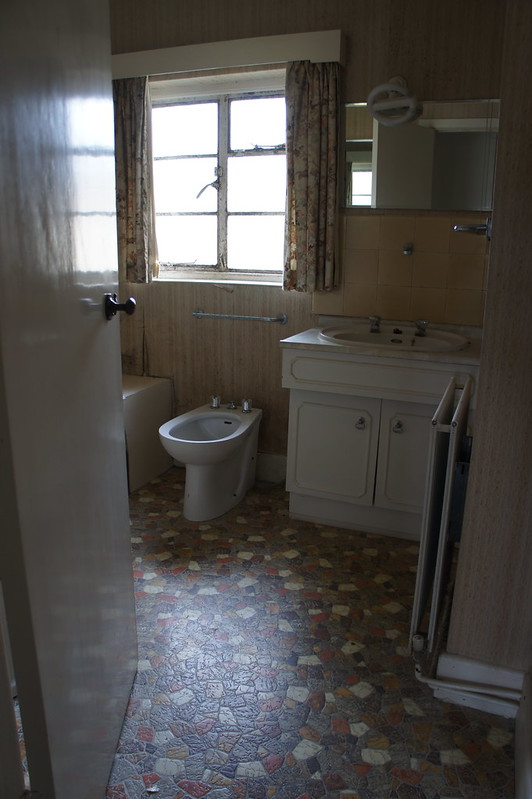} % public
    \includegraphics[height=0.11\linewidth]{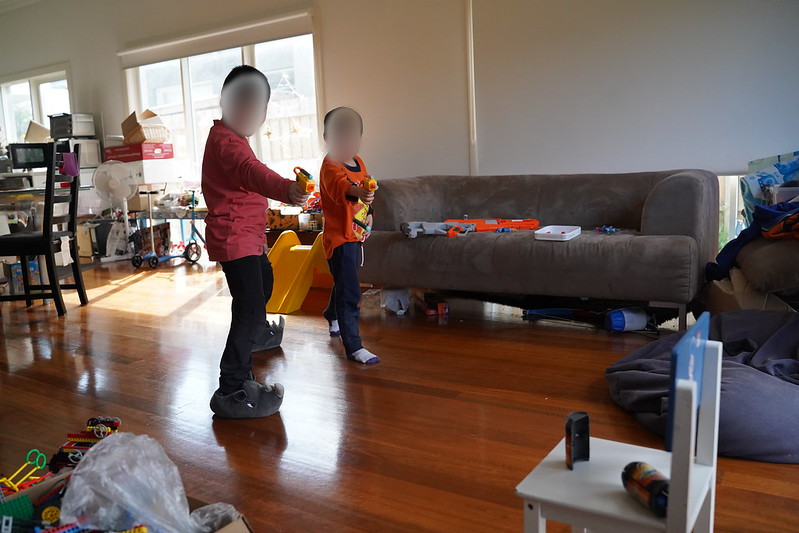} % public
    \includegraphics[height=0.11\linewidth]{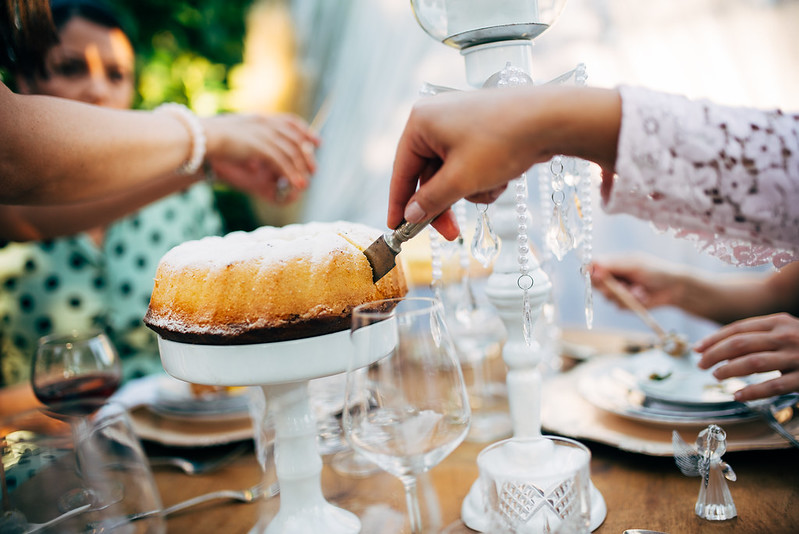} % public   
    \includegraphics[height=0.11\linewidth]{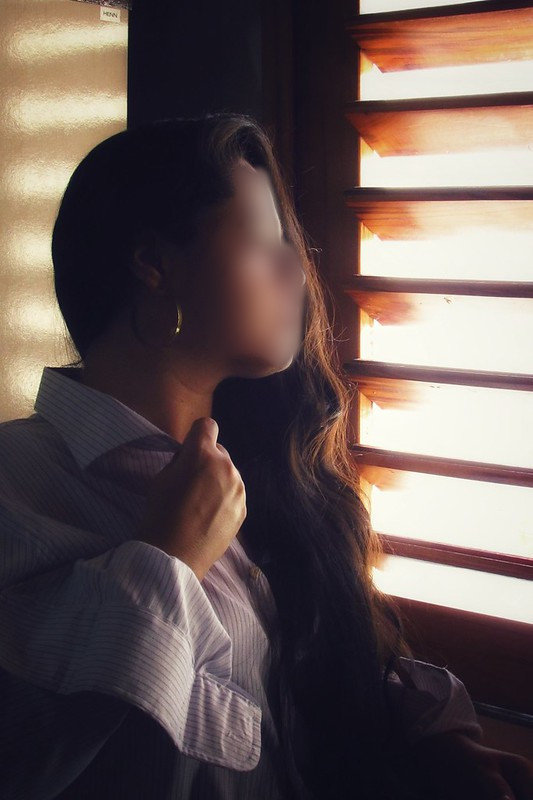} % public
    \includegraphics[height=0.11\linewidth]{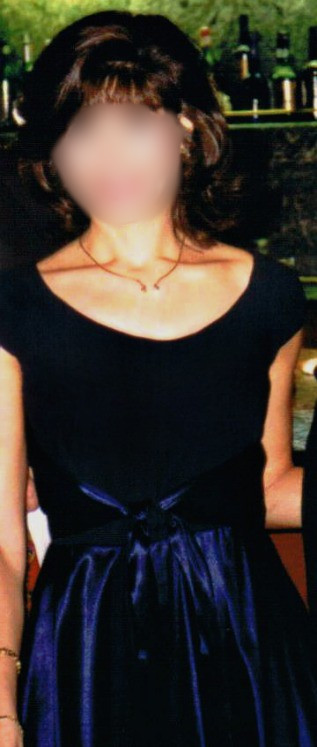} % public
    \includegraphics[height=0.11\linewidth]{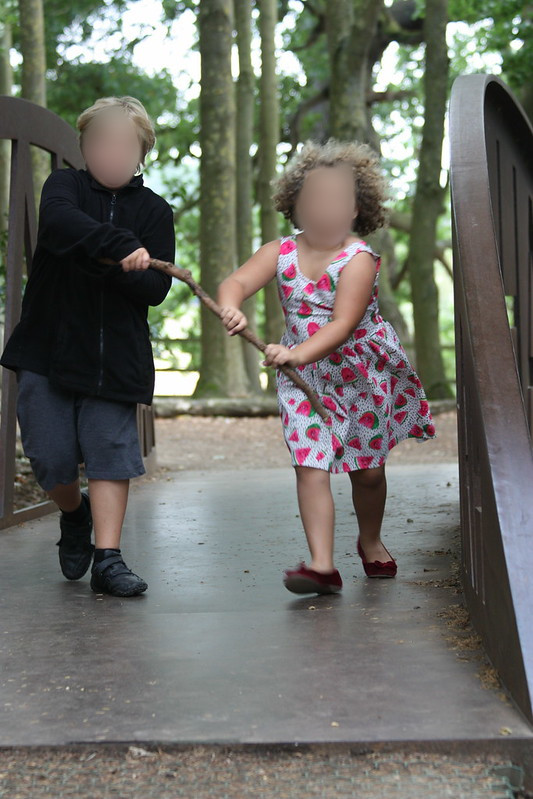} % public
    \includegraphics[height=0.11\linewidth]{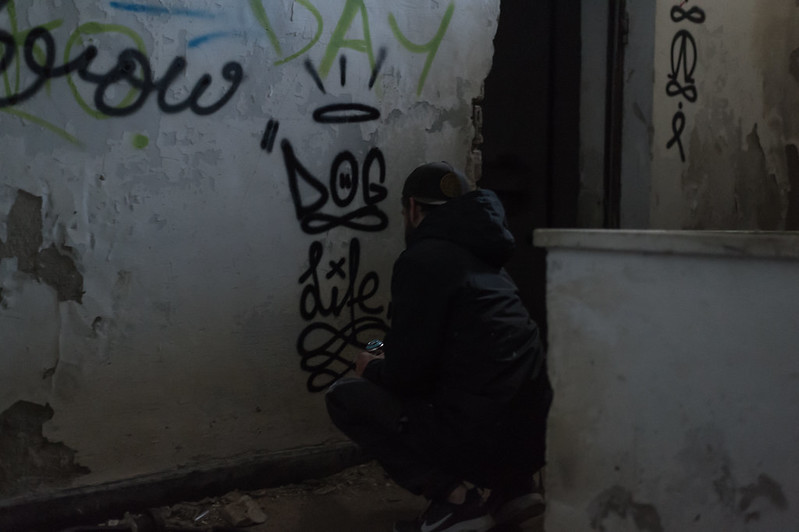} % public
    \caption{Sample of PrivacyAlert images whose annotation could be incorrect or ambiguous. First row: images are labelled as private. Second row: images are labelled as public. Obfuscation added to visible people's faces.}
    \label{fig:labelambiguities}
    \vspace{-10pt}
\end{figure*}

\noindent{\bf Inconsistencies.} Authors of PrivacyAlert performed various analyses of the dataset, including inter-annotation agreement, label distribution across the categories, and identification of potentially mislabelled or ambiguous images using the Area Under the Margin metrics of deep learning models' predictions. From these analyses, they found that annotators were not always consistent in their labelling due to being inattentive or inaccurate, and visually similar images does not always share the same label or tags~\cite{Zhao2022ICWSM_PrivacyAlert}. Despite these observations, the available dataset still contains images with labels that can be subjectively interpreted as incorrect or ambiguous when looking at the images. Fig.~\ref{fig:labelambiguities} shows a sample of these images and their manual annotations. For example, images with children are usually expected to be private as a sensitive piece of information, however both images depicting children are labelled as public. For IPD, labels of images that do not contain people can be annotated inconsistently because PicAlert and VISPR were annotated with different procedures and instructions. VISPR discarded images containing copyright watermark, representing a historic photograph, captured with a poor quality, and containing non-English text, and images of people were labelled as private. Contrarily, PicAlert and PrivacyAlert can contain those images and their labelling could be either private or public depending on the type of quality control and instructions provided to the annotators. Resolving these inconsistencies is beyond the scope of this work and results and conclusions are limited to the current status of the datasets. 
For example, models trained on these datasets fit the incorrect or inconsistent labels and hence could predict images in an ambiguous way compared to how a user would expect.

\section{Comparisons with other prior works}
\label{app:extracomparison}

In this appendix, we compare our proposed model, S2P, with 5 multimodal models (Sec.~\ref{app:multimodal}), and 4 similar existing works using transfer learning, CNN methods, and other classifiers (Sec.~\ref{app:compsvm}). We compare and discuss the results in Sec.~\ref{app:comparisonresults}.

\subsection{Multimodal models}
\label{app:multimodal}

We consider the multimodal models evaluated by Zhao et al.~\cite{Zhao2022ICWSM_PrivacyAlert} on PrivacyAlert: Privacy-CNH (PCNH)~\cite{Tran2016AAAI_PCNH}, Concat~\cite{Tonge2018AAAI}, Dynamic Multimodal Fusion for Privacy prediction (DMFP)~\cite{Tonge2019WWW}, VilBERT fine-tuned for image privacy (P-VilBERT)~\cite{Lu2018NeurIPS_VilBERT,Zhao2022ICWSM_PrivacyAlert}, and Gated Multi-Modal Fusion (GMMF)~\cite{Zhao2022ICWSM_PrivacyAlert,Zhao2023TWEB}. Note that these models take multiple inputs from different modalities, such as text and images, whereas S2P is a unimodal model that uses a single image as input.

\textit{PCNH}~\cite{Tran2016AAAI_PCNH} is a two-branch network that takes a single image as input and combines the features extracted by the two branches with a late fusion mechanism to predict whether an image is private. One branch is based on AlexNet pre-trained on ImageNet and this components is trained with transfer learning by replacing the last FC layer with an ad-hoc number of classes (from 1,000 to 204), and then is kept fixed during the second phase of the full model's training. This branch enables the model to automatically predict objects in the image. The second branch has 2 convolutional layers and 3 FC layers to transform the image into a 24 dimensional feature vector. This branch learns to extract privacy-specific features. The second branch, the feedforward network (3 FC layers that transform the 204 object features and align with the 24 features of the second branch), and the last FC layer that fuses the features concatenated from the two branches are all trained together on the image privacy dataset (PicAlert~\cite{Zerr2012CIKM_PicAlert}). PCNH is not strictly speaking a multimodal approach as defined in this work, however we report the performance of this model as presented by Zhao et al.~\cite{Zhao2022ICWSM_PrivacyAlert} and for the discussion. 

\textit{Concat}~\cite{Tonge2018AAAI} concatenates features extracted by an object recognition model and a scene recognition model, and user tags, and trains a Support Vector Machine (SVM) as a privacy classifier. To extract visual features, this method uses CNNs (e.g., AlexNet) pre-trained on ImageNet~\cite{Deng2009CVPR_ImageNet} for object recognition and Places365~\cite{Zhou2017Vision_Places,Zhou2018TPAMI_Places365} for scene recognition and selects the top-$k$ classes for each type (object and scene) to construct binary vectors, denoted as object tags and scene tags\footnote{Tonge et al.~\cite{Tonge2018AAAI} reported to have used $k=10$ in their experiments for consistency with the previous work on only object tags~\cite{Tonge2016AAAI} and showed the best results compared to using only $k=2$.}. Therefore, for a given image, the binary vector for object tags has a dimensionality of 1,000 elements with only $k$ objects enabled with a value of 1. Similarly, the binary vector for scene tags has a dimensionality of 365 with only $k$ scenes enabled with a value of 1. Additionally, another binary vector can be constructed with the list of tags generated by the users in the dataset and enabling only the tags identified by the user for the specific image. The concatenation of these binary vectors results in sparse vectors that aim at being very discriminative for the SVM classifier. 

\textit{DMFP}~\cite{Tonge2019WWW} fuses the predictions of a set of classifiers specialised on predicting privacy for each modality, such as objects, scenes, and user tags. Unlike Concat, features related to objects and scenes are extracted from the last FC layer of the pre-trained CNNs and are not converted into binary vectors. The predictions of the modality-specific privacy classifiers are fused with a weighted majority voting strategy that uses the ``competence'' of each classifier (weight) predicted by a binary classifier trained on a set of visually similar images (dynamic ensemble approach).

\textit{P-VilBERT}~\cite{Zhao2022ICWSM_PrivacyAlert} is a privacy classifier based on a pre-trained vision-language model, VilBERT~\cite{Lu2018NeurIPS_VilBERT}, that is fine-tuned for image privacy classification using images and tags as inputs. This model has a two-stream architecture, one for visual input and another for the text inputs, and each consisting of attention-based bidirectional encoding blocks (Bidirectional Encoder Representations from Transformers, or BERT), followed by fusion blocks through co-attentional transformer layers. The fusion blocks enable the model to exchange and align information extracted from each modality-specific stream. 

\textit{GMMF}~\cite{Zhao2022ICWSM_PrivacyAlert,Zhao2023TWEB} is a decision-level multimodal fusion model that uses a learnable gating network to weight the predictions of single-modality classifiers. As Concat and DMFP, GMMF uses pre-trained CNN models for object recognition and scene recognition. GMMF also uses the pre-trained BERT model to obtain a feature vector from the input image tags (user generated or extracted by a CNN). The three models are fine-tuned on PicAlert~\cite{Zerr2012CIKM_PicAlert} to compute the probability distribution over the public and private classes. The gating fusion network learns to weigh the predictions from each modality and uses the weights to compute a weighted average fusion of the privacy predictions of each modality, giving dynamically higher importance to the stronger modalities for the input image. 

\subsection{Other  works}
\label{app:compsvm}

The additional methods for comparison are: Image Tags~\cite{Tonge2016AAAI}, Scene Tags~\cite{Tonge2016AAAI}, a scene classifier coupled with an SVM classifier~\cite{Zhao2022ICWSM_PrivacyAlert}, and a fine-tuning of the scene classifier~\cite{Zhao2022ICWSM_PrivacyAlert}. We refer the reader to Tonge and Caragea's work~\cite{Tonge2020TWEB} for a more in-depth analysis of different classifiers, such as Naive Bayes, Logistic Regression, Random Forest, and SVM, with features extracted by pre-trained CNNs, a comparison of different CNNs (AlexNet, GoogleNet, VGG-16, and ResNet), and a comparison of fine-tuned CNNs with pre-trained CNNs. For our analysis, we only consider SVM as an alternative classifier and ResNet as a CNN because Tonge and Caragea's work~\cite{Tonge2020TWEB} showed to be the best choices in terms of classification performance on the PicAlert dataset that was used for the experiments. We also compare S2P with 2 variants that replace the FC layer of S2P with a 1-layer MLP and 2-layer MLP, respectively, where 1-layer means the number of hidden layers. 

\begin{table*}[t!]
    \centering
    \footnotesize
    \setlength\tabcolsep{4.5pt}
    \caption{Image privacy classification results and comparisons of multimodal models on the testing sets of IPD and PrivacyAlert. 
    \vspace{-10pt}
    }
    \begin{tabular}{ll ccccc cc ccc ccc ccc}
        \toprule
        \multicolumn{1}{l}{\textbf{Dataset}} &  \multicolumn{1}{l}{\textbf{Method}} & 
        \multicolumn{5}{c}{\textbf{Modalities}} & \multicolumn{2}{c}{\textbf{Training}} & \multicolumn{3}{c}{Private} & \multicolumn{3}{c}{Public} & \multicolumn{3}{c}{Overall}  \\
        \cmidrule(lr){3-7}\cmidrule(lr){8-9}\cmidrule(lr){10-12}\cmidrule(lr){13-15}\cmidrule(lr){16-18}
        & & Obj. & Scenes & Tags & V & L & TL & FT & P & R & F1 & P & R & F1 & P & BA & ACC \\
        \midrule
        \multirow{9}{*}{PrivacyAlert~\cite{Zhao2022ICWSM_PrivacyAlert}}
        & All private & -- & -- & -- & -- & -- & -- & -- & 25.00 & 100.00 & 40.00 & 0.00 & 0.00 & 0.00  & 12.50 & 50.00 & 25.00 \\ 
        & All public & -- & -- & -- & -- & -- & -- & -- & 0.00 & 0.00 & 0.00 & 75.00 & 100.00 & 85.71  & 37.50 & 50.00 & 75.00 \\ 
        & Random & -- & -- & -- & -- & -- & -- & -- & 74.27 & 50.67 & 60.24 & 24.23 & 47.33 & 32.05 &  49.25 & 49.00 & 49.83 \\ 
        \cmidrule(lr){2-18}
        & *PCNH~\cite{Tran2016AAAI_PCNH} & \bbox & \wbox & \wbox & \bbox & \wbox & \bbox & \wbox & 70.60 & 51.10 & 59.30 & 85.10 & 92.90 & 88.80 & 77.85 & 72.00 & 83.17 \\
        & *Concat~\cite{Tonge2018AAAI} & \bbox & \bbox & \bbox & \bbox & \wbox & \wbox & \wbox & 62.60 & 71.60 & 66.80 & 90.00 & 85.80 & 87.90 & 76.30 & 78.70 & 82.22 \\
        & *DMFP~\cite{Tonge2019WWW} & \bbox & \bbox & \bbox & \bbox & \wbox & \bbox & \wbox & 66.60 & 65.60 & 66.10 & 88.60 & 89.00 & 88.80 & 77.60 & 77.30 & 83.17 \\
        & *P-VilBERT~\cite{Zhao2022ICWSM_PrivacyAlert,Lu2018NeurIPS_VilBERT} & \wbox & \wbox & \bbox & \bbox & \bbox & \bbox & \bbox & 65.80 & 69.70 & 67.70 & 89.70 & 87.90 & 88.80 & 77.75 & 78.80 & 83.37 \\
        & *GMMF~\cite{Zhao2022ICWSM_PrivacyAlert,Zhao2023TWEB} & \bbox & \bbox & \bbox & \bbox & \wbox & \bbox & \bbox & 77.90 & 72.20 & 75.00 & 91.00 & 93.20 & 92.10 & 84.45 & \textbf{82.70} & \textbf{87.94} \\
        \cmidrule(lr){2-18}
         \rowcolor{mylightgray} \cellcolor{white} & S2P & \wbox & \bbox & \wbox & \bbox & \wbox & \bbox & \wbox & 63.11 & 63.11 &  63.11 & 87.67 &  87.67 & 87.67 & 75.39 & 75.39 & 81.51 \\
      \bottomrule \addlinespace[\belowrulesep]
      \multicolumn{18}{l}{\parbox{0.95\linewidth}{\scriptsize{
      *Results are taken from Zhao et al.'s work on PrivacyAlert~\cite{Zhao2022ICWSM_PrivacyAlert}. As some of the images of the dataset are no longer available, performance may be higher for these methods. Unlike Zhao et al.'s work that computes the weighted average precision and recall (BA) across the two classes, giving higher emphasis to the public class that has a higher number of samples, we computed the macro averaging (unweighted mean), treating the two classes equally. 
      \\
      KEY -- Obj.:~objects, V:~vision, L:~language, TL:~transfer learning, FT:~fine-tuning, P:~precision, R:~recall, ACC:~accuracy, BA:~Balanced accuracy (corresponds to overall recall).}}}
    \end{tabular}
    \label{tab:multimodalcomparison}
    \vspace{-10pt}
\end{table*}

\textit{Image Tags} uses two CNNs, one pre-trained on ImageNet for object recognition (ResNet-101 with 1,000 output classes) and another pre-trained on Places365 for scene recognition (ResNet-50 with 365 output classes) and converts the outputs of each CNN into a binary vector by selecting the top-$k$ classes with the highest probability. Selected classes are denoted as object tags and scene tags, respectively. The resulting vector has a dimensionality of 1,365 with $2k$ elements set to 1. Following the parameter setting of Tonge et al.~\cite{Tonge2018AAAI}, we set $k=10$. The image tags are then provided as input to an SVM trained for image privacy prediction. We count the number of parameters of the pre-trained CNNs, totalling 68,805,077 of trainable parameters. None of the CNN parameters are optimised specifically for privacy. \textit{Scene Tags} is a variant of Image Tags considering only the CNN trained for scene classification and totalling 24,255,917 trainable parameters. The third method (\textit{ResNet-50 + SVM}) is also based on the pre-trained CNN for scene classification and provides the features extracted from the last FC layer predicted for the 365 classes as input to a linear SVM classifier instead of selecting the scene tags. The fourth method (\textit{ResNet-50-FT}) uses the same CNN as previous methods and transfer learning as a training strategy. The last fully connect layer is replaced with a FC layer that maps the 2,048 features into the two logits for the public and private classes instead of the 365 classes for scene recognition. The parameters of the layer are initialised from a Xavier's uniform distribution~\cite{Glorot2010ICAIS_XavierInit}, whereas the parameters of all other layers are pre-trained on Places365. During training, all model parameters (23,512,130) are fine-tuned for image privacy.

For the S2P variants, the MLP is designed to halve the number of features for each hidden layer and the last FC transforms the features into the logits of the two classes. 
All the MLP parameters are initialised from a Xavier's uniform distribution~\cite{Glorot2010ICAIS_XavierInit}.  

For the three methods using the SVM as a privacy classifier, we use the 5-fold stratified cross-validation strategy on the combination of the training and validation sets to select the hyper-parameters and train the SVM. For ResNet-50 + SVM on IPD, the best regularisation parameter (C in scikit-learn) is 0.5, and the training is performed with balancing the classes, whereas C is set to 0.1 on PrivacyAlert. Note that features are not normalised as we observed that standardisation led to comparable classification performance to not using it. For Scene Tags on IPD, C is set 0.5 and the classifier is trained with balanced weights across the two classes.

\subsection{Results and discussion}
\label{app:comparisonresults}

We assess how S2P ranks with respect to the multi-modal models even if using an additional input modality makes the comparison unfair as multimodal models are expected to perform better than unimodal models but this is not guaranteed in practice. As source codes of the multimodal models are not publicly available, we simply report the results on PrivacyAlert from Zhao et al.~\cite{Zhao2022ICWSM_PrivacyAlert}, whereas results on IPD are not available. Table~\ref{tab:multimodalcomparison} shows that the multimodal baselines evaluated by Zhao et al~\cite{Zhao2022ICWSM_PrivacyAlert}, except for PCNH~\cite{Tran2016AAAI_PCNH}, achieve higher classification performance than S2P, especially in terms of recall in the private class, balanced accuracy, and accuracy. These results are taken directly from the evaluation done by Zhao et al~\cite{Zhao2022ICWSM_PrivacyAlert} and a fully fair comparison is not possible due to some missing images in the dataset. The dynamic gating fusion of GMMF with privacy predictions from objects, scenes, and user tags achieves the best performance with recall at 72.20\% on the private class, 82.70\% for balanced accuracy, and 87.94\% for accuracy. This shows that fine-tuning multiple models for privacy prediction and then fusing this information in a learnable way is the most effective solution so far, especially outperforming other fusion strategies as Concat and DMFP. The vision-language model, P-VilBERT, fine-tuned for image privacy classification, obtains classification performance comparable to DMFP and hence lower than GMMF. Concat, DMFP, P-VilBERT, and GMMF especially achieves a higher recall on the private class  (71.60\%, 65.60\%, 69.70\%, and 72.20\%, respectively) than the recall of S2P (63.11\%). The best performance of GMMF is also given by the higher recall on the public class at 93.20\%. On the contrary, PCNH is a unimodal model based only on the input image as S2P, but the classification performance are lower than S2P despite having a branch whose parameters are specifically learned for privacy prediction. This shows that simply relating scenes to privacy with an FC layer is more effective than designing a more complex architecture.

\begin{table*}[t!]
    \centering
    \footnotesize
    \setlength\tabcolsep{6.5pt}
    \caption{Results and comparisons of S2P (ResNet50  + FC) with prior methods using different classifier, training strategy, and input features on the testing set of the two image datasets for image privacy classification. Note also the two variants of S2P with the fully connected layer replaced by an MLP with 1 and 2 hidden layers, respectively.
    }
    \vspace{-10pt}
    \begin{tabular}{ll ccc ccc ccc  cc }
        \toprule
        \multicolumn{1}{l}{\textbf{Dataset}} & \multicolumn{1}{l}{\textbf{Method}} & \multicolumn{3}{c}{Private} & \multicolumn{3}{c}{Public} & \multicolumn{3}{c}{Overall} & \multicolumn{2}{c}{\textbf{\# of trainable parameters}$^\diamond$}  \\
        \cmidrule(lr){3-5}\cmidrule(lr){6-8}\cmidrule(lr){9-11}\cmidrule(lr){12-13}
        & & P & R & F1 & P & R & F1 & P & BA & ACC & Optimised & Total\\
        \midrule
        \multirow{5}{*}{IPD~\cite{Yang2020PR}}
        & Image Tags + SVM & 69.53 & 81.03 & 74.84 & 89.66 & 82.25 & 85.80 & 79.60 & 81.64 & 81.84 & 0 & 68,805,077 \\
        & Top-10 Scene Tags + SVM & 66.93 & 76.43 & 71.37 & 87.32 & 81.12 & 84.10 & 77.12 & 78.78 & 79.56 & 0 & 24,255,917 \\
        & ResNet50 + SVM & 72.74 & \textbf{81.42} & 76.84 & 90.12 & 84.74 & 87.35 & 81.43 & \textbf{83.08} & 83.64 & 0 & 24,255,917 \\
        & ResNet50-FT & 69.93 & 70.36 & 70.14 & 85.13 & 84.87 & 85.00 & 77.53 & 77.62 & 80.03 & 23,512,130 & 23,512,130 \\ 
        \rowcolor{mylightgray} \cellcolor{white} & ResNet50 + FC & 75.83 & 72.44 & 74.10 & 86.52 & 88.45 & 87.48 & 81.18 & 80.45 & 83.12 & 732 & 24,256,649 \\
        & ResNet50 + MLP (1) & 75.27 & 76.74 & 75.99 & 88.25 & 87.39 & 87.82 & 81.76 & 82.06 & \textbf{83.84} & 66,978 & 24,322,895 \\ 
        & ResNet50 + MLP (2) & 71.30 & 81.21 & 75.93 & 89.90 & 83.66 & 86.67 & 80.60 & 82.43 & 82.84 & 83,449 & 24,339,366 \\
        \midrule
        \multirow{8}{*}{PrivacyAlert~\cite{Zhao2022ICWSM_PrivacyAlert}}
        & *ResNet50 + SVM~\cite{Zhao2022ICWSM_PrivacyAlert} & 63.90 & 64.40 & 64.20 & 88.10 & 87.90 & 88.00 & 76.00 & 76.43 & 82.00 & -- & -- \\ 
        & *ResNet50-FT~\cite{Zhao2022ICWSM_PrivacyAlert} & 69.90 & 61.30 & 65.30 & 87.60 & 91.20 & 89.40 & 78.75 & 76.25 & 83.70 & -- & --\\
        \cmidrule(lr){2-13}
        & Image Tags + SVM & 53.00 & 74.67 & 61.99 & 90.19 & 77.86 & 83.57 & 71.59 & 76.26 & 77.06 & 0 & 68,805,077 \\
        & Top-10 Scene Tags + SVM & 45.26 & 74.22 & 56.23 & 89.04 & 69.99 & 78.37 & 67.15 & 72.10 & 71.05 & 0 & 24,255,917  \\ 
        & ResNet50 + SVM & 53.90 & \textbf{79.78} & 64.34 & 91.95 & 77.19 & 83.93 & 72.93 & \textbf{78.48} & 77.84 & 0 & 24,255,917  \\ 
        & ResNet50-FT & 45.22 & 70.44 & 55.08 & 87.85 & 71.47 & 78.82 & 66.54 & 70.96 & 71.21 & 23,512,130 & 23,512,130 \\ 
        \rowcolor{mylightgray} \cellcolor{white}& ResNet50 + FC & 63.11 & 63.11 &  63.11 & 87.67 &  87.67 & 87.67 & 75.39 & 75.39 & \textbf{81.51} & \textbf{732} & 24,256,649 \\
        & ResNet50 + MLP (1) & 55.59 & 71.78 & 62.66 & 89.55 & 80.83 & 84.97 & 72.57 & 76.30 & 78.56 & 66,978 & 24,322,895\\ 
        & ResNet50 + MLP (2) & 56.13 & 72.22 & 63.17 & 89.73 & 81.13 & 85.21 & 72.93 & 76.68 & 78.90 & 83,449 & 24,339,366 \\
      \bottomrule \addlinespace[\belowrulesep]
      \multicolumn{13}{l}{\parbox{0.95\linewidth}{\scriptsize{
      Highlighted in gray is the model used for the discussion in the main paper.\\
      $^\diamond$Only neural network parameters are included in this count. Coefficients of SVM are not considered.\\
      *Results are taken from Zhao et al.'s work on PrivacyAlert~\cite{Zhao2022ICWSM_PrivacyAlert}. As some of the images of the dataset are no longer available, performance may be higher for these methods. Unlike Zhao et al.'s work that computes the weighted average precision and recall (BA) across the two classes, giving higher emphasis to the public class that has a higher number of samples, we computed the macro averaging (unweighted mean), treating the two classes equally. \\
      KEY -- P:~precision, R:~recall, ACC:~accuracy, BA:~Balanced accuracy (corresponds to overall recall); S2P:~Scene-to-Privacy; SVM:~Support Vector Machine; FT:~fine-tuning; FC:~fully connected layer.}}}
    \end{tabular}
    \label{tab:scenemodels}
    \vspace{-5pt}
\end{table*}

Table~\ref{tab:scenemodels} compares the classification results of these models on both IPD and PrivacyAlert, and reports and compares the number of trainable parameters and the number of parameters optimised for privacy for all models. For PrivacyAlert, we also report the results of ResNet50+SVM and ResNet50-FT from Zhao et al.'s work~\cite{Zhao2022ICWSM_PrivacyAlert} as a reference for discussion, but we remind the reader of the difference in the data available compared to our experiment. Our pre-trained ResNet-50 combined with the trained SVM achieves the best recall on the private class (81.42\% on IPD and 79.78\% on PrivacyAlert) and the best balanced accuracy (83.08\% on IPD and 78.48\% on PrivacyAlert). Given the small size of the datasets and the features extracted by a pre-trained CNN, the good performance of SVM are expected. However, scaling to larger dataset size is a known drawback of this model to consider. S2P (referred to as ResNet-50 + FC in the table) achieves lower performance, especially because of the lower recall on the private class at 72.44\% on IPD and and 63.11\% on PrivacyAlert. However, the higher recall on the public class, and given the imbalanced distribution of the classes in the datasets, makes S2P achieve a comparable accuracy on IPD (83.12\% for S2P and 83.64\% for ResNet50+SVM) and the highest accuracy  (81.51\%) on PrivacyAlert. The increasing number of parameters optimised for privacy by the variants of S2P allows the model to improve the classification performance compared to S2P, especially in terms of recall of the private class and balanced accuracy on both datasets, and achieve performance more comparable to ResNet-50 with SVM. However, increasing from 1 to 2 hidden layers provides a minimal overall improvement (even if there an increase of about 5~pp on IPD for recall on the private class) and an increase in false positives towards predicting images as private. On IPD, precision of the private class reduces from 75.27\% for the variant with 1 hidden layer to 71.30\% for the variant with 2 hidden layers. Image Tags also achieves comparable performance to S2P in both datasets but at the cost of using an additional CNN, whereas using only scene tags obtains lower overall performance. The latter is caused by the higher number of false positives for the private class and hence a lower recall for the public class. Fine-tuning all the parameters of the ResNet-50 pre-trained for scene recognition to the task of image privacy classification achieves the lowest performance on both datasets and therefore this is not a good strategy to train a model for privacy. Note the difference in performance between our ResNet50+SVM and ResNet50-FT and the same models in Zhao et al.'s work~\cite{Zhao2022ICWSM_PrivacyAlert}. This can denote both a reproducibility issue and the influence of the missing images. 

\section{Learning privacy with objects and MLP}
\label{app:mlp}

We compare the classification performance of 12 variants for MLP with different design choices (Sec.~\ref{app:mlpchoices}), including object features, architecture, and training loss, and analyse the hyper-parameters of the MLP baseline (Sec.~\ref{app:mlphyperparams}). We also discuss MLP-I that uses the resized and reshaped image as input to the classifier (Sec.~\ref{app:mlpimage}).

\subsection{Analysis of different design choices}
\label{app:mlpchoices}

As object features, we consider either only using cardinality or using cardinality and confidence. The first case is chosen for consistency with the feature used in GPA~\cite{Stoidis2022BigMM}. The second case is based on the additional feature used in the same model of a previous work~\cite{Xompero2024CVPRW_XAI4CV}. We also consider a variant of the model architecture in which a batch normalisation layer is added after each FC layer before the output layer. To deal with a highly class imbalance dataset, we consider training the model with the weighted cross-entropy loss giving higher importance to the private class based on the distribution of the training data. The variants based on the batch normalisation and the weighted cross-entropy loss are trained for both cases of object features, resulting in six different models. We also train other six variants using the same design choices as before but normalising the input features with the z-score transformation\footnote{The z-score transformation is commonly referred to as standardisation: the difference between the value of a variable and the mean of the distribution from where the variable is sampled from is divided by the standard deviation of the same distribution.} to handle the sensitivity of MLP to the range of the input features and by using the statistics of the training data. 

Table~\ref{tab:analysismlp} compares the classification performance of the trained models on IPD and PrivacyAlert. The most relevant result is the use of normalisation for the input features, achieving a recall on the private class and a balanced accuracy above 70\% for almost all variants. The most significant improvement is for the model using only the cardinality as feature in both datasets and for the model using both cardinality and confidence as features in PrivacyAlert. For the first model compared to the same model trained without feature normalisation, the improvement is approximately of $+26$ percentage points (pp) for recall on the private class and $+6$~pp for balanced accuracy on IPD, and $+7$~pp for recall on the private class on PrivacyAlert. However, when using batch normalisation in the model architecture, the classification performance are comparable to those of the model trained without the batch normalisation layer, either using or not using feature normalisation. This also shows that batch normalisation negatively impact the effect of the feature normalisation on the input features during training. Using the weighted cross-entropy loss combined with only cardinality as input object feature during the training of the model achieves a recall on the private class of approximately 78\% on IPD and 85\% on PrivacyAlert, outperforming other variants, and a balanced accuracy of approximately 74-75\%, independently of using feature normalisation. When also using confidence as part of the object features, results for the two performance measures are similar to the model with only cardinality as object feature except for the recall on the private class in PrivacyAlert when feature normalisation is not applied (80.22\%). However, this improvement is less significant due to the decrease in the overall accuracy caused by more false positives or less true positives for the public class. 

Ranking the models across the two datasets with a balance between the three performance measures is not straightforward. For a fair comparison with GPA, we selected the variant using feature normalisation and only cardinality as object feature (see Section~\ref{sec:experiments}).

\subsection{Hyper-parameters analysis}
\label{app:mlphyperparams}

The design of an MLP classifier requires to define the number of hidden layers (depth) and the number of neurons for each layer (width). We analyse the classification performance of this baseline while varying these two hyper-parameters. We perform a grid search with the following values: \{1, 2, 3, 5, 7, 10\} for depth, and \{4, 8, 16, 32, 64, 128\} for width. Note that we use the same width for all hidden layers, and the width 128 is larger than the input feature vector (cardinality values for the 80 objects). Specifically, we use the choices made in the previous analysis: only cardinality as feature, feature normalisation, and no weight loss during training. 

Fig.~\ref{fig:mlphyperparams} shows the variations in recall on the private class, balanced accuracy, and accuracy, based on each combination of hyper-parameter values on both IPD and PrivacyAlert. Curves are not necessarily consistent between the two datasets and the best value for the two hyper-parameters is not easily identifiable, depending on the reference performance measure. Considering balanced accuracy and recall on the private class, performance of the model with different parameter values is comparable to each other on IPD, except for when the width is 4. Setting the width to 4 makes the model degenerate to predict only public images when the depth is higher than 5 hidden layers in IPD and when the depth increases to 3 hidden layers or above in PrivacyAlert. PrivacyAlert has a smaller number of images whose content can vary compared to IPD, and therefore the performance measures have larger variations depending on the values of the hyper-parameters. As for the width set to 4, the model degenerates to predict only public images when the width is set to 8 and the depth is set to 7 hidden layers. Using larger widths can lead to a higher recall even with only 3 hidden layers, e.g., using 32 neurons, but the overall accuracy decreases to about 70\% denoting a higher number of false positives. A similar result can be observed when the number of hidden neurons is set to 128 and the depth is set to 7 hidden layers. Other configurations make the model achieve more comparable results on PrivacyAlert. We remind the reader that this classification performance is affected by other choices and hyper-parameters, such as batch size or learning rate, that are not analysed in this experiment.

\subsection{Resized image as input feature vector}
\label{app:mlpimage}

The classifier has 2 FC hidden layers, each halving the dimension of the input feature vector and followed by ReLU, and an FC layer to output the logits of the two classes before applying the softmax for the classification. The input RGB image is resized to a resolution of 64$\times$64 pixels, the values are normalised with respect to the statistics of the ImageNet training set, and then the transformed image is reshaped into a vector with the colour channel concatenated one after the other. These design choices make the number of trainable parameters to be optimised to 99,104,258 (randomly initialised before training the model). The number of parameters of MLP-I is also much higher than MLP or GA-MLP (<2,000 parameters). Table~\ref{tab:compartiveanalysis} reports the classification results of the model on both the IPD and PrivacyAlert datasets, respectively. In both datasets, MLP-I achieves lower performance than MLP: about -10~pp for accuracy and -20~pp for balanced accuracy. Simply resizing, normalising, and reshaping an image as input to an MLP classifier is not sufficient to predict an image as private. 

\begin{table}[t!]
    \centering
    \footnotesize
    \setlength\tabcolsep{1.1pt}
    \caption{
    Comparison of classification performance between MLP variants based on different design choices.
    \vspace{-10pt}
    }
    \begin{tabular}{ccc cc ccc ccc}
    \toprule
    \multicolumn{3}{c}{\textbf{Object features}} & & & \multicolumn{3}{c}{\textbf{Image Privacy Dataset}} & \multicolumn{3}{c}{\textbf{PrivacyAlert}}  \\
    \cmidrule(lr){1-3}\cmidrule(lr){6-8}\cmidrule(lr){9-11}
    Card. & Conf. & Norm. & BN & WL & R (Priv) & BA & ACC & R (Priv) & BA & ACC \\
    \midrule
    \bbox & \wbox & \wbox & \wbox & \wbox & 49.13 & 68.44 & 74.87 & 63.56 & 71.49 & 75.45 \\
    \bbox & \bbox & \wbox & \wbox & \wbox & 69.79 & 75.04 & 76.79 & 59.11 & 71.20 & 77.23 \\
    \bbox & \wbox & \wbox & \bbox & \wbox & 48.83 & 68.22 & 74.68 & 61.78 & 70.78 & 75.28 \\
    \bbox & \bbox & \wbox & \bbox & \wbox & 69.44 & 75.23 & 77.16 & 57.33 & 70.87 & 77.62 \\
    \bbox & \wbox & \wbox & \wbox & \bbox & 78.82 & 74.78 & 73.44 & 85.78 & 73.94 & 68.04 \\
    \bbox & \bbox & \wbox & \wbox & \bbox & 76.43 & 75.49 & 75.17 & 80.22 & 74.10 & 71.05 \\
    \midrule
    \bbox & \wbox & \bbox & \wbox & \wbox & 75.09 & 74.65 & 74.51 & 70.67 & 73.78 & 75.33 \\
    \bbox & \bbox & \bbox & \wbox & \wbox & 72.01 & 75.28 & 76.37 & 75.33 & 74.96 & 74.78 \\
    \bbox & \wbox & \bbox & \bbox & \wbox & 49.39 & 68.36 & 74.68 & 62.67 & 70.82 & 74.89 \\
    \bbox & \bbox & \bbox & \bbox & \wbox & 70.14 & 74.90 & 76.49 & 58.44 & 70.49 & \textbf{76.50} \\
    \bbox & \wbox & \bbox & \wbox & \bbox & \textbf{78.04} & 74.63 & 73.50 & 84.67 & 74.54 & 69.49 \\
    \bbox & \bbox & \bbox & \wbox & \bbox & 73.09 & \textbf{75.76} & \textbf{76.65} & \textbf{85.33} & \textbf{75.47} & 70.55 \\
    \bottomrule \addlinespace[\belowrulesep]
    \multicolumn{11}{l}{\parbox{0.96\linewidth}{\scriptsize{KEY -- Card.:~cardinality, Conf.:~confidence, Norm:~input features are normalised based on training statistics; BN:~batch normalisation; WL:~weighted cross-entropy loss; R:~recall, BA:~balanced accuracy, ACC:~accuracy, Priv.:~private; \wbox~not considered, \bbox~considered.}}}
    \end{tabular}
    \label{tab:analysismlp}
\end{table}

%%%%%%
\pgfplotstableread{mlp_params_privacyalert.txt}\bamlparams
\pgfplotstableread{mlp_params_ipd.txt}\mlparamsipd

\begin{figure}[t!]
    \centering
    \footnotesize
    \begin{tikzpicture}
        \begin{axis}[
        width=0.60\columnwidth,
        height=0.5\columnwidth,
        ylabel={Recall (Priv)},
        xmin=0, xmax=11,
        ymin=0, ymax=90,
        label style={font=\footnotesize},
        tick label style={font=\scriptsize},
        xticklabels={},
        xlabel near ticks,
        ylabel near ticks,
        title={PrivacyAlert},
        title style={font=\footnotesize},
    ]
    \addplot+ [solid, color=dc1, mark=*, mark options={mark size=1.5pt,fill=dc1}] table[x=Depth, y=RPriv1]{\bamlparams};
    \addplot+ [solid, color=dc2, mark=*, mark options={mark size=1.5pt,fill=dc2}] table[x=Depth, y=RPriv2]{\bamlparams};
    \addplot+ [solid, color=dc3, mark=*, mark options={mark size=1.5pt,fill=dc3}] table[x=Depth, y=RPriv3]{\bamlparams};
    \addplot+ [solid, color=dc4, mark=*, mark options={mark size=1.5pt,fill=dc4}] table[x=Depth, y=RPriv4]{\bamlparams};
    \addplot+ [solid, color=dc5, mark=*, mark options={mark size=1.5pt,fill=dc5}] table[x=Depth, y=RPriv5]{\bamlparams};
    \addplot+ [solid, color=dc6, mark=*, mark options={mark size=1.5pt,fill=dc6}] table[x=Depth, y=RPriv6]{\bamlparams};
    \end{axis}
    \end{tikzpicture}
    \begin{tikzpicture}
        \begin{axis}[
        width=0.60\columnwidth,
        height=0.5\columnwidth,
        xmin=0, xmax=11,
        ymin=0, ymax=90,
        label style={font=\footnotesize},
        tick label style={font=\scriptsize},
        yticklabels={},
        xticklabels={},
        xlabel near ticks,
        ylabel near ticks,
        title={IPD},
        title style={font=\footnotesize},
    ]
    \addplot+ [solid, color=dc1, mark=*, mark options={mark size=1.5pt,fill=dc1}] table[x=Depth, y=RPriv1]{\mlparamsipd};
    \addplot+ [solid, color=dc2, mark=*, mark options={mark size=1.5pt,fill=dc2}] table[x=Depth, y=RPriv2]{\mlparamsipd};
    \addplot+ [solid, color=dc3, mark=*, mark options={mark size=1.5pt,fill=dc3}] table[x=Depth, y=RPriv3]{\mlparamsipd};
    \addplot+ [solid, color=dc4, mark=*, mark options={mark size=1.5pt,fill=dc4}] table[x=Depth, y=RPriv4]{\mlparamsipd};
    \addplot+ [solid, color=dc5, mark=*, mark options={mark size=1.5pt,fill=dc5}] table[x=Depth, y=RPriv5]{\mlparamsipd};
    \addplot+ [solid, color=dc6, mark=*, mark options={mark size=1.5pt,fill=dc6}] table[x=Depth, y=RPriv6]{\mlparamsipd};
    \end{axis}
    \end{tikzpicture}
    \\
    \begin{tikzpicture}
        \begin{axis}[
        width=0.60\columnwidth,
        height=0.5\columnwidth,
        ylabel={Balanced accuracy},
        xmin=0, xmax=11,
        ymin=45, ymax=80,
        label style={font=\footnotesize},
        tick label style={font=\scriptsize},
        xticklabels={},
        xlabel near ticks,
        ylabel near ticks,
    ]
    \addplot+ [solid, color=dc1, mark=*, mark options={mark size=1.5pt,fill=dc1}] table[x=Depth, y=BA1]{\bamlparams};
    \addplot+ [solid, color=dc2, mark=*, mark options={mark size=1.5pt,fill=dc2}] table[x=Depth, y=BA2]{\bamlparams};
    \addplot+ [solid, color=dc3, mark=*, mark options={mark size=1.5pt,fill=dc3}] table[x=Depth, y=BA3]{\bamlparams};
    \addplot+ [solid, color=dc4, mark=*, mark options={mark size=1.5pt,fill=dc4}] table[x=Depth, y=BA4]{\bamlparams};
    \addplot+ [solid, color=dc5, mark=*, mark options={mark size=1.5pt,fill=dc5}] table[x=Depth, y=BA5]{\bamlparams};
    \addplot+ [solid, color=dc6, mark=*, mark options={mark size=1.5pt,fill=dc6}] table[x=Depth, y=BA6]{\bamlparams};
    \end{axis}
    \end{tikzpicture}
    \begin{tikzpicture}
        \begin{axis}[
        width=0.60\columnwidth,
        height=0.5\columnwidth,
        xmin=0, xmax=11,
        ymin=45, ymax=80,
        label style={font=\footnotesize},
        tick label style={font=\scriptsize},
        yticklabels={},
        xticklabels={},
        xlabel near ticks,
        ylabel near ticks,
    ]
    \addplot+ [solid, color=dc1, mark=*, mark options={mark size=1.5pt,fill=dc1}] table[x=Depth, y=BA1]{\mlparamsipd};
    \addplot+ [solid, color=dc2, mark=*, mark options={mark size=1.5pt,fill=dc2}] table[x=Depth, y=BA2]{\mlparamsipd};
    \addplot+ [solid, color=dc3, mark=*, mark options={mark size=1.5pt,fill=dc3}] table[x=Depth, y=BA3]{\mlparamsipd};
    \addplot+ [solid, color=dc4, mark=*, mark options={mark size=1.5pt,fill=dc4}] table[x=Depth, y=BA4]{\mlparamsipd};
    \addplot+ [solid, color=dc5, mark=*, mark options={mark size=1.5pt,fill=dc5}] table[x=Depth, y=BA5]{\mlparamsipd};
    \addplot+ [solid, color=dc6, mark=*, mark options={mark size=1.5pt,fill=dc6}] table[x=Depth, y=BA6]{\mlparamsipd};
    \end{axis}
    \end{tikzpicture}
    \\
    \begin{tikzpicture}
        \begin{axis}[
        width=0.60\columnwidth,
        height=0.5\columnwidth,
        xlabel={Number of hidden layers},
        ylabel={Accuracy (\%)},
        xmin=0, xmax=11,
        ymin=65, ymax=80,
        label style={font=\footnotesize},
        tick label style={font=\scriptsize},
        xtick={1,2,3,5,7,10},
        xlabel near ticks,
        ylabel near ticks,
        title style={font=\footnotesize,at={(0.5,0.9)}},
    ]
    \addplot+ [solid, color=dc1, mark=*, mark options={mark size=1.5pt,fill=dc1}] table[x=Depth, y=ACC1]{\bamlparams};
    \addplot+ [solid, color=dc2, mark=*, mark options={mark size=1.5pt,fill=dc2}] table[x=Depth, y=ACC2]{\bamlparams};
    \addplot+ [solid, color=dc3, mark=*, mark options={mark size=1.5pt,fill=dc3}] table[x=Depth, y=ACC3]{\bamlparams};
    \addplot+ [solid, color=dc4, mark=*, mark options={mark size=1.5pt,fill=dc4}] table[x=Depth, y=ACC4]{\bamlparams};
    \addplot+ [solid, color=dc5, mark=*, mark options={mark size=1.5pt,fill=dc5}] table[x=Depth, y=ACC5]{\bamlparams};
    \addplot+ [solid, color=dc6, mark=*, mark options={mark size=1.5pt,fill=dc6}] table[x=Depth, y=ACC6]{\bamlparams};
    \end{axis}
    \end{tikzpicture}
    \begin{tikzpicture}
        \begin{axis}[
        width=0.60\columnwidth,
        height=0.5\columnwidth,
        xlabel={Number of hidden layers},
        xmin=0, xmax=11,
        ymin=65, ymax=80,
        label style={font=\footnotesize},
        tick label style={font=\scriptsize},
        yticklabels={},
        xtick={1,2,3,5,7,10},
        xlabel near ticks,
        ylabel near ticks,
        title style={font=\footnotesize,at={(0.5,0.9)}},
    ]
    \addplot+ [solid, color=dc1, mark=*, mark options={mark size=1.5pt,fill=dc1}] table[x=Depth, y=ACC1]{\mlparamsipd};
    \addplot+ [solid, color=dc2, mark=*, mark options={mark size=1.5pt,fill=dc2}] table[x=Depth, y=ACC2]{\mlparamsipd};
    \addplot+ [solid, color=dc3, mark=*, mark options={mark size=1.5pt,fill=dc3}] table[x=Depth, y=ACC3]{\mlparamsipd};
    \addplot+ [solid, color=dc4, mark=*, mark options={mark size=1.5pt,fill=dc4}] table[x=Depth, y=ACC4]{\mlparamsipd};
    \addplot+ [solid, color=dc5, mark=*, mark options={mark size=1.5pt,fill=dc5}] table[x=Depth, y=ACC5]{\mlparamsipd};
    \addplot+ [solid, color=dc6, mark=*, mark options={mark size=1.5pt,fill=dc6}] table[x=Depth, y=ACC6]{\mlparamsipd};
    \end{axis}
    \end{tikzpicture}
    \caption{Analysis of the MLP performance when varying the number of hidden layers and number of neurons in the hidden layers. The number of neurons are kept fixed across the hidden layers. Note the different y-axis limits for visualisation purposes.
    Legend:
    \protect\tikz \protect\fill[dc1,fill=dc1] (1,1) circle (0.5ex);~4 neurons,
    \protect\tikz \protect\fill[dc2,fill=dc2] (1,1) circle (0.5ex);~8 neurons,
    \protect\tikz \protect\fill[dc3,fill=dc3] (1,1) circle (0.5ex);~16 neurons,
    \protect\tikz \protect\fill[dc4,fill=dc4] (1,1) circle (0.5ex);~32 neurons,
    \protect\tikz \protect\fill[dc5,fill=dc5] (1,1) circle (0.5ex);~64 neurons,
    \protect\tikz \protect\fill[dc6,fill=dc6] (1,1) circle (0.5ex);~128 neurons. 
    }
    \label{fig:mlphyperparams}
    \vspace{-10pt}
\end{figure}
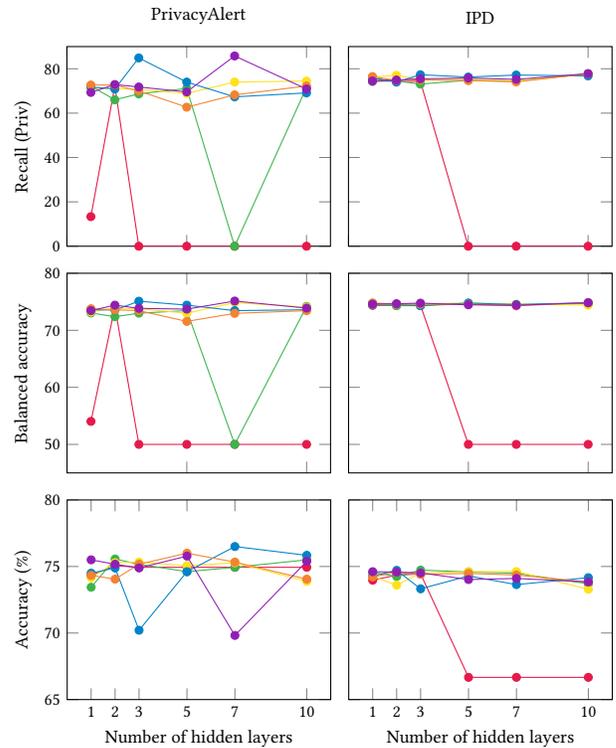
%%%%%

\begin{table*}[t!]
    \centering
    \footnotesize
    \caption{Classification results and comparison of GA-MLP variants.
    \vspace{-10pt}
    }    
    \begin{tabular}{cccc cc cccc cccc}
    \toprule
    \multicolumn{4}{c}{\textbf{Object features}} & & & \multicolumn{4}{c}{\textbf{Image Privacy Dataset}} & \multicolumn{4}{c}{\textbf{PrivacyAlert}}  \\
    \cmidrule(lr){1-4}\cmidrule(lr){7-10}\cmidrule(lr){11-14}
    Card. & Conf. & Norm. & Tran. & BN & WL & R (Priv) & BA & ACC & Rep. & R (Priv) & BA & ACC & Rep.*\\
    \midrule
    \bbox & \wbox & \wbox & \wbox & \wbox & \wbox & 28.60 & 59.01 & 69.14 & \bbox &  4.00	& 51.11 & 74.61 & \bbox \\
    \bbox & \wbox & \wbox & \bbox & \wbox & \wbox & 28.60 & 59.01 & 69.14 & \bbox & 14.00	& 53.36 & 73.00 & \wbox \\
    \bbox & \bbox & \wbox & \wbox & \wbox & \wbox & 47.79 & 64.71 & 70.36 & \wbox &  2.22	& 50.52 & 74.61 & \wbox \\
    \bbox & \bbox & \wbox & \bbox & \wbox & \wbox & 36.59 & 62.22 & 70.76 & \wbox &  4.00	& 51.00 & 74.44 & \bbox \\
    \bbox & \wbox & \wbox & \wbox & \bbox & \wbox & \textbf{75.65} & \textit{74.71} & \textit{74.39} & \wbox & 68.89	& \textbf{72.82} & \textit{74.78} & \wbox \\
    \bbox & \bbox & \wbox & \wbox & \bbox & \wbox & \textit{70.83} & \textbf{74.98} & \textbf{76.36} & \wbox & 60.67	& \textit{71.27}	& \textbf{76.56} & \wbox \\
    \bbox & \wbox & \wbox & \wbox & \wbox & \bbox & 53.26 & 62.28 & 65.29 & \bbox & \textbf{96.67}	& 58.88	& 40.03 & \bbox \\
    \bbox & \bbox & \wbox & \wbox & \wbox & \bbox & 62.80 & 66.57 & 67.82 & \bbox & \textit{86.00}	& 65.81	& 55.73 & \wbox \\    
    \midrule
    \bbox & \wbox & \bbox & \wbox & \wbox & \wbox &  0.00 & 50.00 & 66.67 & \bbox &  4.00 & 51.11 & 74.61 & \bbox \\
    \bbox & \wbox & \bbox & \bbox & \wbox & \wbox & 34.85 & 60.88 & 69.56 & \bbox &  4.44 & 51.18 & 74.50 & \wbox \\
    \bbox & \bbox & \bbox & \wbox & \wbox & \wbox & 38.50 & 62.38 & 70.34 & \wbox &  2.00 & 50.41 & 74.55 & \wbox \\
    \bbox & \bbox & \bbox & \bbox & \wbox & \wbox & 39.93 & 62.52 & 70.05 & \wbox &  4.00 & 51.00 & 74.44 & \bbox \\
    \bbox & \wbox & \bbox & \wbox & \bbox & \wbox &  0.00 & 50.00 & 66.67 & \bbox & 69.11 & 72.93 & 74.83 & \wbox \\
    \bbox & \bbox & \bbox & \wbox & \bbox & \wbox & 68.14 & 74.87 & 77.11 & \wbox & 66.44 & 72.60 & 75.67 & \wbox \\
    \bbox & \wbox & \bbox & \wbox & \wbox & \bbox & 53.26 & 62.28 & 65.29 & \bbox & 96.67 & 58.88 & 40.03 & \bbox \\
    \bbox & \bbox & \bbox & \wbox & \wbox & \bbox & 60.11 & 66.50 & 68.63 & \wbox & 78.89 & 66.08 & 59.69 & \bbox \\
    \bottomrule \addlinespace[\belowrulesep]
    \multicolumn{14}{l}{\parbox{0.7\linewidth}{\scriptsize{*Note that implementation of the method is affected by non-deterministic behaviours across various variants\\KEY -- Card.:~cardinality, Conf.:~confidence, Norm:~input features are normalised based on training statistics; Trans.:~each input feature is transformed into a higher dimensional vector via a fully connected layer; BN:~batch normalisation; WL:~weighted cross-entropy loss; R:~recall, BA:~balanced accuracy, ACC:~accuracy, Priv.:~private; Rep.:~repeatability of training procedure and results; \wbox~not considered, \bbox~considered.}}}
    \end{tabular}
    \label{tab:analysisgraphanostic}
    \vspace{-10pt}
\end{table*}

\begin{table}[t!]
    \centering
    \footnotesize
    \caption{Analysis of GA-MLP over five runs to quantify the variability of batch normalisation.
    \vspace{-10pt}
    }
    % \vspace{-7pt}
    \begin{tabular}{c ccc ccc}
    \toprule
    \textbf{Run} & \multicolumn{3}{c}{\textbf{Image Privacy Dataset}} & \multicolumn{3}{c}{\textbf{PrivacyAlert}} \\
    \cmidrule(lr){2-4}\cmidrule(lr){5-7}
    & R (Priv) & BA & ACC & R (Priv) & BA & ACC \\
    \midrule
        1 & 75.65 & 74.71 & 74.39 & 68.89 & 72.82 & 74.78 \\ 
        2 & 76.22 & 74.57 & 74.02 & 71.33 & 74.23 & 75.67\\ 
        3 & 76.56 & 74.62 & 73.97 & 70.22 & 73.56 & 75.22 \\
        4 & 75.74 & 74.67 & 74.32 & 67.11 & 72.71 & 75.50 \\
        5 & 78.30 & 74.27 & 72.93 & 71.56 & 74.04 & 75.28 \\
    \midrule
        mean & 76.49 & 74.57 & 73.93 & 69.82 & 73.47 & 75.29\\
        std & $\pm1.07$ & $\pm0.17$ & $\pm0.59$ & $\pm1.85$ & $\pm0.69$ & $\pm0.34$ \\
    \bottomrule
    \end{tabular}    
    \label{tab:gamlpruns}
    \vspace{-10pt}
\end{table}

\section{Graph-agnostic model}
\label{app:gamlp}

We compare the classification performance of sixteen variants for GA-MLP based on different design choices, including object features, architecture, and training loss.

A previous work~\cite{Xompero2024CVPRW_XAI4CV} used confidence and cardinality as object features, and included an FC layer to transform the input features and a batch normalisation layer after the FC layer of each block. We analyse variants of this model adding each of these components to the simplest variant that only uses the object cardinality as node feature in input. The second variant includes the confidence as second object feature. Other two variants use the FC layer to transform the input features, each variant depending on the object features used as input. Similarly, other two variants use the batch normalisation layer in the model architecture, each variant depending on the input object features. The last two variants use a weighted cross-entropy loss as objective function to give higher importance to the private class based on the distribution of the training data. To understand if the behaviour observed for MLP is also repeating with GA-MLP, we consider other eight variants that use the same design choices as the previous ones but normalise the input features with the z-score transformation based on the statistics of the training data. 

Table~\ref{tab:analysisgraphanostic} compares the classification performance of the trained models on IPD and PrivacyAlert. Avoiding using batch normalisation or the weighted cross-entropy loss makes the performance low, especially for the recall on the private class that is lower than 50\%. For PrivacyAlert, the performance of these trained models, either using or not using feature normalisation (rows 1-4 and 9-12), is close to the degenerate case of predicting only the public class for all images in the testing set. On the contrary, using batch normalisation makes the model achieve the highest performance measures, especially when considering accuracy and balanced accuracy jointly. However, adding feature normalisation to these models does not provide a significant impact and when using only cardinality as object feature, the model degenerates to predict only the public class in IPD. As expected, when training the model with weighted cross-entropy loss, the recall on the private class can achieve performance higher than 50\% on IPD and above 78\% (up to 96.67\%) on PrivacyAlert. However, accuracy is significantly and negatively affected with values lower than 70\% (40\% on PrivacyAlert, see rows 7 and 15), showing that there is higher number of false positive and less images correctly predicted as public. Using confidence as additional feature can improve the performance of the re-trained model but not sufficiently to be comparable to the model using batch normalisation. Moreover, adding feature normalisation does not have any effect when using only cardinality.

For a fair comparison with GPA~\cite{Stoidis2022BigMM}, we select the variant that uses batch normalisation and only cardinality as input object feature. As the training and results of this model cannot be replicated (row 5 of Table~\ref{tab:analysisgraphanostic}), we analyse the performance over 5 runs and report the classification results, mean and standard deviation in Table~\ref{tab:gamlpruns}. Keeping fixed the values of all hyper-parameters and seed, the training and results of the model can be reproduced with a small standard deviation (note the largest standard deviation of $\pm$1.85 for the recall on the private class in PrivacyAlert). We use the results of the model trained in the first run for the discussion in the main paper (see Section~\ref{sec:experiments}). 

\end{document}